\documentclass[sigconf, screen]{acmart}
\AtBeginDocument{%
  }

\setcopyright{none}

\copyrightyear{2026}
\acmYear{2026}

\acmDOI{}
\acmISBN{}

\acmDOI{XXXXXXX.XXXXXXX}
\acmConference[Manuscript under review]{}{Manuscript under review}{}

\begin{document}

\title{From Weight Perturbation to Feature Attribution for Explaining Fully Connected Neural Networks.}



\author{Thodoris Lymperopoulos}
\affiliation{%
  \institution{NCSR Demokritos}
  \city{Athens}
  \country{Greece}}
\email{t.lymperopoulos@iit.demokritos.gr}

\author{Denia Kanellopoulou}
\affiliation{%
  \institution{NCSR Demokritos}
  \city{Athens}
  \country{Greece}}
\email{denia@iit.demokritos.gr}

\renewcommand{\shortauthors}{Lymperopoulos et al.}

\begin{abstract}
Fully Connected Neural Networks (FCNNs) are often regarded as simple and intuitive architectures, yet they serve as the foundation for more complex models. 
Nonetheless, the lack of consensus on their interpretability continues to pose challenges, underscoring the enduring relevance of simpler, attribution-based 
approaches for understanding even the most advanced neural architectures. In this regard, we explore a novel idea for estimating feature attribution, by 
applying perturbation to the features' attached weights instead of their values. This method offers a fresh perspective aimed at mitigating common limitations in 
Occlusion techniques, such as Added Bias and Out-of-Distribution data. The application of this rule leads to the formation of a pair of novel attribution methods we call 
XWP and XWP\textsuperscript{c}. 
Founded on simple rules, our methods achieve competitive performance in identifying image signals for simple DNNs, competing with the most established 
attribution methods on standard baseline metrics. Our work thus contributes to the field of Explainability by introducing a robust framework that paves the way 
for addressing these long-standing vulnerabilities, and leads to more reliable and interpretable model explanations.
\end{abstract}




\begin{teaserfigure}
    \centering
    \setlength{\tabcolsep}{2pt}  
    \begin{tabular}{c c c c c c}
        \includegraphics[width=0.148\textwidth]{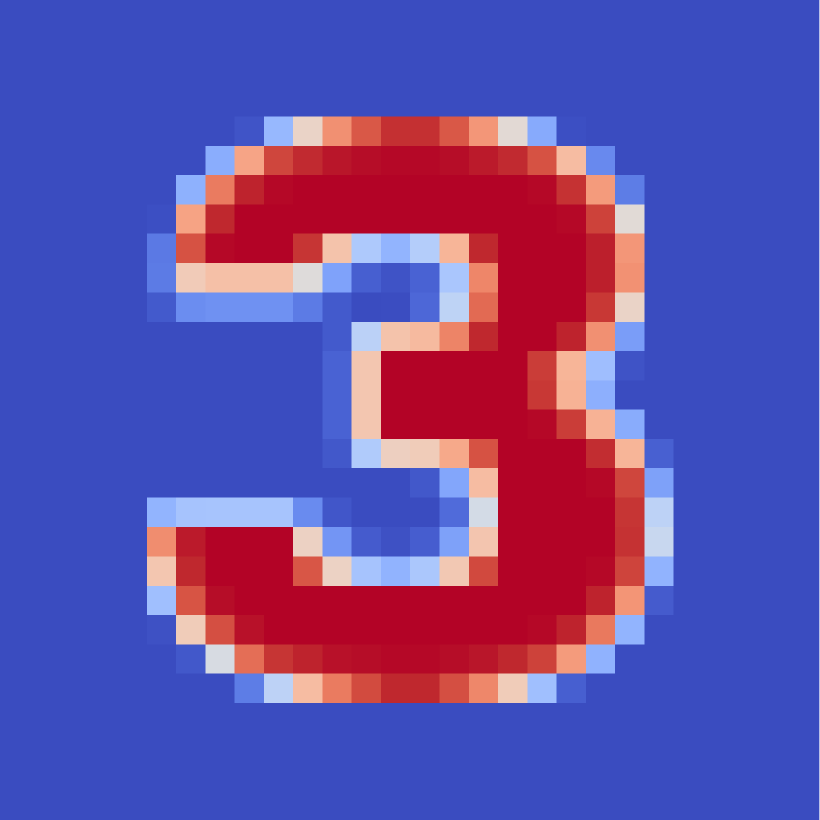} &
        \includegraphics[width=0.148\textwidth]{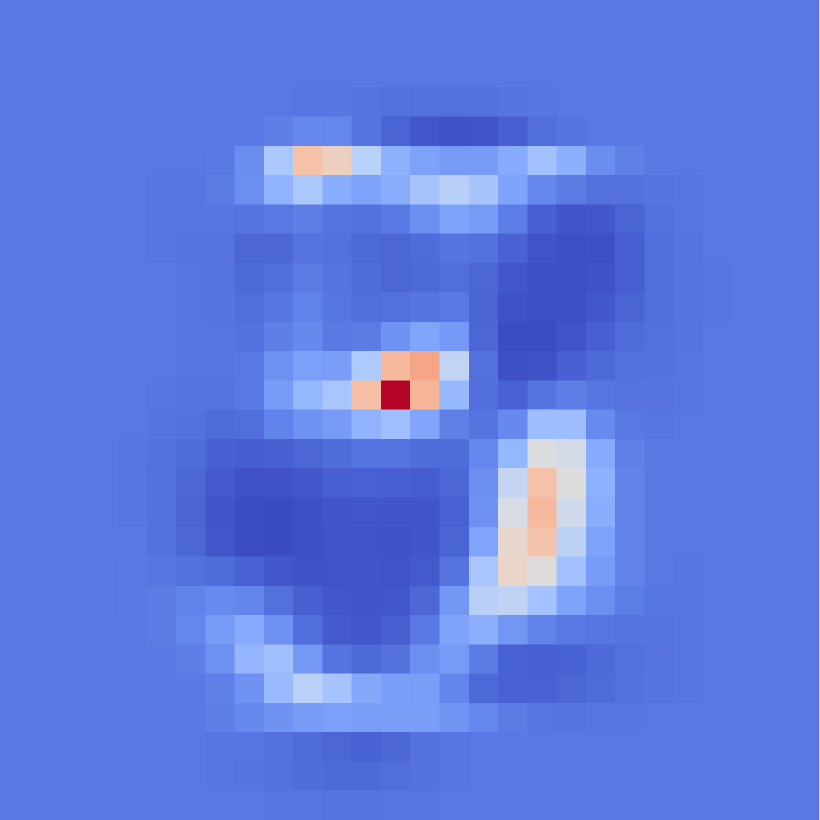} &
        \includegraphics[width=0.148\textwidth]{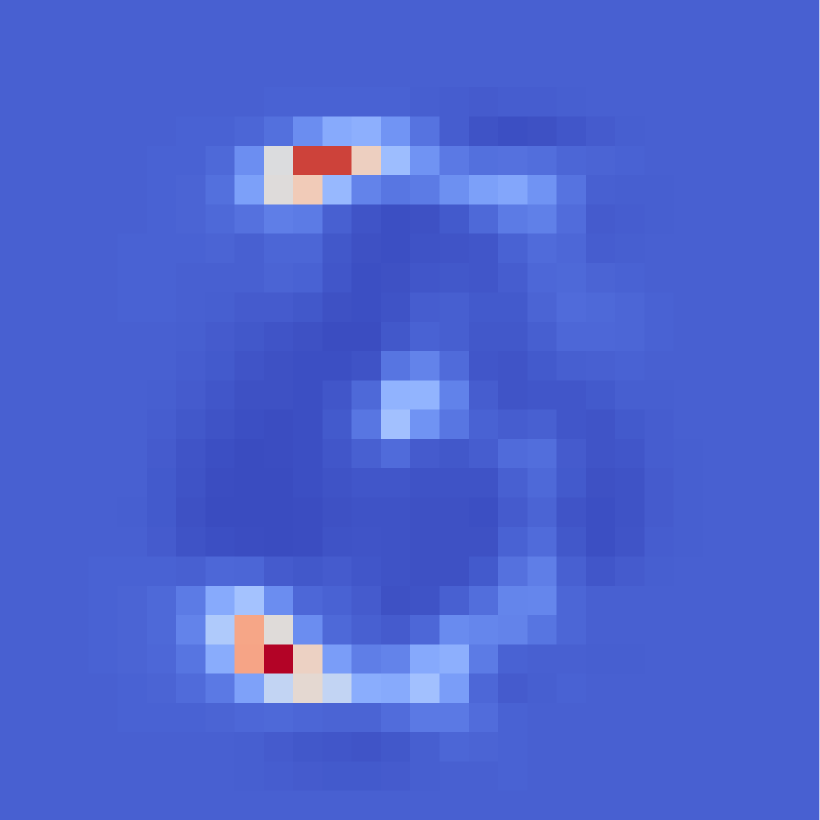} &
        \hspace{1cm}
        \includegraphics[width=0.148\textwidth]{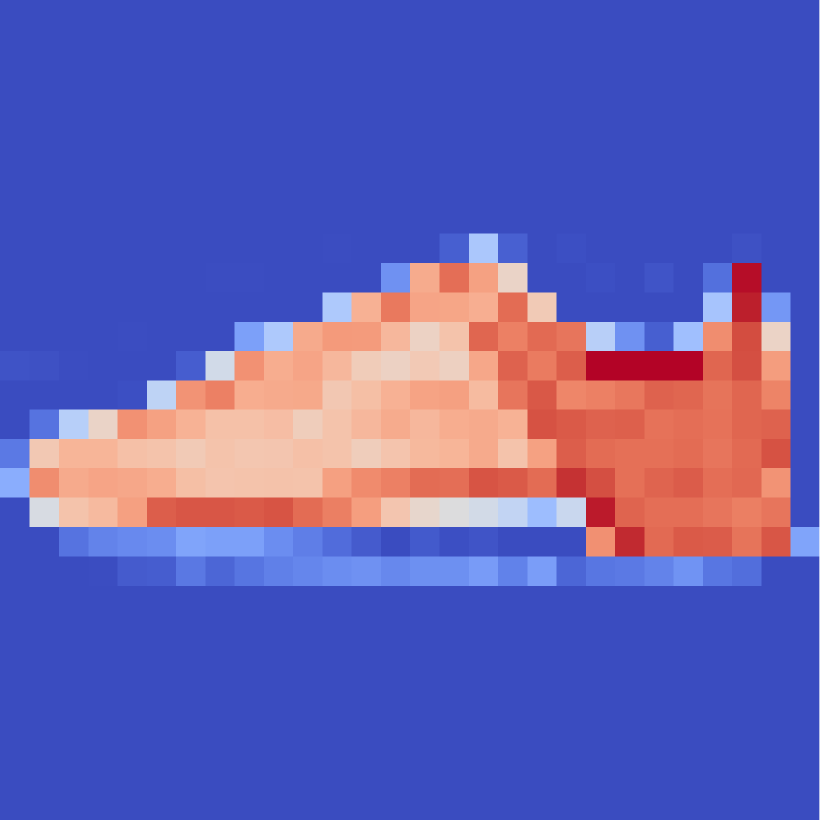} &
        \includegraphics[width=0.148\textwidth]{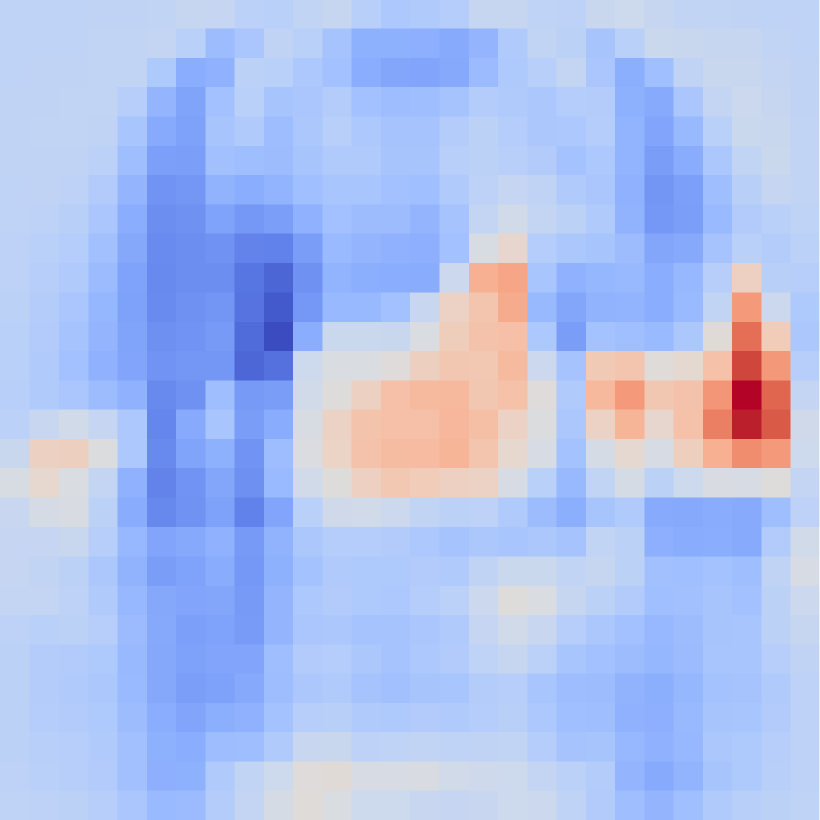} &
        \includegraphics[width=0.148\textwidth]{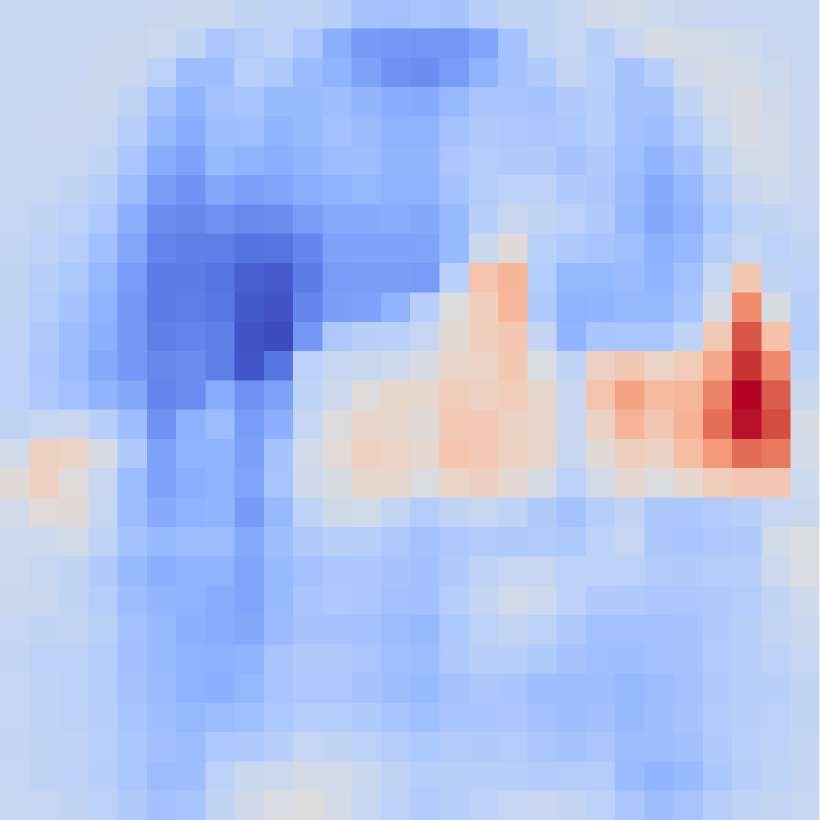} \\[3pt]
        \includegraphics[width=0.148\textwidth]{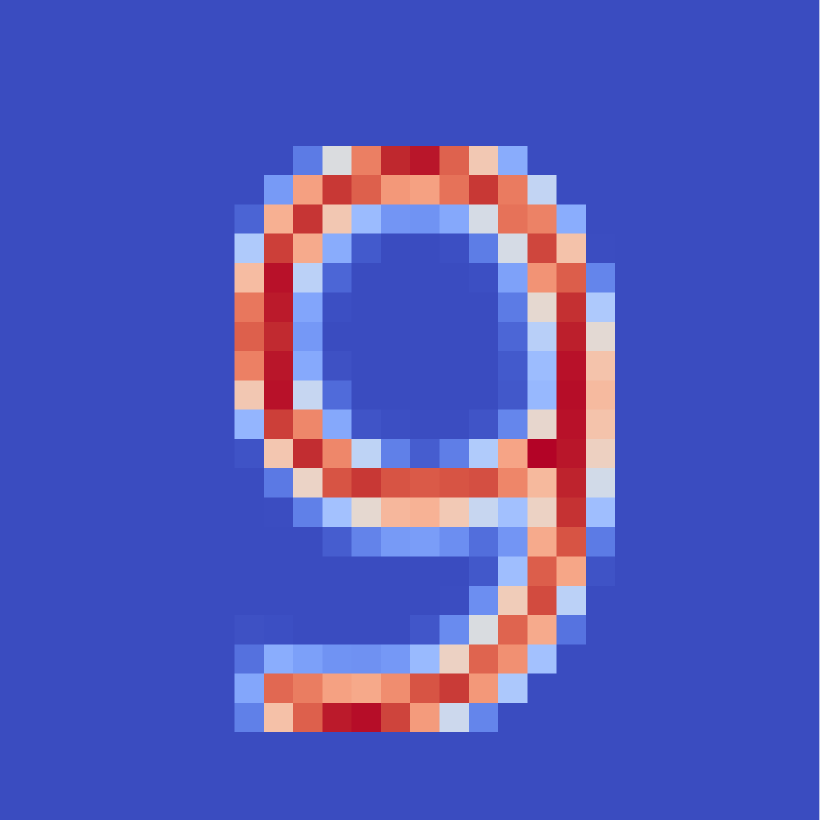} &
        \includegraphics[width=0.148\textwidth]{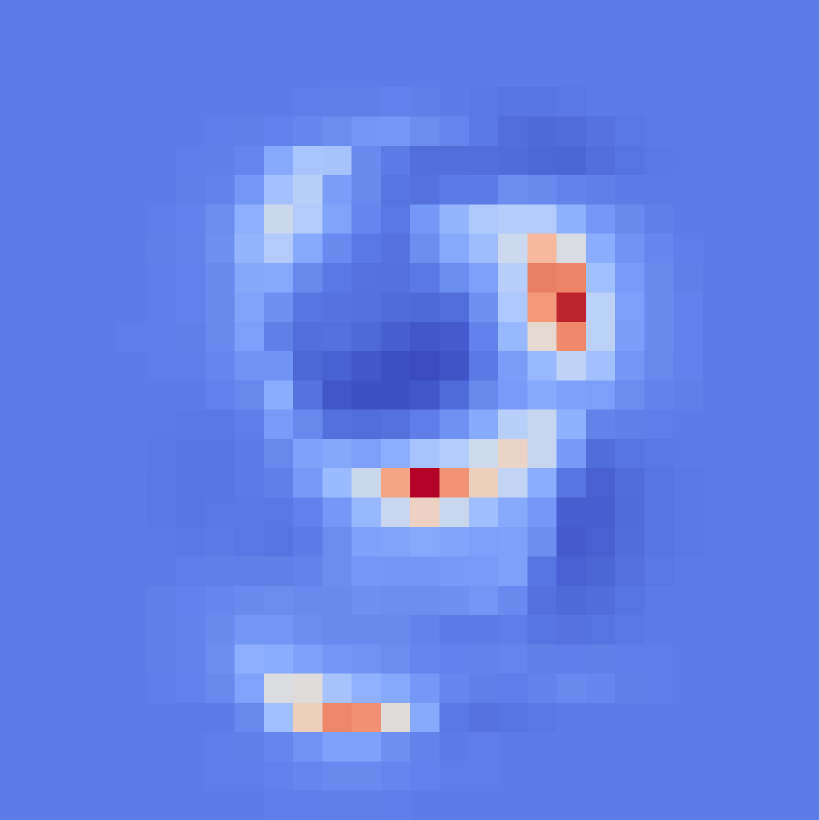} &
        \includegraphics[width=0.148\textwidth]{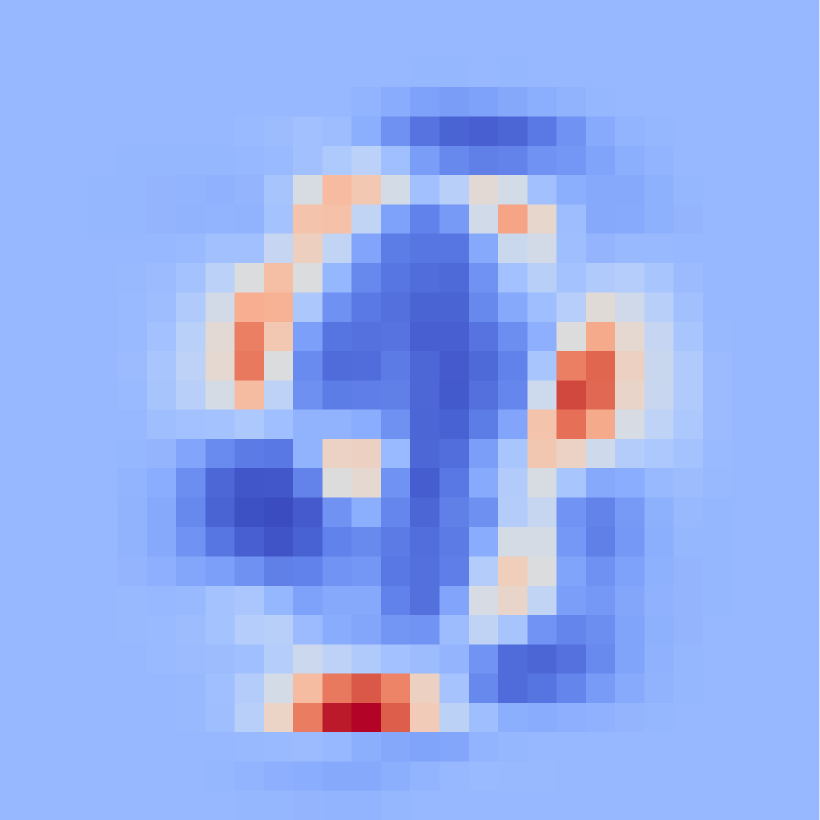} &
        \hspace{1cm}
        \includegraphics[width=0.148\textwidth]{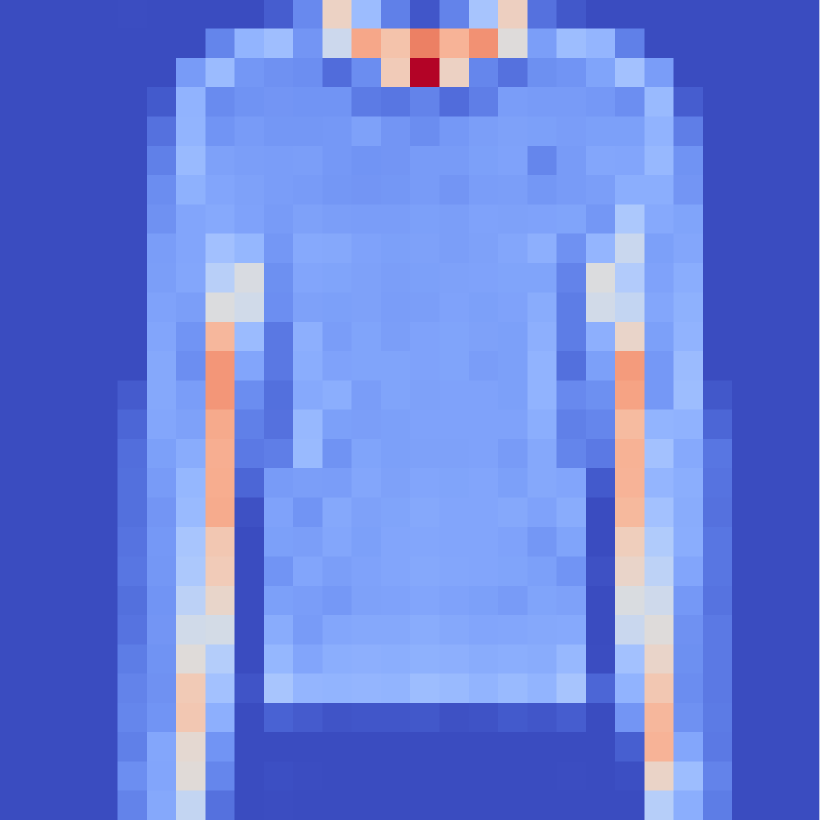} &
        \includegraphics[width=0.148\textwidth]{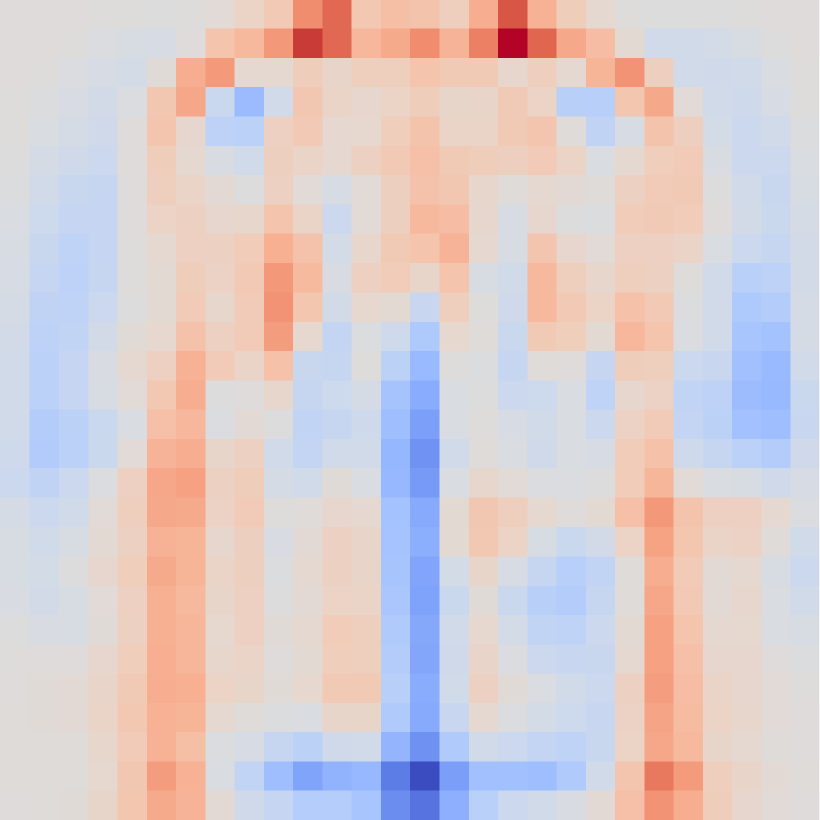} &
        \includegraphics[width=0.148\textwidth]{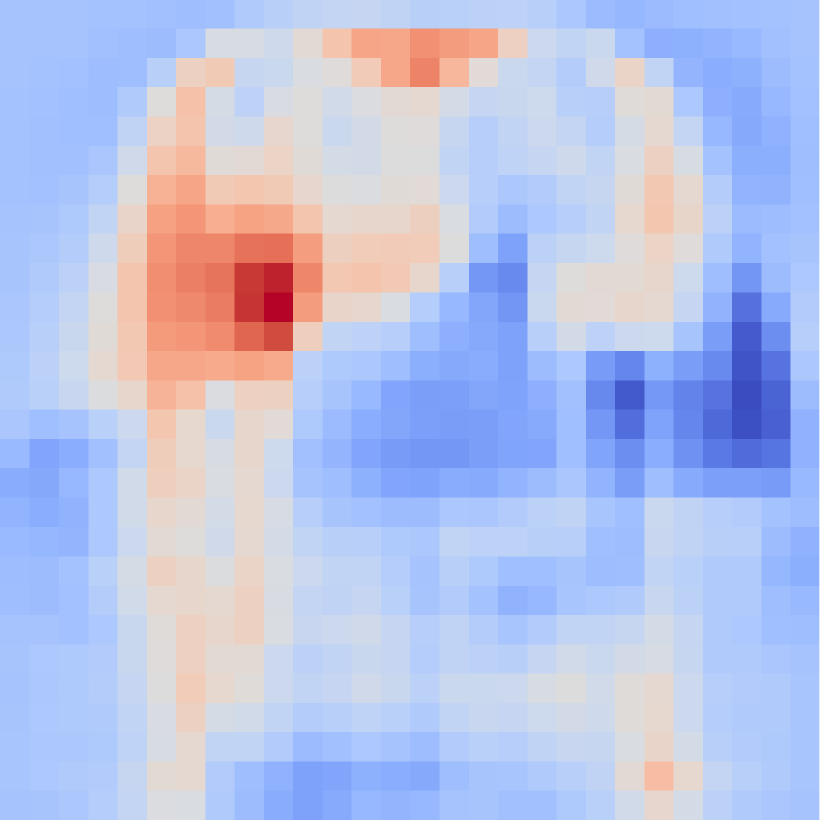} \\
    \end{tabular}
    \caption{The produced attribution maps for our proposed methods, using weight perturbation for estimating feature attribution.}
    \label{fig:teaser}
\end{teaserfigure}


\maketitle

\section{Introduction}

Modern deep learning models have achieved remarkable performance across a wide range of tasks, from image recognition to natural language processing. 
Yet, this success comes at a cost: as models grow in scale and complexity, their internal decision-making process becomes increasingly opaque, rendering 
them, in effect, black boxes. This opacity poses significant challenges in high-stakes domains such as medicine and autonomous systems, where understanding 
and trusting a model's predictions is as important as the predictions themselves. The field of Explainable AI (XAI) \cite{BARREDOARRIETA202082} emerged precisely 
to address this gap, seeking to provide human-interpretable insights into the model's behavior \cite{Linardatos2020ExplainableAA}.

Attribution methods represent the earliest and most widely adopted family of approaches within XAI, aiming to quantify the contribution of each input 
feature to the model's output \cite{10.1007/978-3-319-10590-1_53}. The core intuition is to produce a saliency map — a score assigned to each input 
dimension — reflecting its relative importance for a given prediction. Over time, a range of methodologies has been developed to estimate these scores. 
Gradient-based methods, such as Vanilla Gradients \cite{Simonyan2013DeepIC} and 
Integrated Gradients \cite{sundararajan2017axiomaticattributiondeepnetworks}, derive these scores directly from the model's gradients with respect to 
the input. Backpropagation-based methods, such as Layer-wise Relevance Propagation (LRP) \cite{lrpBinder} and DeepLIFT \cite{10.5555/3305890.3306006}, 
instead propagate the output signal backwards through the network, redistributing relevance across layers according to defined propagation rules. 
Perturbation-based methods, such as LIME \cite{10.1145/2939672.2939778} and SHAP \cite{10.5555/3295222.3295230}, approximate the model locally by observing 
how the output changes as portions of the input are systematically masked or altered. Collectively, these methods constitute a robust framework for interpreting 
a model's decisions.

In a separate body of work \cite{olah2020zoom} authors introduced a complementary perspective, applying well-established techniques — weight visualization and neuron 
optimization — to probe the model's internal representations directly. Yet, the scale and complexity of modern architectures rendered manual inspection of 
individual components intractable, motivating the development of Mechanistic Interpretability (MI) \cite{bereska2024mechanisticinterpretabilityaisafety}
— a systematic framework for reverse-engineering a model's decision-making process by identifying the circuits and components responsible for specific 
computations. As the field matured, attention shifted to more complex architectures \cite{elhage2021mathematical}, and MI established a natural 
connection to the field of Causality \cite{kaddour2022causalmachinelearningsurvey, geiger2025causalabstractiontheoreticalfoundation}: both disciplines 
share a commitment to interventional reasoning, seeking not merely to correlate inputs with outputs but to identify the underlying mechanisms that govern
model behavior. This convergence gave rise to structured methods for uncovering causal relationships among model components, yielding foundational 
insights into how neural networks encode and process information \cite{bereska2024mechanisticinterpretabilityaisafety, geiger2025causalabstractiontheoreticalfoundation}.

Despite the recent success of MI to interpret different aspects of the model's inner processes, no definite consensus on the model's interpretability 
has been reached to this day, even for the simplest among complex architectures — the Fully Connected Neural Networks (FCNNs) \cite{elhage2021mathematical}. This 
comprises an essential challenge, since FCNNs serve as core components in Transformers \cite{10.5555/3295222.3295349}. 
In this work, we depart from conventional occlusion approaches by revisiting the fundamentals of model computation through the lens of weight perturbation.
Adopting this perspective, we formulate a novel framework for attribution estimation, which circumvents common pitfalls of input perturbation in existing methodologies,
such as the problem of Out-of-Distribution data. Within this framework, we introduce two variants—XWP and XWP\textsuperscript{c}— designed to assess attribution scores 
with improved accuracy.  We validate their efficacy on FCNNs, where heatmaps generated for the Typeface MNIST and Fashion MNIST datasets (Figure~\ref{fig:teaser}) 
reveal clean, input-aligned patterns, devoid of the artifacts prevalent in prior work. Together, these variants provide a synergistic approach to explaining a model’s 
decision-making process, establishing a novel paradigm for perturbation-based interpretability. 

\section{Related Work}

\textbf{Occlusion Methods.} They comprise a prominent family of attribution methods relying on occlusion and perturbation of input data to quantify 
the contribution of individual features or regions to a model's prediction. One of the earliest approaches, Occlusion Sensitivity \cite{10.1007/978-3-319-10590-1_53}, 
slides a mask patch across an input image and observes the resulting change in output confidence, producing a saliency map that reflects each region's 
importance. Building on this principle, LIME (Local Interpretable Model-agnostic Explanations) \cite{10.1145/2939672.2939778} generates perturbed 
samples around a given input, weights them by their proximity to the original, and fits a sparse linear model to approximate the local decision 
boundary of any black-box classifier. Similarly, SHAP (SHapley Additive exPlanations) \cite{10.5555/3295222.3295230} grounds feature attribution in 
cooperative game theory, computing Shapley values that fairly distribute the prediction output among all input features by averaging their marginal 
contributions across all possible feature subsets — a process inherently based on systematic feature inclusion and exclusion.

Subsequent work has expanded and refined perturbation-based attribution in several directions. KernelSHAP \cite{10.5555/3295222.3295230} and 
SampleSHAP \cite{Aydoan2025AnIA} provide computationally efficient approximations of Shapley values, since its estimation is computationally intractable.
Meaningful Perturbations \cite{8237633} formulates the search for an explanatory mask as a continuous optimization problem, penalizing perturbations 
that minimally alter the input while maximally reducing model confidence. More recently, RISE (Randomized Input Sampling for Explanation) 
\cite{petsiuk2018riserandomizedinputsampling} estimates importance maps by feeding thousands of randomly masked versions of an image through the network 
and computing a weighted average of the masks, scaled by the corresponding output scores. Collectively, however, these methods suffer from severe 
theoretical limitations, particularly in the process of selecting the baseline values that are meant to represent the absence of information. Specifically, 
the Added Bias problem \cite{sturmfels2020visualizing, hsieh2021evaluationsmethodsexplanationrobustness, fel2023dontliemerobust} typically lead to 
information addition with respect to the model's inherent biases and image characteristics, while the Out-Of-Distribution problem \cite{gomez2022metricssaliencymapevaluation} 
highlights that the resulting perturbed samples may lie outside the margins of the training data distribution, rendering the model's responses on such 
inputs unreliable as a basis for attribution.

\textbf{Weight Contribution.} A different line of approaches focuses on the critical role of weights to the model's decision making process.
\textit{Weight perturbation} is based on techniques for altering a model's parameters and is used for improving the model's learning process \cite{9528460}, 
robustness to adversarial examples \cite{NEURIPS2020_1ef91c21} and data reconstruction \cite{9533539}. In the context of interpretability, authors of \cite{olah2020zoom} 
found matching pairs of weights and optimal neuronal activations for interpreting individual neurons. Authors of \cite{NEURIPS2024_72d50a87}
replaced standard linear transformations in deep networks with a new B-cos operation that enforces alignment between model weights and task-relevant input 
patterns during training. This \textit{ante-hoc} approach manages to effectively summarize model behaviour in a single, highly interpretable linear map.
In another line of work, \textit{Embedded Methods} \cite{Importance2024Veas} introduced an artificial layer between the input and the first hidden layer to 
track the evolution of training statistics—most commonly weights and gradients - which are then used to infer feature importance using different techniques 
to the collected gradient or weight profiles. However, the interpretation of these statistics remains elusive.

\section{Methodology}

\begin{figure*}[h]
  \centering
  \includegraphics[width=\linewidth]{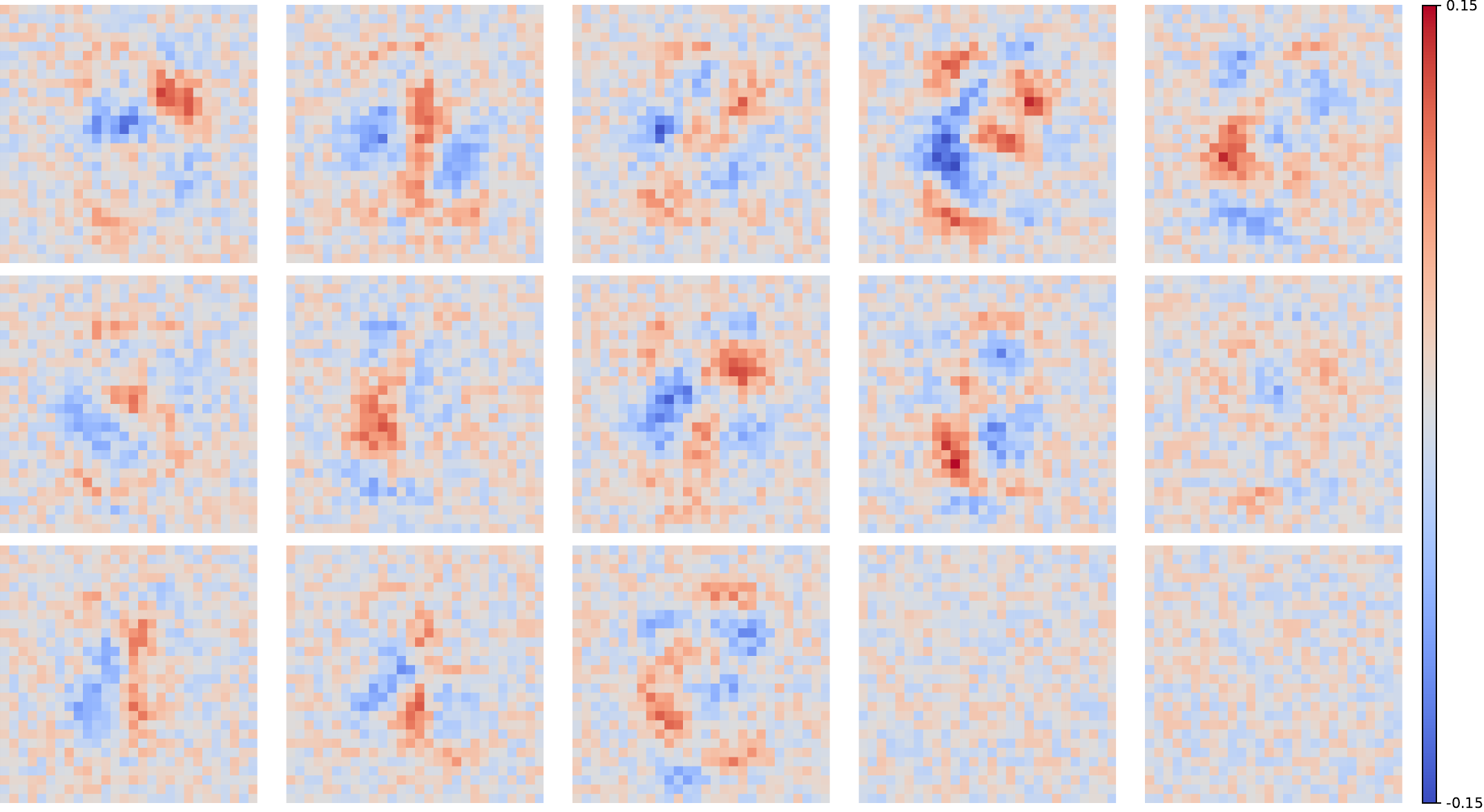}
  \caption{A visualization of first layer weights of a trained FCNN on TMNIST, for specific neurons of the first layer. 
  Distinct patterns corresponding to the digits 0-9 can be identified (first two rows), among other, more abstract patterns (third row). 
  The   case for pattern absence is also present (last two images), signifying neutral neurons with unclear functionality.}
  \Description{}    
  \label{weight_viz}
\end{figure*}

\subsection{Mathematical Annotation} 

Consider an arbitrary fully connected neural network (FCNN) $f = f^L \circ f^{L-1}\circ... \circ f^0$ receiving input $x \in \mathcal{X}$, from the input space. Let  
$l$ denote an arbitrary inner layer of the network, consisting of $n_l$ neurons with weights $W^l \in \mathbb{R}^{n_{l-1}\times n_{l}}$ and bias $b\in \mathbb{R}^l$.
The layer's operation is described by the output $z^l$ obtained by applying a non-linear activation function to the input $x^l_j$. Thus, for neurons $i$ and $j$ in layers 
$l-1$ and $l$, respectively, the following holds:
\begin{align}
    \label{eq:fcnn}
    x^{l}_j &= \sum_{i \in[n_{l-1}]} z_i^{l-1} * w_{ij}^l + b_j^l, \\
    z_j^l &= \sigma(x_j^l).
\end{align}
$\sigma$ is the ReLU activation function, applied to all intermediate layers (except the final layer, where Softmax is used). For a neuron $i$ in the input layer, 
we adopt the annotation $W_i=\{w_{ij} | j \in l^1\}$ denoting the set of its attached weights of the first layer, which will be our primary focus in our research. 

\subsection{Model Parameters and Pattern Formation}

Our methodology treats a deep model as a computational graph—a structured sequence of operations involving neuron activations and model parameters. In this 
framework, parameters act as control mechanisms, shaping the formation of meaningful representations to minimize loss and enable accurate decision-making. Their 
role is thus pivotal for aligning machine and human objectives, comprising the starting point of our research towards interpreting the resulting patterns and 
representations.

In the context of FCNNs, the visualization of weights connecting input features to specific neurons of the first layer 
$W^T_j = \{w_{ij} | i in [n^0]\}$ reveals patterns that are interpretable (as demonstrated in Figure~\ref{weight_viz})) and can be well understood: they illustrate 
the model's motifs that either enhance or suppress a neuron's activation. Such dual behavior reflects the polysemantic nature of neurons \cite{olah2020zoom}, 
where a single unit integrates multiple feature detectors. In the first layer of FCNNs, this functionality manifests as strong activation in response to specific 
patterns and suppression in response to others. 

The primary contribution of visualizing weights $W^T_j$ is the revelation of spatial structure among input features. This is a key observation which will be exploited 
in our work. However, the individual interpretation of a 
weight does not directly translate to estimating the contribution of input neurons, which collectively influence all patterns encoded in the weights. In this regard,
authors in \cite{olah2020zoom} advanced our understanding by constructing optimal input signals: maximizing activation for specific inner neurons and matching their 
patterns with those of the neuron's weights. Although this approach enables interpretation of inner neurons and introduces a novel framework for 
integrating neuronal interactions with weight patterns, it does not generalize to a quantitative measure of neuronal contribution.

\subsection{Introducing Explainable Weight Perturbation}
We introduce a novel attribution method, bridging the gap between pattern formation and neuronal contribution. This is achieved through a theoretically grounded 
framework rooted in the principles of Occlusion. We term our method \textbf{Explainable Weight Perturbation} (XWP), operating by assessing neuronal contribution
as a distance measure of the model's response at the transition of values $W_i$ from the initial to the trained model state.

Formally, we denote $f(x)$ as the output of the trained model, and $f_{\text{ }W_i = W^o_{i}}(x)$ being the targeted replacement of parameters $W_i$ of the 
trained model by those of the untrained state—denoted as $W^o_{i}$. XWP applies the following rule:
\begin{equation}
    \label{eq:xwp}
    R_i(x) = \mathcal{I}^t*(f(x) - f_{\text{ }W_i = W^o_{i}}(x)),
\end{equation}
where $\mathcal{I}$ equals one when output neuron $n^l$ matches the target $t$:
\begin{equation}
    \mathcal{I}^t_{n^L} = 
    \begin{cases}
    -1, & \text{if } n^L \neq t \\
    \text{ }\text{ }1, & \text{if } n^L = t.
    \end{cases}
\end{equation}
It thus functions as a reward signal, by positively awarding increases in target neuron prediction (or decreases in non-target neurons) in this weight 
transition, signaling a favourable change of $i$'s contribution to the model's prediction capacity. Any other case meets its unfavorable response. 

Unlike traditional Occlusion approaches, our method follows a distinct path, with perturbations unfolding in the parameter instead of the input space. It 
leverages the model's initial state, considering it a blank spot, where mechanisms detecting patterns are in their early stage. The application of the distance 
measure permits the model's pattern formation to be expressed—through its effect on the input samples—while guiding the selection of the baseline value by the 
model's state. This approach introduces no human bias or ambiguous statistical rules, enabling the safe and reliable application of the Occlusion rule.

In equation~\ref{eq:xwp} the model's inner parameters are selected from the trained state, while in constrast to values of $W_i$, their values remain unaltered.
This static consideration of intermediate parameters stems from the assumption that their effect is equally shared among input neurons. This in turn permitted the 
consideration of input and inner representations as being decoupled, and the application of changes only to the first—further enabling the exploitation of the 
input's spatial structure. Despite this, this independence is deceptive, since the training algorithm considers their effect when applying the chain rule to 
weights $W_i$, rendering our method accountable for internal neuronal interactions.

Having defined a method for identifying and isolating the contribution of feature $i$ to pattern formation, we can express its complementary rule
XWP\textsuperscript{c}, as follows:
\begin{equation}
    R^{\text{ }c}_i(x) = \mathcal{I}^t*(f(x) - f_{\text{ }W_{i'} = W^o_{i'}}(x)), 
\end{equation}
where $i' \neq i$ represents all neurons of the input layer differing from $i$.

To summarize the role of each rule, they jointly perturb weights in the direction indicated by the model's trained state, to derive a feature importance 
score based on the model's response. XWP achieves this for feature $i$ by measuring its effect when excluding it from the trained model, while 
XWP\textsuperscript{c} assesses this effect by dropping the values of all other features back to their initial states. Expressed in simpler terms, a feature in XWP 
asks itself \textit{"what did I achieve for the model's capabilities considering all updates"} while in XWP\textsuperscript{c} the underlying question is 
\textit{"what did I achieve only by myself"}.

We move forward to testing our proposed variants for two baseline datasets in Section~\ref{sec:exper}, while identifying key characteristics in their produced 
heatmaps and comparing them with state-of-the-art approaches.
\section{Experiments}
\label{sec:exper}

To evaluate the effectiveness of our framework, we conduct a comprehensive experimental analysis of its two variants, applied to FCNNs trained on the Typeface MNIST 
and Fashion MNIST datasets. We employ a standard FCNN architecture with ReLU activations in the intermediate layers and a softmax output layer. The model consists 
of four layers with dimensions (784, 400, 100, 10), respectively. Training continues until convergence, typically achieving test accuracy exceeding 90\% within 
10 epochs. All experiments are performed on an NVIDIA GeForce RTX 4060 GPU.

We then compute the attribution scores generated by our proposed variants and compare them against those produced by five state-of-the-art baseline methods:
Occlusion (OCCL) \cite{10.1007/978-3-319-10590-1_53}, Shapley values (SHAP) \cite{10.5555/3295222.3295230}, RISE \cite{petsiuk2018riserandomizedinputsampling}, Integrated Gradients \cite{sundararajan2017axiomaticattributiondeepnetworks} and LRP \cite{binder2016layerwiserelevancepropagationneural}. 
Our evaluation combines quantitative metrics with qualitative analysis, including the visualization of computed heatmaps. The detailed results are presented 
in Section~\ref{sec:test}.

\subsection{Datasets}
\textbf{TMNIST} (Typeface-MNIST). This dataset \cite{tmnist_vyawahare} is composed of 29,900 MNIST-style grayscale 28x28 images of digits from 0 to 9, rendered across a variety of Google font 
styles. Unlike handwritten digit datasets, the typeface-based nature of TMNIST produces clean, well-defined digit shapes with minimal noise and ambiguity. 
This property makes it particularly well-suited for the evaluation of attribution methods, as the clarity of the digit structures allows for a more 
unambiguous interpretation of the resulting attribution maps.

\textbf{FashionMNIST}. This is a dataset of grayscale images \cite{xiao2017/online} introduced as a more challenging alternative to the original MNIST benchmark. 
It consists of 70,000 28x28 images spanning ten classes of clothing items, including t-shirts, trousers, and ankle boots, split into 60,000 training and 10,000 
test samples. Despite its simplicity in resolution, the dataset presents non-trivial classification challenges due to inter-class visual similarity, making it a
standard benchmark for evaluating models on simple image recognition tasks.

\subsection{Evaluation Metrics}

\begin{figure*}[t]
    \centering
    \setlength{\tabcolsep}{2pt}  

    \begin{tabular}{c c c c c c c c}
        \textbf{Sample} & \textbf{OCCL} & \textbf{SHAP} & \textbf{RISE} & \textbf{IG} &
        \textbf{LRP} & \textbf{XWP} & \textbf{XWP\textsuperscript{c}} \\[3pt]

        \includegraphics[width=0.115\textwidth]{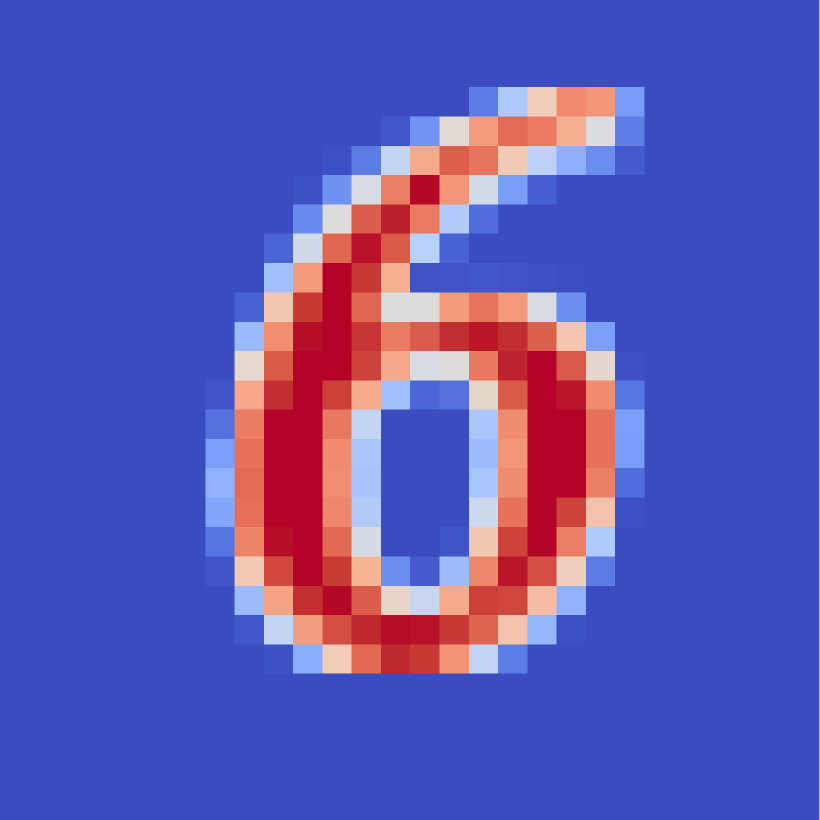} &
        \includegraphics[width=0.115\textwidth]{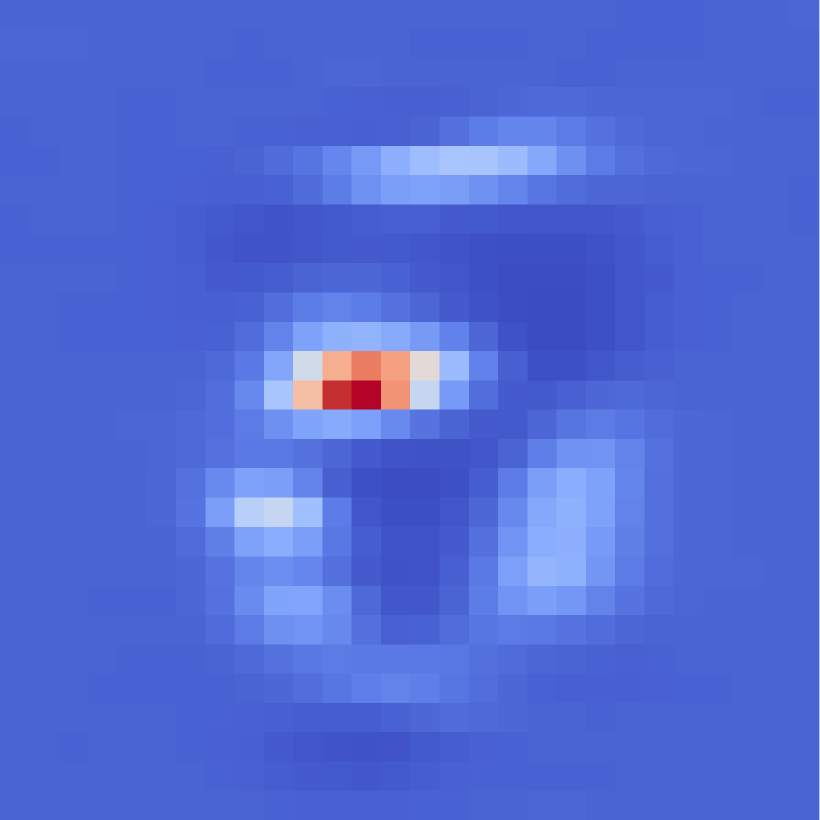} &
        \includegraphics[width=0.115\textwidth]{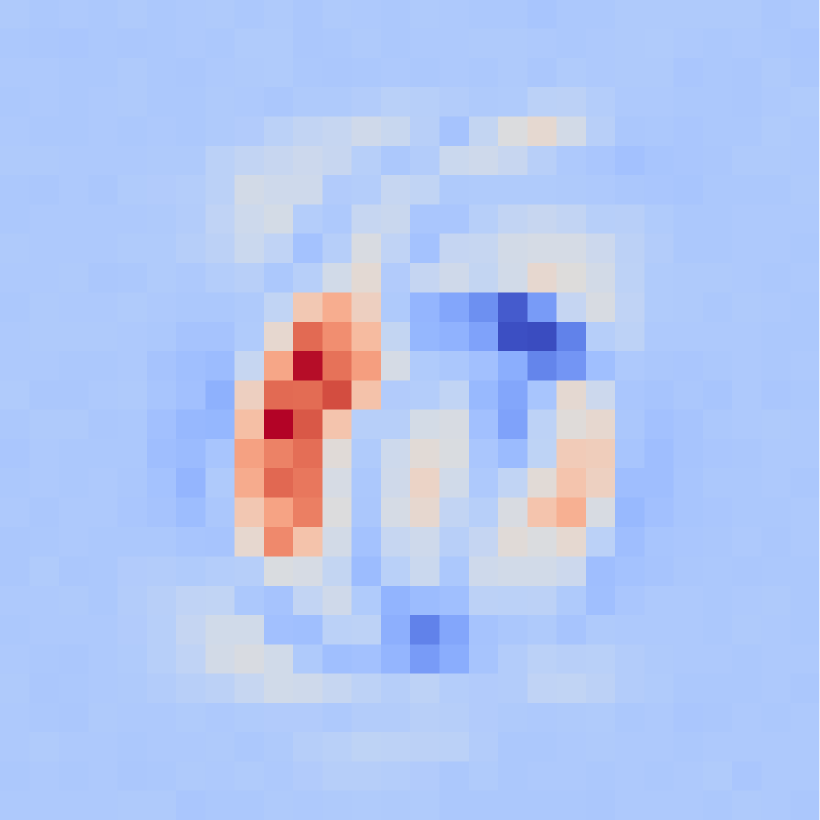} &
        \includegraphics[width=0.115\textwidth]{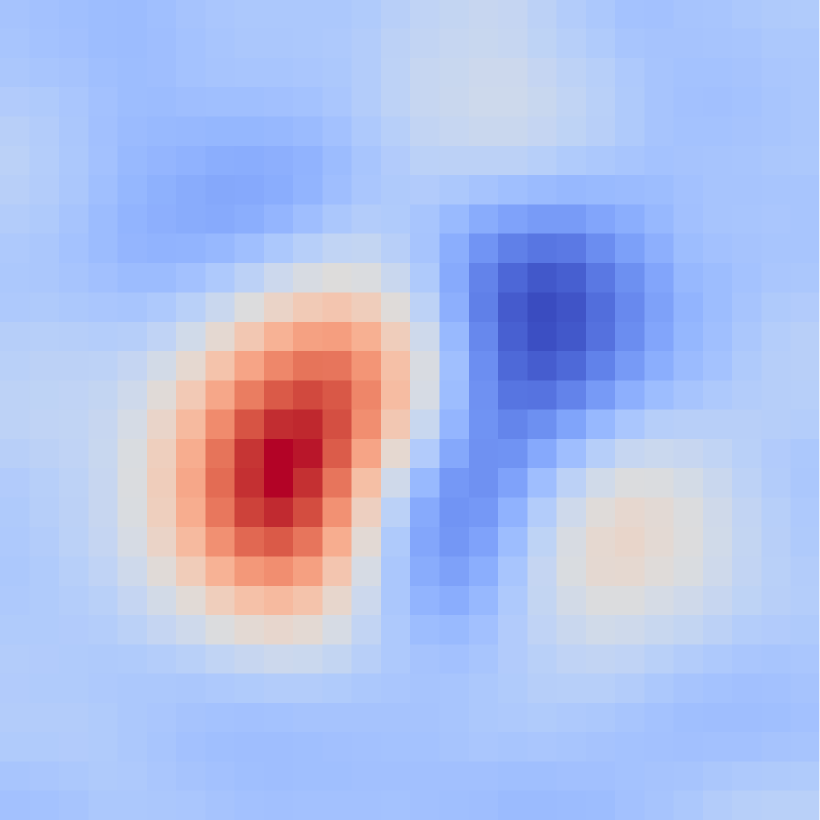} &
        \includegraphics[width=0.115\textwidth]{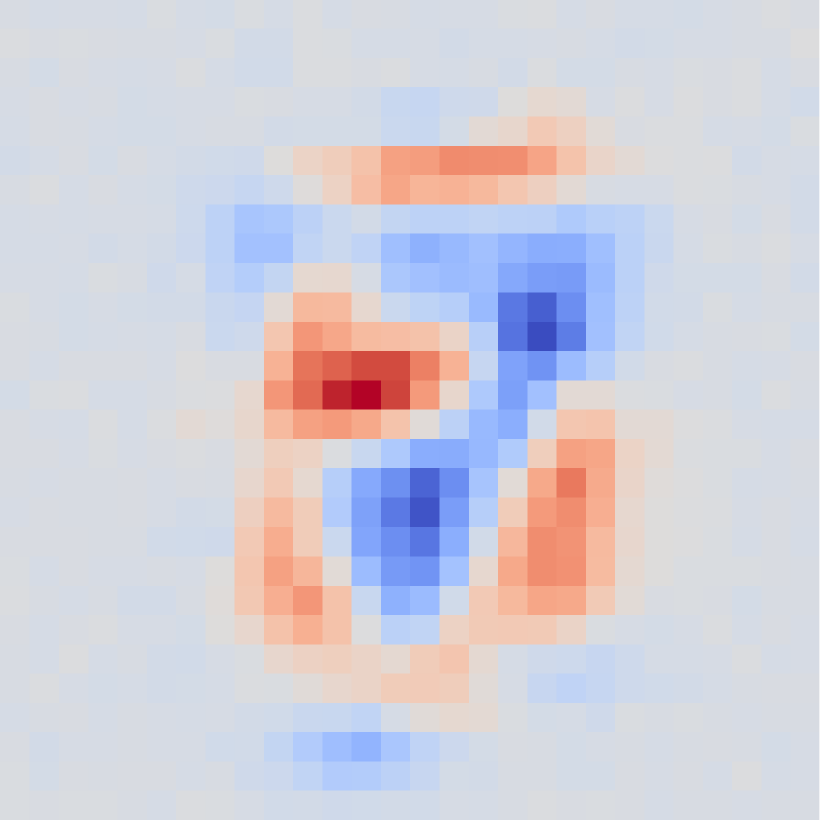} &
        \includegraphics[width=0.115\textwidth]{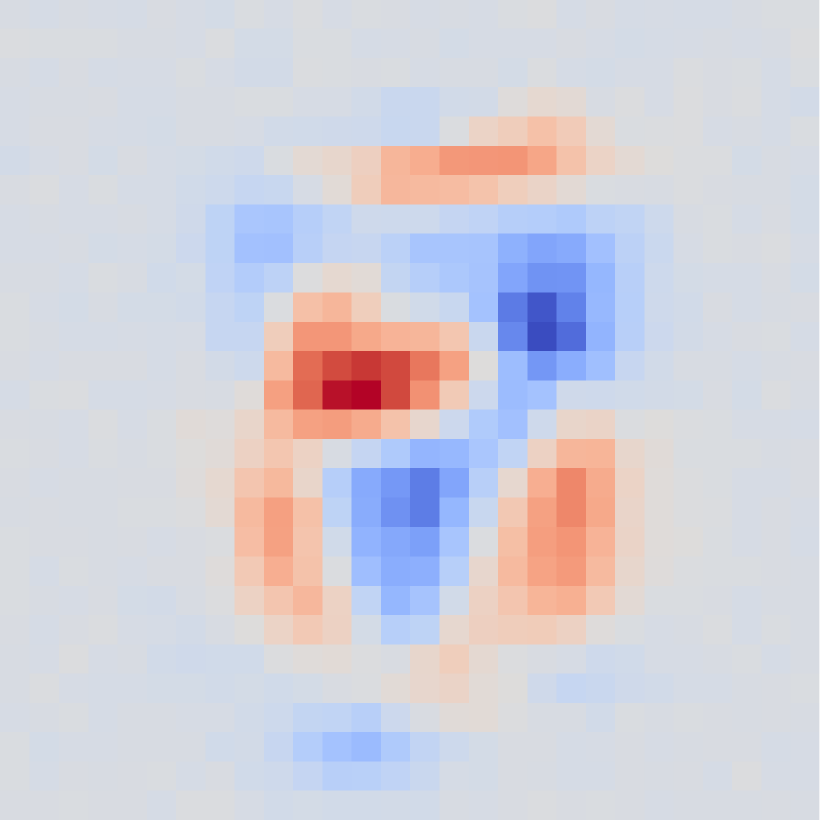} &
        \includegraphics[width=0.115\textwidth]{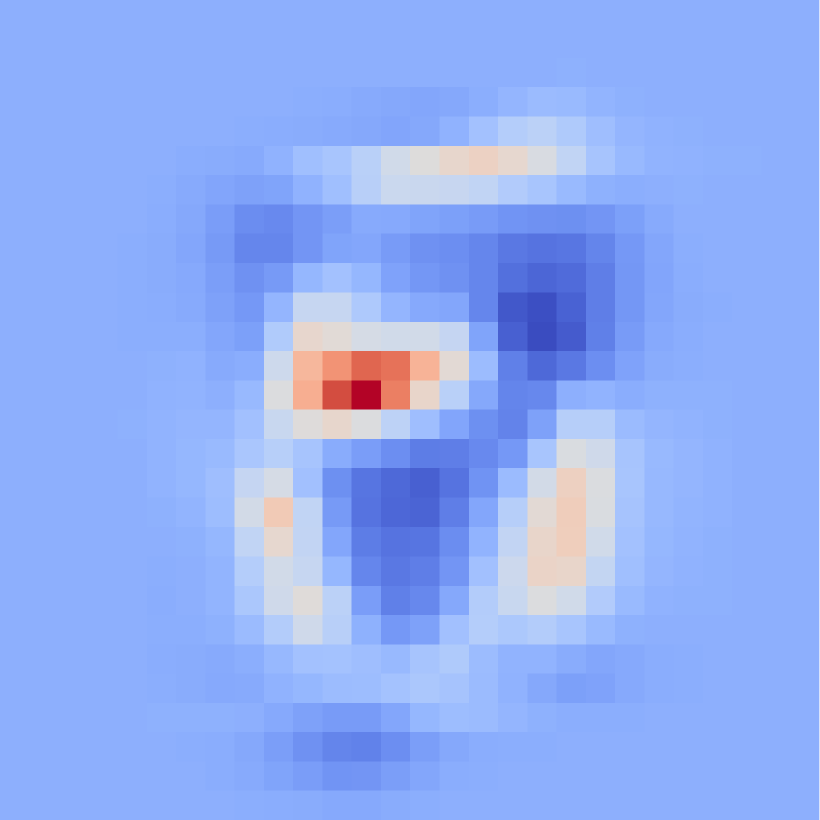} &
        \includegraphics[width=0.115\textwidth]{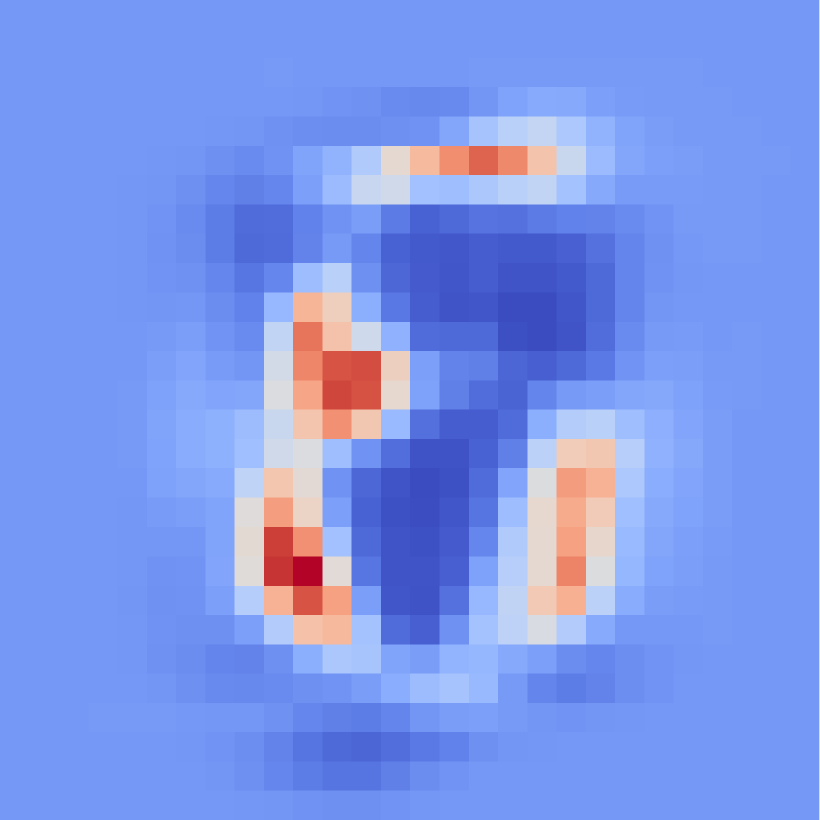} \\[3pt]

        \includegraphics[width=0.115\textwidth]{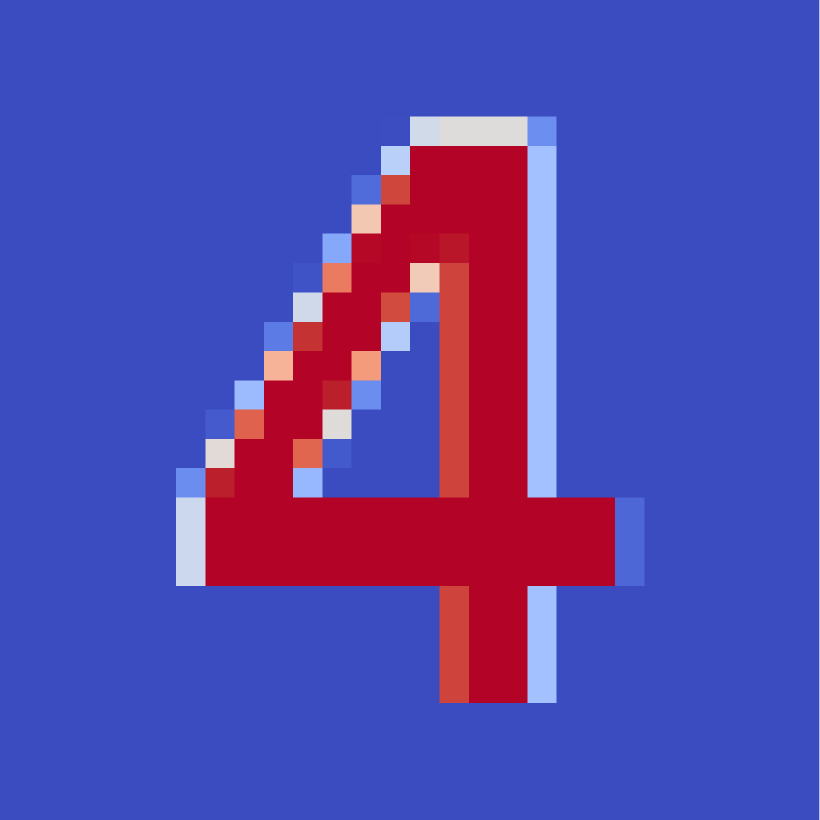} &
        \includegraphics[width=0.115\textwidth]{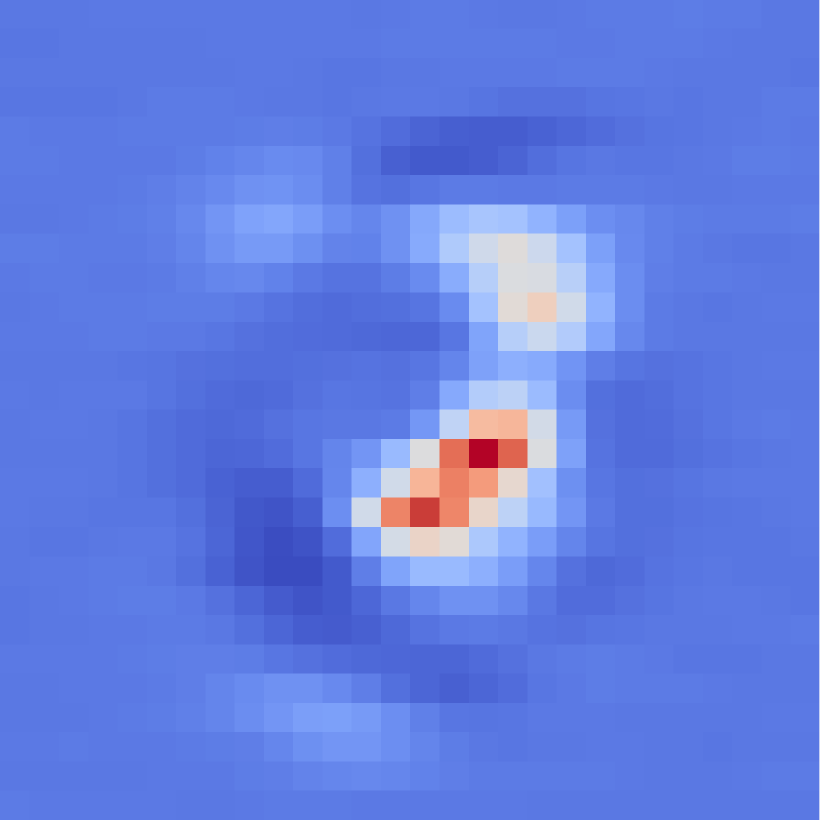} &
        \includegraphics[width=0.115\textwidth]{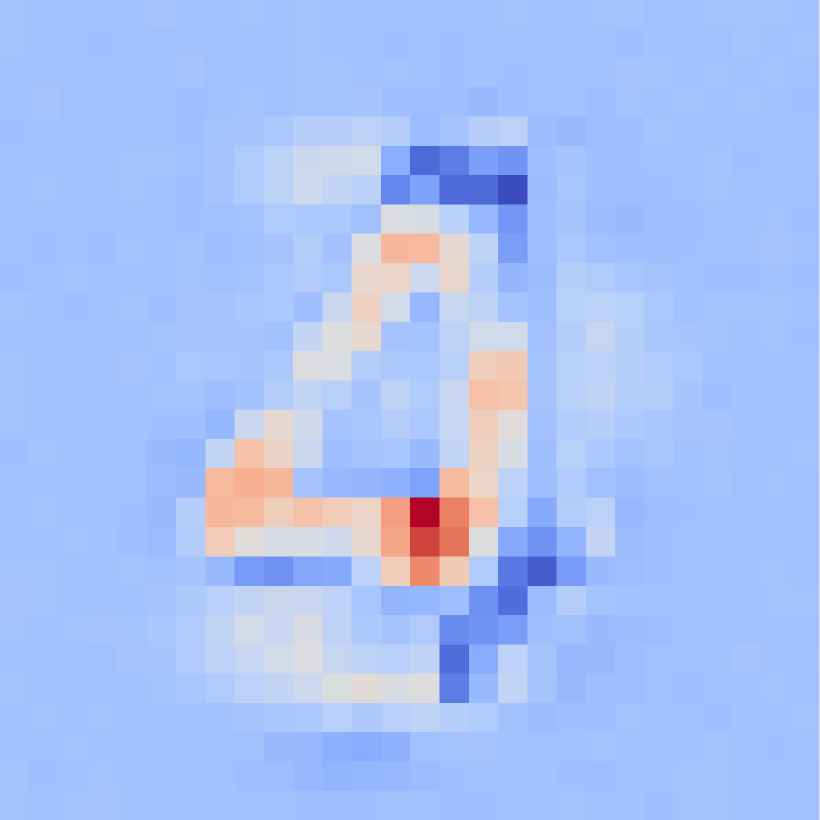} &
        \includegraphics[width=0.115\textwidth]{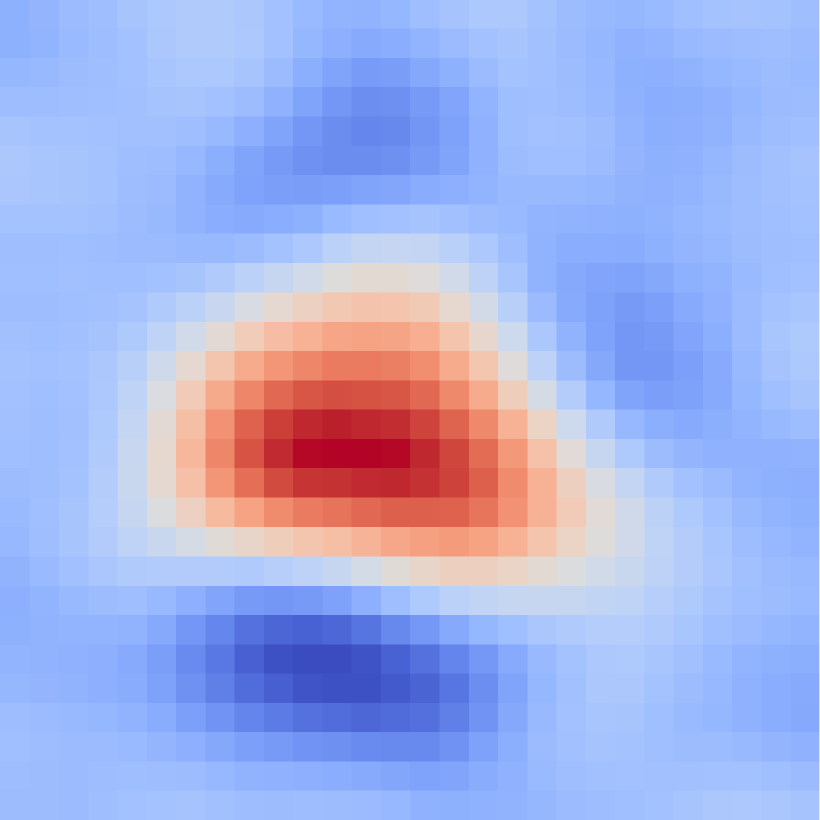} &
        \includegraphics[width=0.115\textwidth]{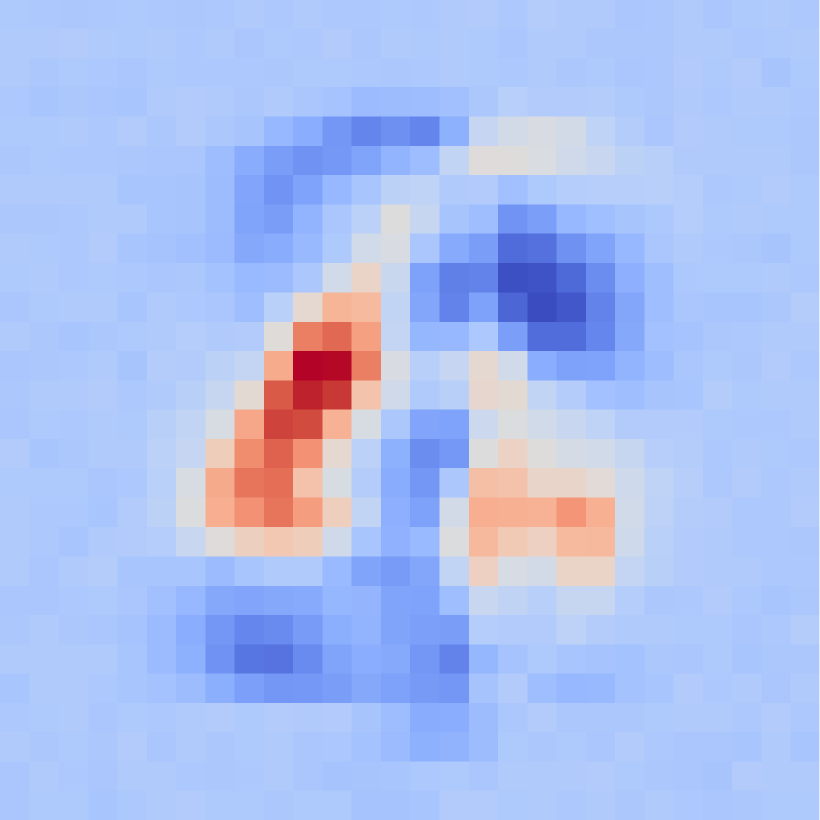} &
        \includegraphics[width=0.115\textwidth]{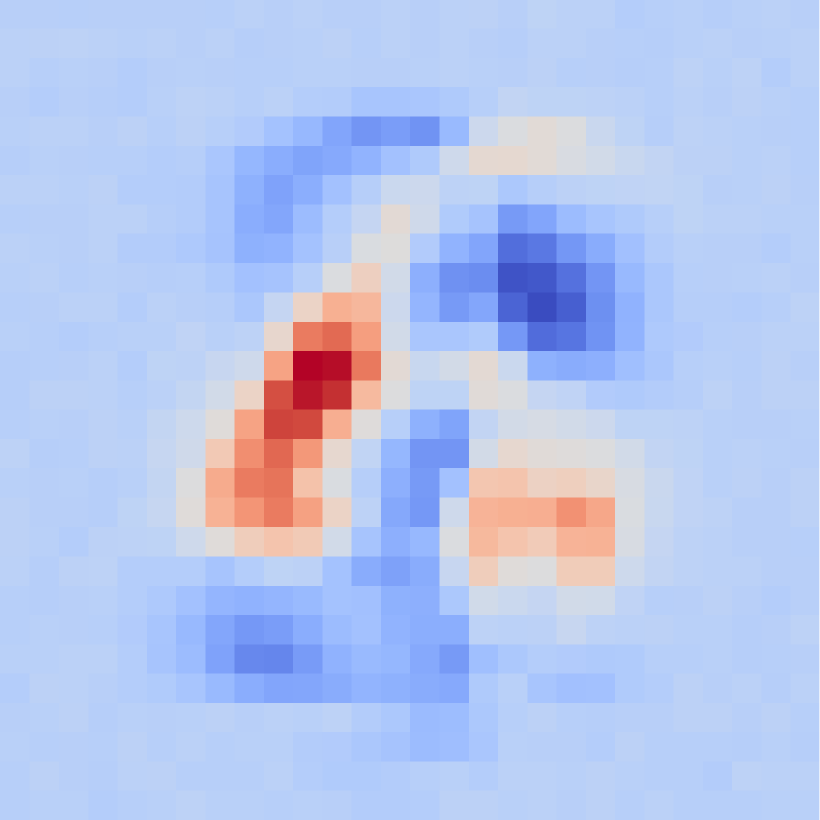} &
        \includegraphics[width=0.115\textwidth]{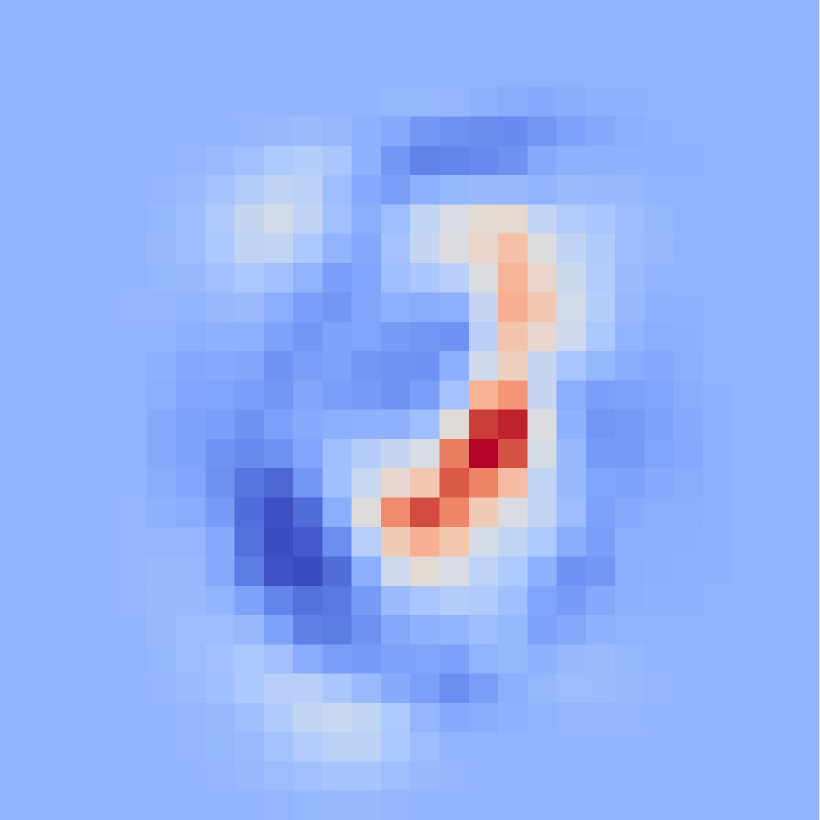} &
        \includegraphics[width=0.115\textwidth]{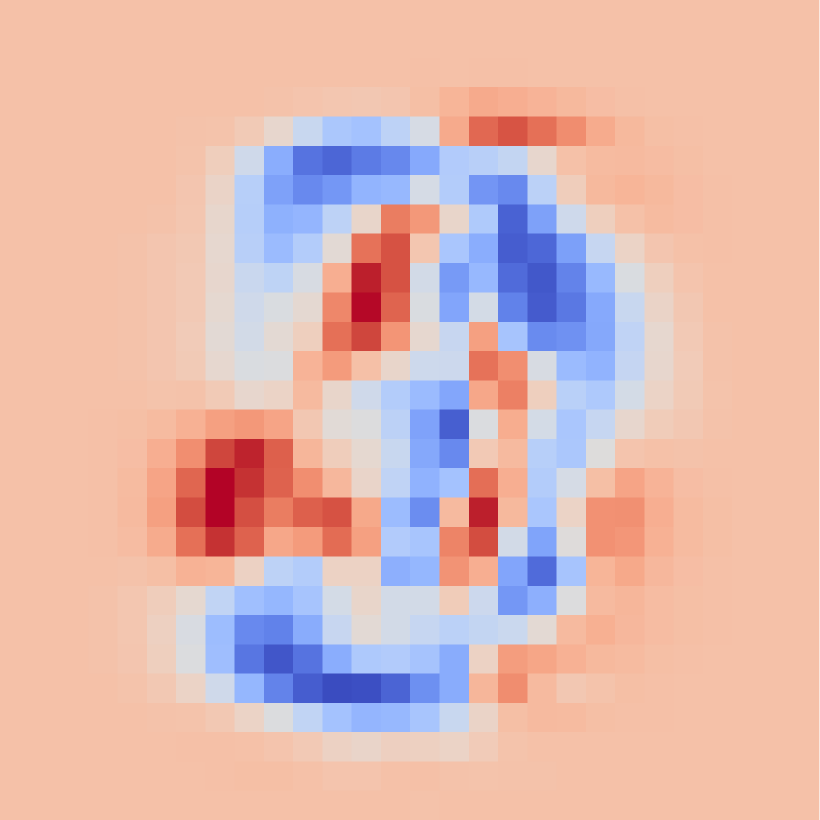} \\[3pt]

        \includegraphics[width=0.115\textwidth]{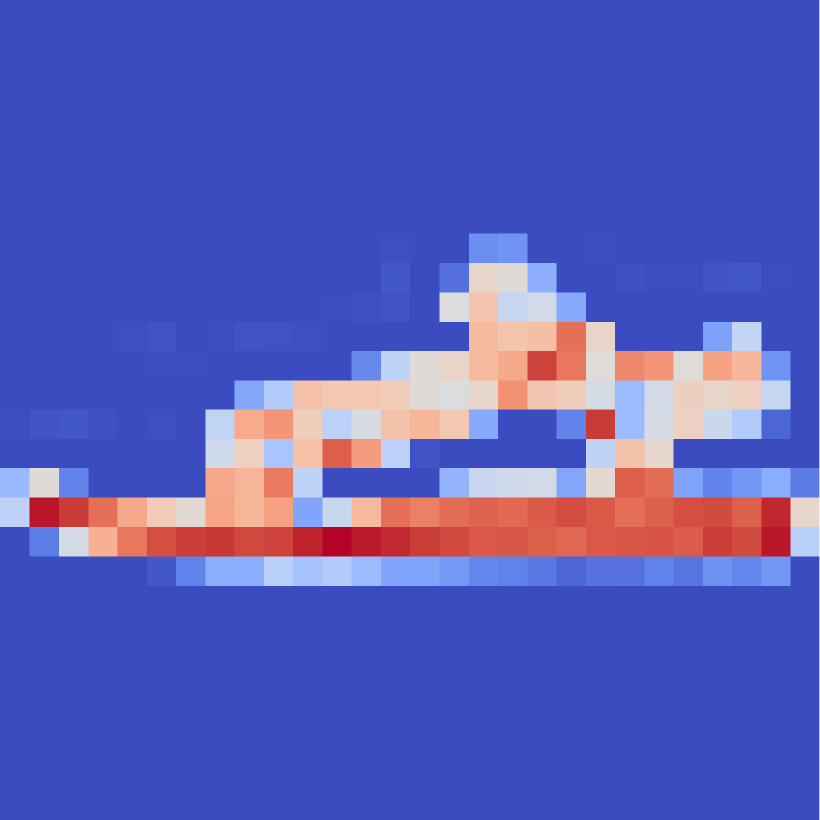} &
        \includegraphics[width=0.115\textwidth]{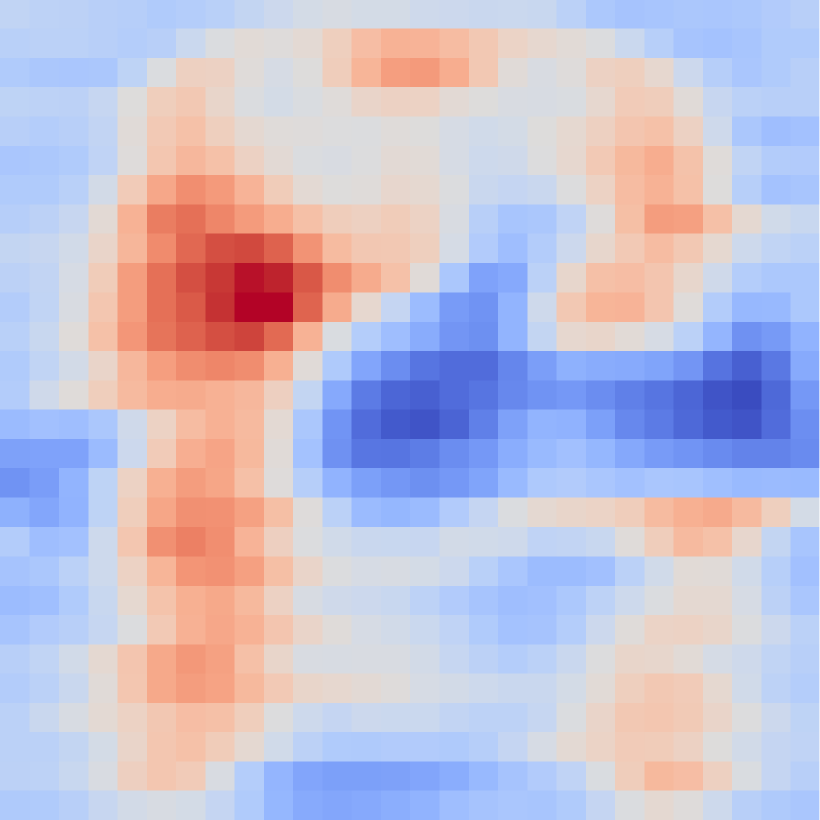} &
        \includegraphics[width=0.115\textwidth]{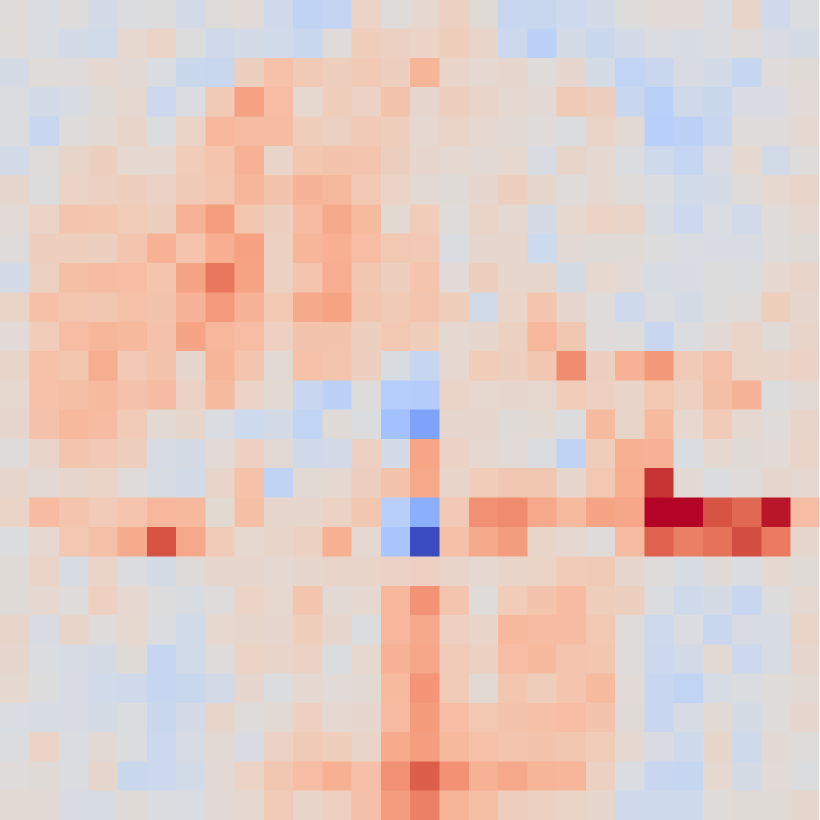} &
        \includegraphics[width=0.115\textwidth]{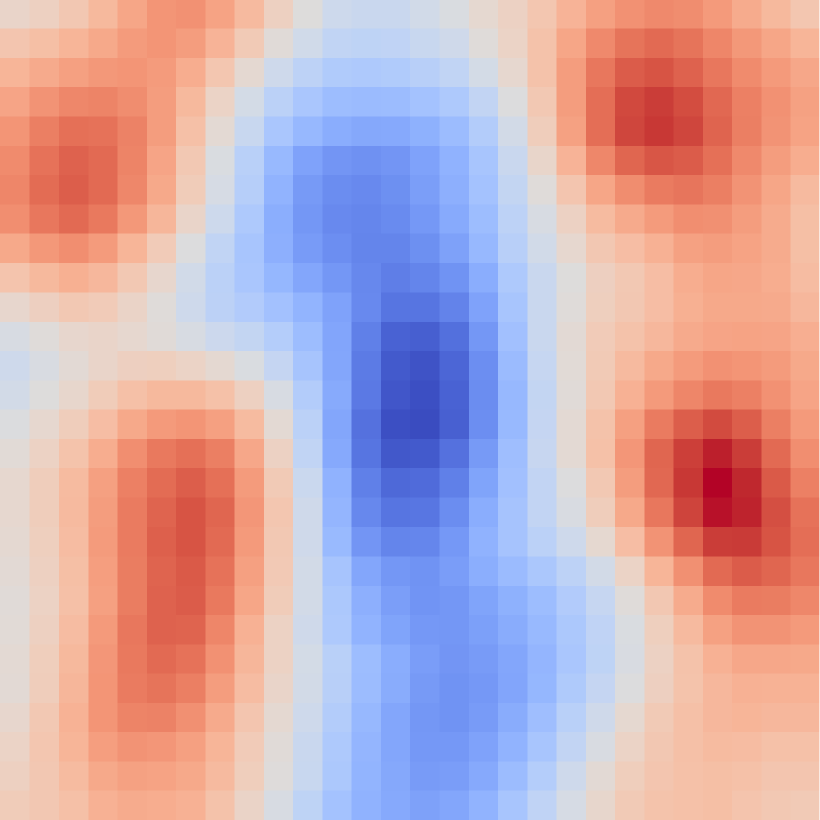} &
        \includegraphics[width=0.115\textwidth]{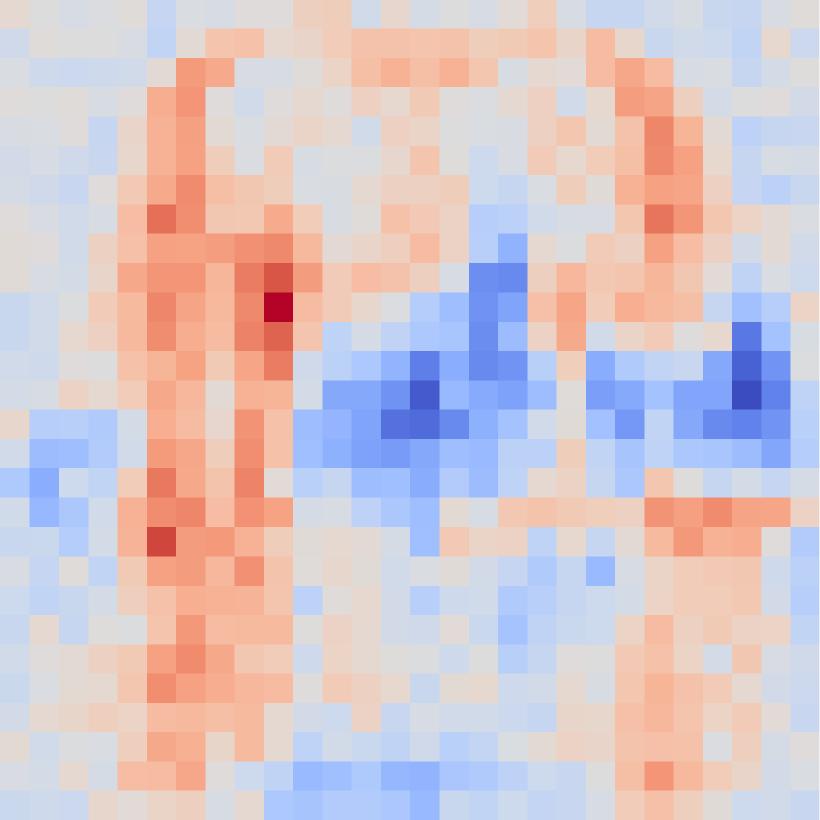} &
        \includegraphics[width=0.115\textwidth]{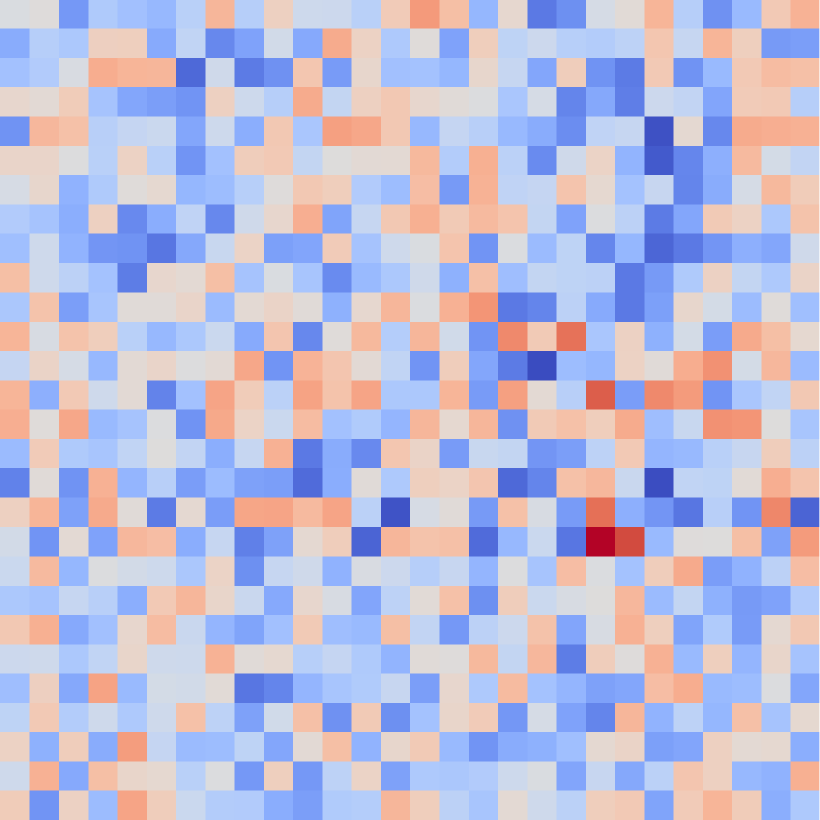} &
        \includegraphics[width=0.115\textwidth]{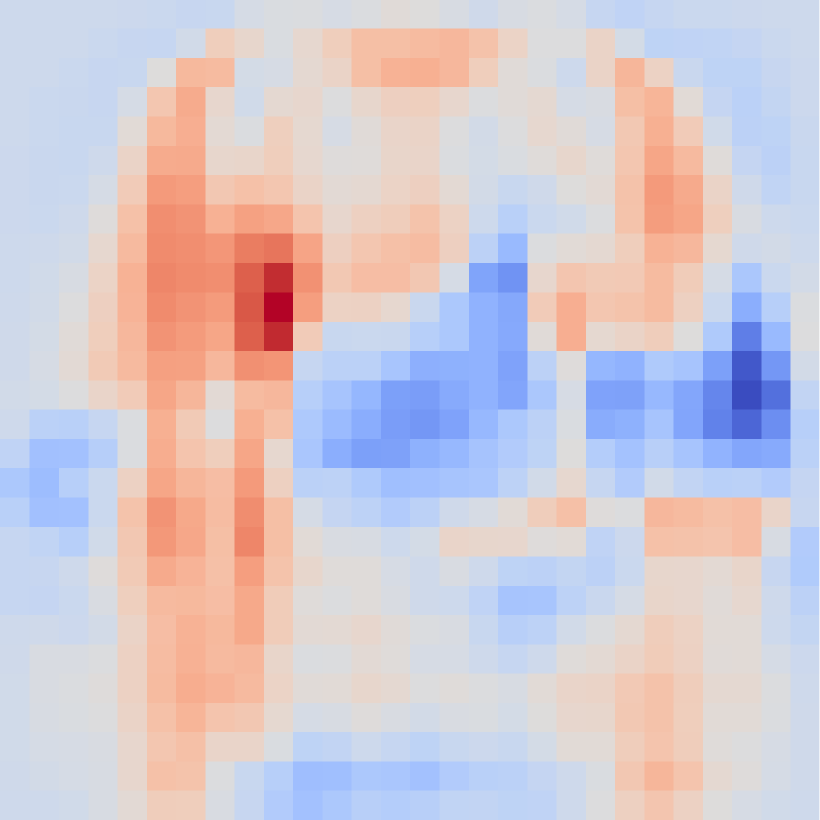} &
        \includegraphics[width=0.115\textwidth]{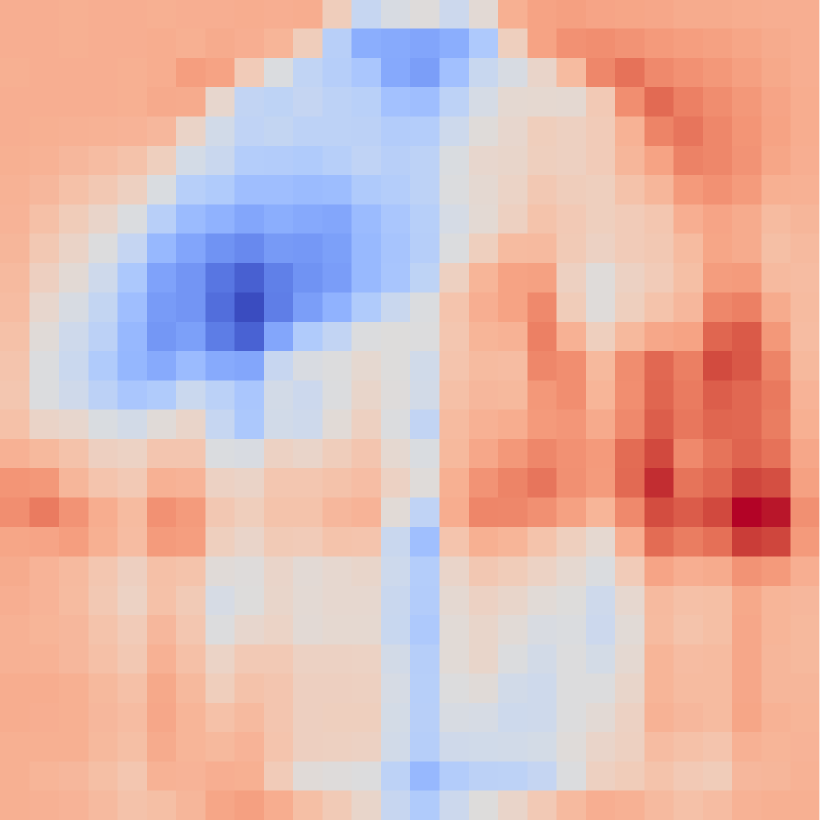} \\[3pt]

        \includegraphics[width=0.115\textwidth]{Experiments/fashion_mnist/sample_11.pdf} &
        \includegraphics[width=0.115\textwidth]{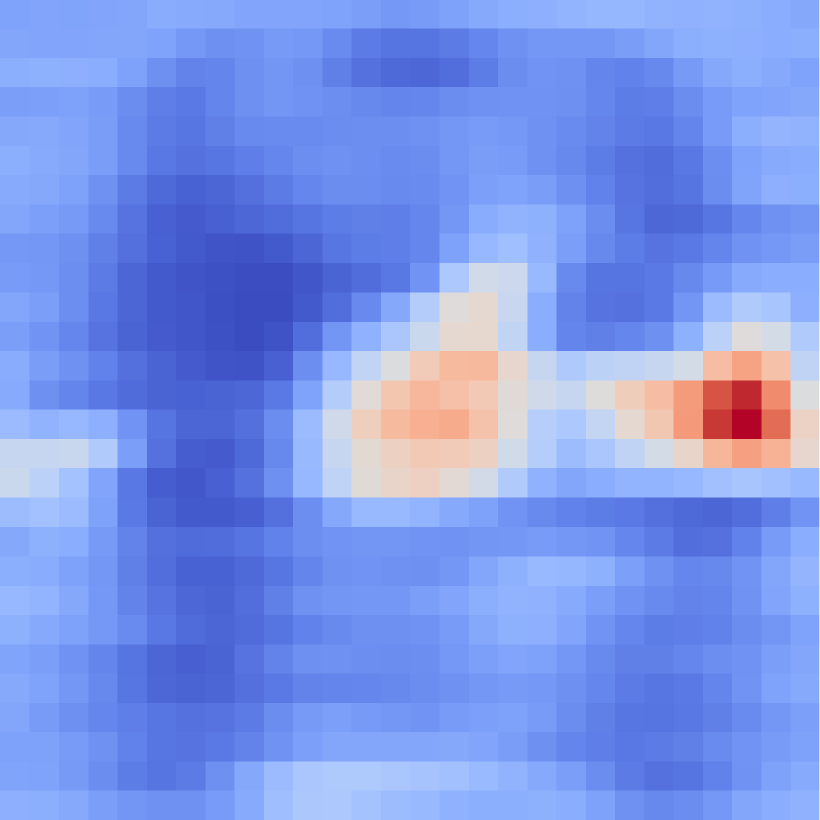} &
        \includegraphics[width=0.115\textwidth]{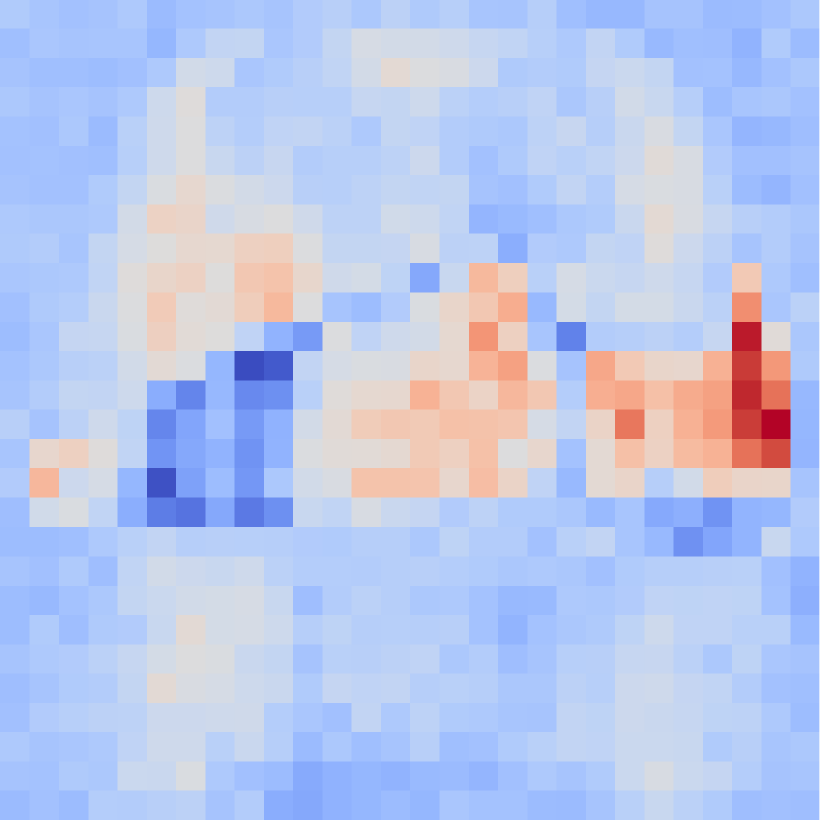} &
        \includegraphics[width=0.115\textwidth]{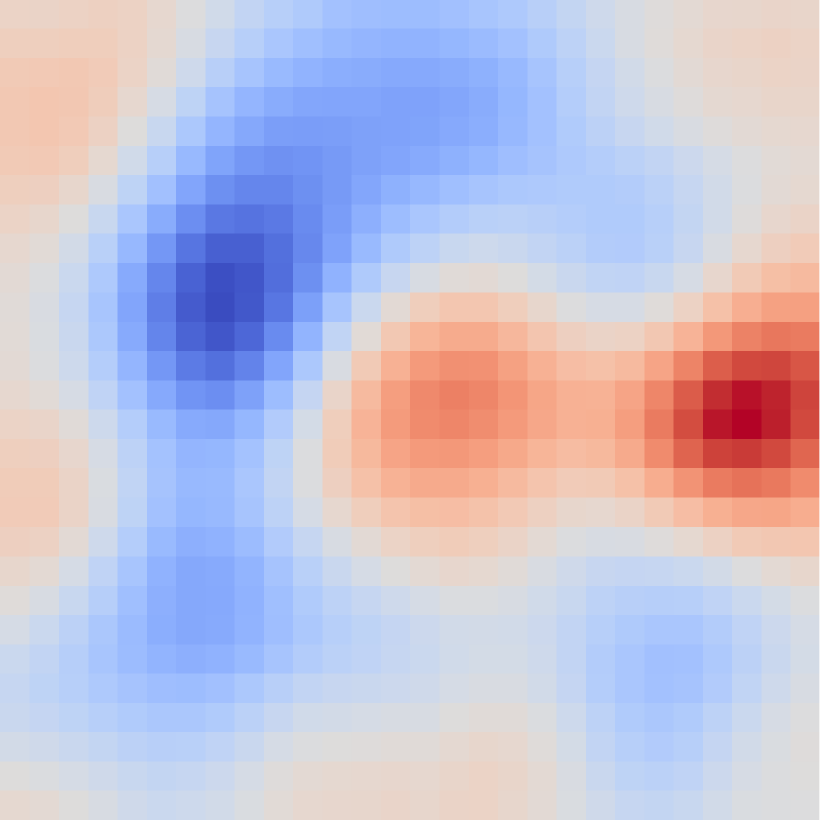} &
        \includegraphics[width=0.115\textwidth]{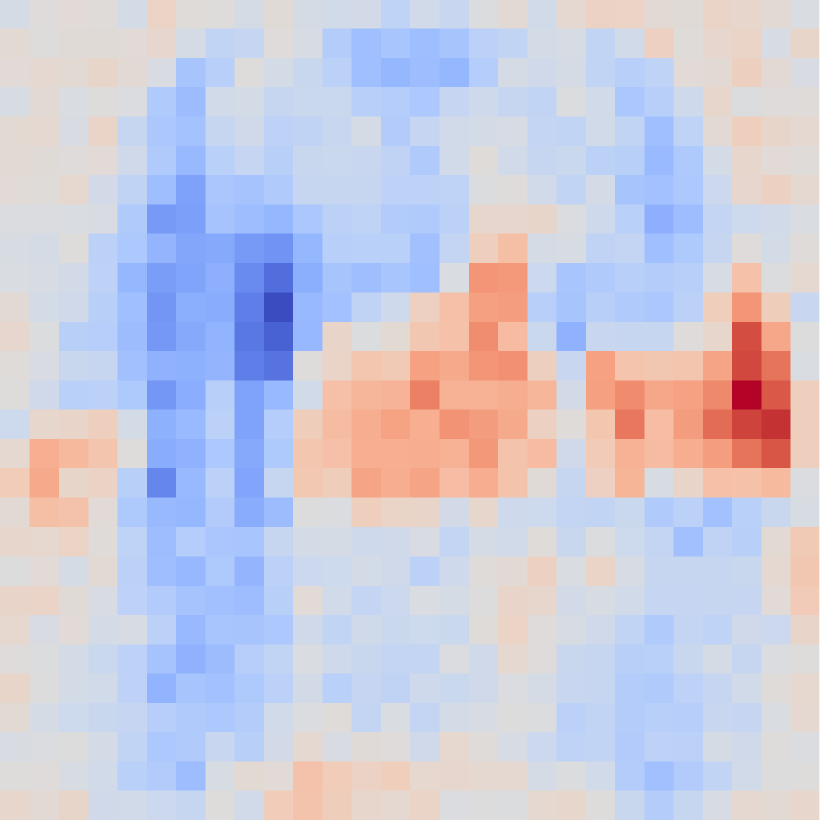} &
        \includegraphics[width=0.115\textwidth]{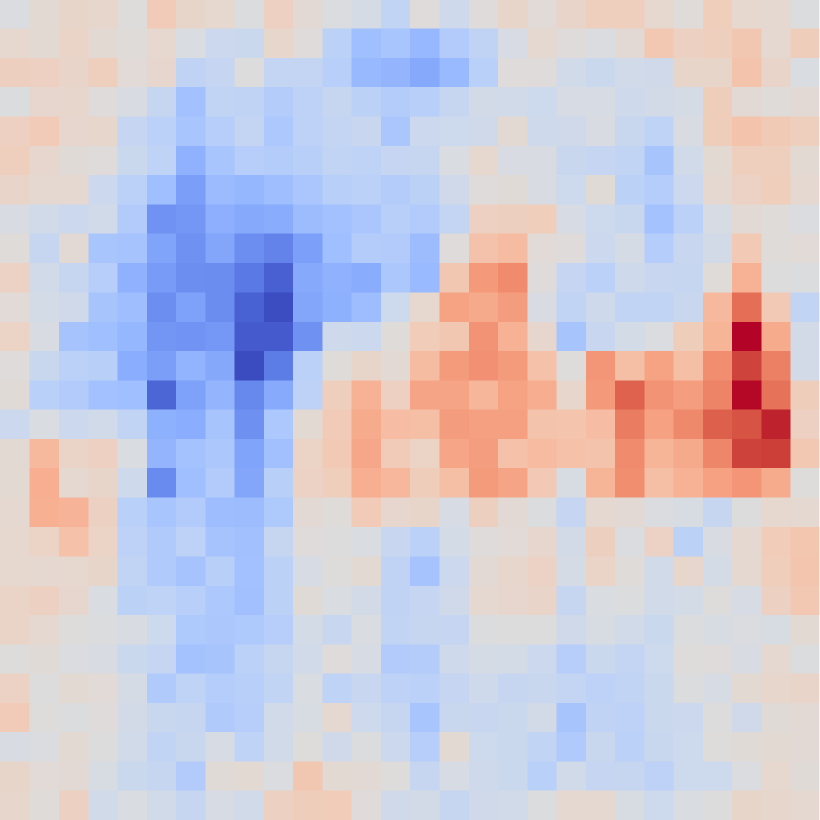} &
        \includegraphics[width=0.115\textwidth]{Experiments/fashion_mnist/XWP_11.pdf} &
        \includegraphics[width=0.115\textwidth]{Experiments/fashion_mnist/XWP_C_11.pdf} \\[3pt]
       
    \end{tabular}

    \caption{The importance scores of different attribution methods for the Typeface MNIST and Fashion MNIST datasets. A visual inspection suggests that 
    XWP\textsuperscript{c} is slightly more capable than XWP in detecting patterns within the input signal, while they jointly produce the cleanest heatmaps
    with respect to their competitors.}
    \label{fig:both_res}

    \Description{A grid of 4x8 importance scores of different attribution methods for the Typeface MNIST and Fashion MNIST datasets (two for each). Starting
    from the leftmost column the image sample is presented, followed by different attribution methods: Occlusion, Shapley Values, Integrated Gradients, LRP, XWP and 
    XWP\textsuperscript{c}. The heatmaps of the later seem slightly more accurate than these of XWP in detecting patterns within the input signal, while they 
    jointly produce the cleanest heatmaps with respect to other techniques.}
\end{figure*}

\textbf{Deletion AUC}. The Deletion AUC criterion \cite{petsiuk2018riserandomizedinputsampling} evaluates an attribution method by progressively removing input 
features in decreasing order of their assigned importance scores and measuring the resulting drop in model confidence. The area under the resulting confidence 
curve is then computed — a lower value indicates a more faithful attribution map, as removing the most important features should cause a sharp and sustained 
drop in confidence. This metric directly tests whether the features identified as important are indeed the ones the model relies upon.

\textbf{Average Drop}. The Average Drop metric \cite{8354201} measures the mean percentage decrease in model confidence when a percentage of the most important
image regions highlighted by the attribution map are dropped. Formally, it computes the relative drop between the model's confidence on the original image and its 
confidence on the masked image, averaged across the dataset. A higher average drop indicates a more accurate attribution method, as the highlighted regions should 
hide crucial information for the model's prediction. For our experimental setup, we selected the value of 20\% as the masking percentage. 

\subsection{Results}
\label{sec:test}

\begin{table*}[h!]
    \caption{Quantitative analysis, based on Average Drop (AD) and Deletion AUC (AUC) for the datasets TMNIST and FashionMNIST (FTMNIST). 
    Shapley values and Integrated Gradients outperform all other techniques by a small margin, yet XWP and XWP\textsuperscript{c} closely match their performance, 
    achieving greater stability. Best results are highlighted in bold.}
    \label{tab:main_results}
    \begin{center}
    \begin{footnotesize}
    \begin{sc}
    \setlength{\tabcolsep}{3.5pt}
    \begin{tabular*}{\textwidth}{@{\extracolsep{\fill}}lcccccccccccccc}
    \toprule
    & \multicolumn{2}{c}{OCCL} & \multicolumn{2}{c}{SHAP} & \multicolumn{2}{c}{RISE} & \multicolumn{2}{c}{IG} & \multicolumn{2}{c}{LRP} & \multicolumn{2}{c}{XWP} & \multicolumn{2}{c}{XWP\textsuperscript{c}}\\
    \cmidrule(lr){2-3} \cmidrule(lr){4-5} \cmidrule(lr){6-7} \cmidrule(lr){8-9} \cmidrule(lr){10-11} \cmidrule(lr){12-13} \cmidrule(lr){14-15}
    Dataset & AD($\uparrow$) & AUC($\downarrow$) & AD($\uparrow$) & AUC($\downarrow$) & AD($\uparrow$) & AUC($\downarrow$) & AD($\uparrow$) & AUC($\downarrow$) & AD($\uparrow$) & AUC($\downarrow$) & AD($\uparrow$) & AUC($\downarrow$) & AD($\uparrow$) & AUC($\downarrow$) \\
    \midrule
    TMNIST   & 0.327 & 0.126 & \textbf{0.357} & \textbf{0.095} & 0.33 & 0.17 & 0.35 & 0.157 & 0.285 & 0.203 & 0.308 & 0.138 & 0.352 & 0.162 \\
    FMNIST   & 0.636 & 0.489 & 0.633 & 0.491 & 0.622 & 0.564 & \textbf{0.661} & \textbf{0.461}  & 0.645 & 0.551 & 0.647 & 0.473 & 0.655 & 0.524\\
    \bottomrule
    \end{tabular*}
    \end{sc}
    \end{footnotesize}
    \end{center}
\end{table*}

\begin{figure*}[t]
    \centering
    
    \includegraphics[width=0.84\textwidth]{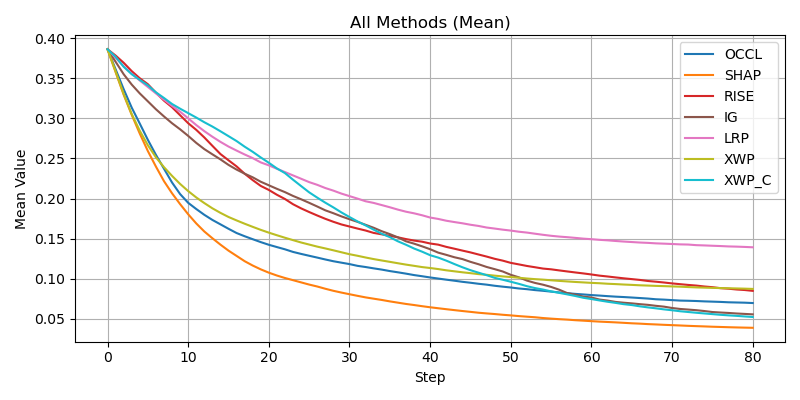} \\
    \includegraphics[width=0.84\textwidth]{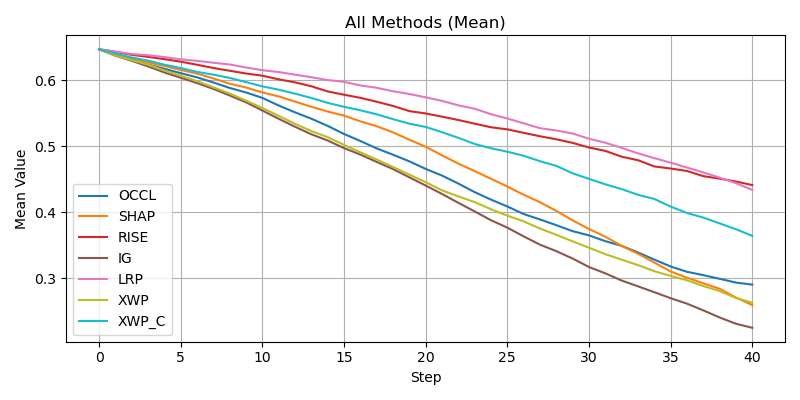} 
       
    \caption{The evolution of Deletion AUC for different attribution methods. Shapley values and Integrated Gradients experience the sharpest drop in AUC score
    in TMNIST and FashionMNIST respectively, 
    with XWP\textsuperscript{c} (denoted as XWP\_C) and Occlusion to follow. The number of steps is chosen at the point just before the curves reach a plateau.}
    \label{fig:auc_scores}
    \Description{The evolution of Deletion AUC for different attribution methods. Shapley values and Integrated Gradients experience the sharpest drop in AUC score
    in TMNIST and FashionMNIST respectively, with XWP\textsuperscript{c} (denoted as XWP\_C) and Occlusion to follow.}
\end{figure*}

This section presents the results of the experiments described in Section~\ref{sec:exper}. As shown in Table~\ref{tab:main_results}, Shapley values achieve 
the highest performance in TMNIST, while Integrated Gradients excel in FashionMNIST. XWP demonstrates strong performance in Deletion AUC (Figure~\ref{fig:auc_scores}), 
while XWP\textsuperscript{c} excels in Average Drop.

A vizualisation of the generated heatmaps (Figure~\ref{fig:both_res}) further clarifies the nature of the computed attribution scores. While the results generally hold, 
we observe cases where the individual effect of feature change does not match the signal itself, exhibiting in many cases strong similarities to Occlusion. 
This translates to a divergence of the objectives of XWP—targeting the model's update needs for better generalization—with the model's current dynamics in response
to the input signal. This gap is filled by XWP\textsuperscript{c}, where the absent effect of all other changes increases the distance measure applied, thus 
premitting more changes to take effect when considering the input signal. Additional results from applying our variants to more samples from the Typeface MNIST and 
Fashion MNIST datasets are shown in Figures~\ref{fig:tmnist} 
and~\ref{fig:fmnist}, respectively.

Although the selected datasets are relatively simple—thereby mitigating some of the theoretical challenges associated with baseline selection in occlusion methods—our 
proposed variants demonstrate greater stability across evaluations. They consistently achieve top-tier performance, while competing methods exhibit significant 
fluctuations As demonstrated by the heatmap visualizations, our variants strike an optimal balance between cleanliness and pattern recognition, while also 
effectively capturing negative patterns arising from polysemantic neurons in a clear and interpretable manner. The complementary strengths of the two methods 
enhance their collective ability to explain a model's decision-making process comprehensively.

In summary, these experiments confirm that the framework of weight perturbation—embodied by XWP and XWP\textsuperscript{c}—demonstrates greater robustness in 
model interpretability, establishing it as a reliable approach for the field of Explainable AI.

\section{Conclusion}

In this paper, we introduced a novel framework for model explainability through a structured approach exploiting pattern formation and weight perturbation. 
Its design stems from the existence of interpretable patterns within the model's weights, paving the way for the exploitation of spatial resolution.
The identification of neuronal interactions where this resolution was preserved, was then combined with weight perturbation and occlusion. By leveraging the 
model's states, this design manages to effectively circumvent persistent challenges in occlusion methods—such as out-of-distribution data and added bias. 
Overall, it comprises a structured framework for model interpretability.

We introduced two variants in this framework and evaluated their performance against state-of-the-art attribution methods. Our approach demonstrated greater 
preformance when considering both datasets, outperforming classical occlusion techniques while producing cleaner and more interpretable visualizations. These 
come with reduced noise and fewer artifacts, yet no exceptions in pattern recognition.
 
This work represents a critical first step toward addressing the theoretical challenges of Occlusion. The simplicity of the selected datasets in our experimental 
setup mitigates the problem of baseline selection. The method's true potential lies in its extension to more complex datasets and architectures, such 
as Transformers or Convolutional Neural Networks, and its integration with causal analysis—paving the way for more robust and interpretable AI systems.

\bibliographystyle{ACM-Reference-Format}
\bibliography{main}

\appendix
\begin{figure*}[p]
    \centering
    \setlength{\tabcolsep}{2pt}  

    \begin{tabular}{c c c c c c c c}
        \textbf{Sample} & \textbf{OCCL} & \textbf{SHAP} & \textbf{RISE} & \textbf{IG} &
        \textbf{LRP} & \textbf{XWP} & \textbf{XWP\textsuperscript{c}} \\[3pt]

        \includegraphics[width=0.115\textwidth]{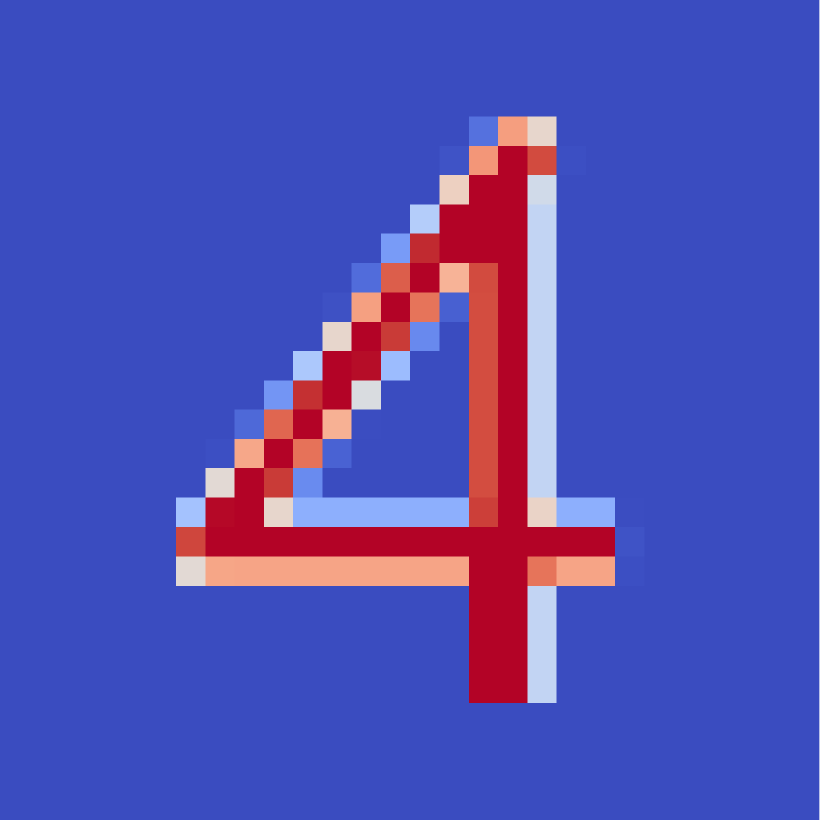} &
        \includegraphics[width=0.115\textwidth]{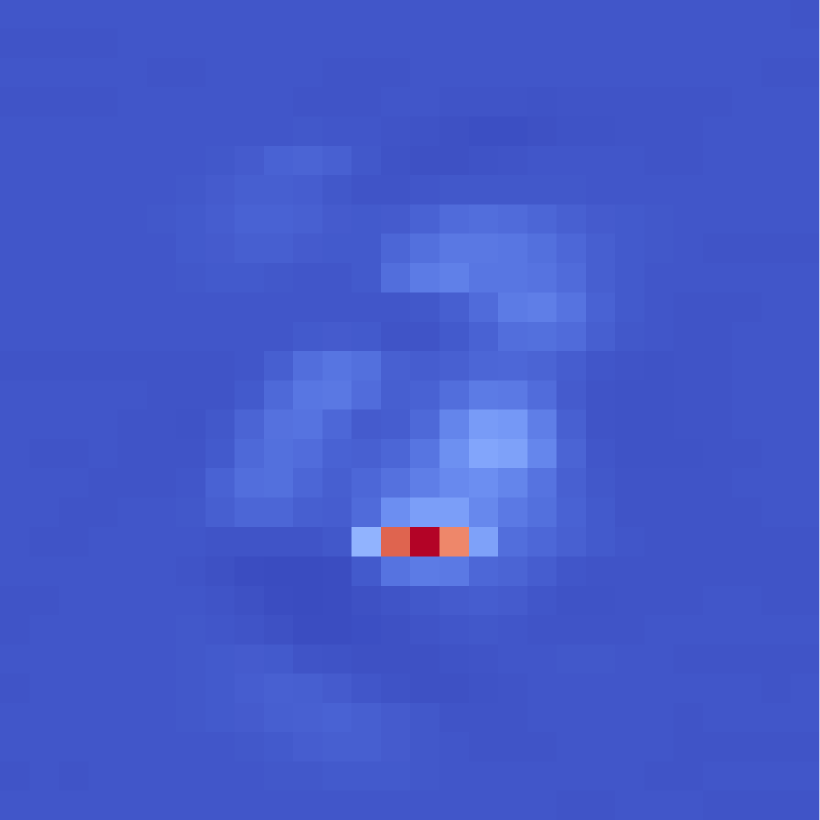} &
        \includegraphics[width=0.115\textwidth]{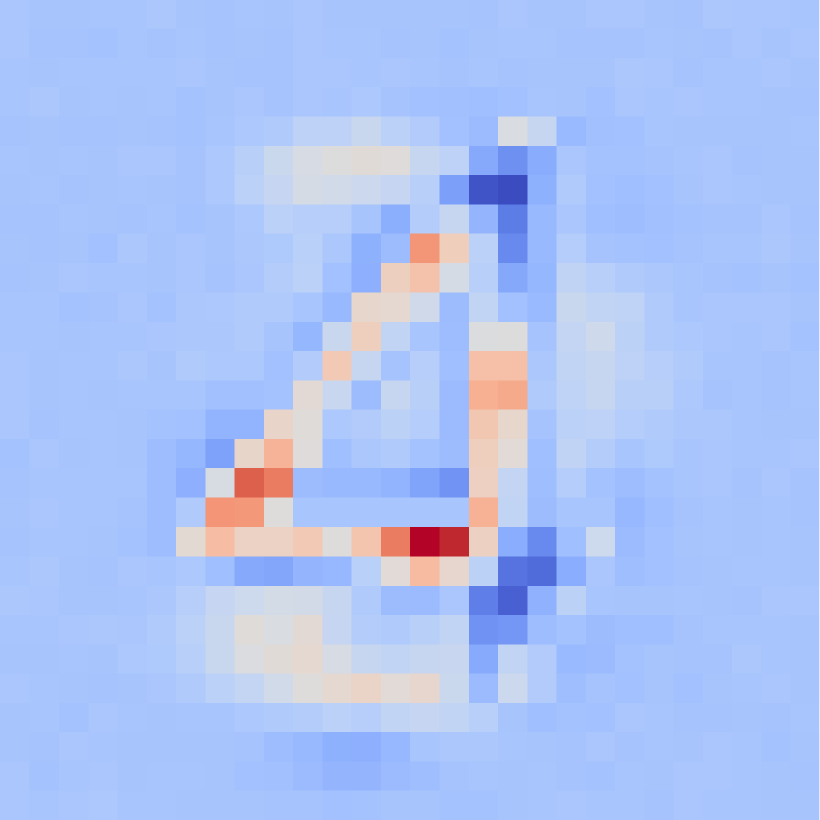} &
        \includegraphics[width=0.115\textwidth]{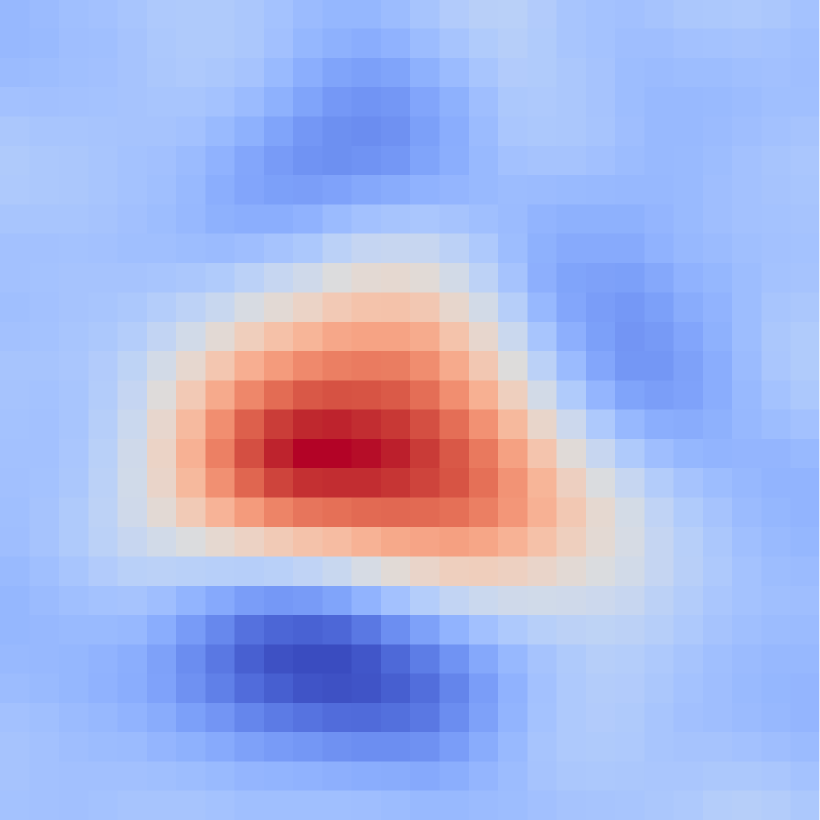} &
        \includegraphics[width=0.115\textwidth]{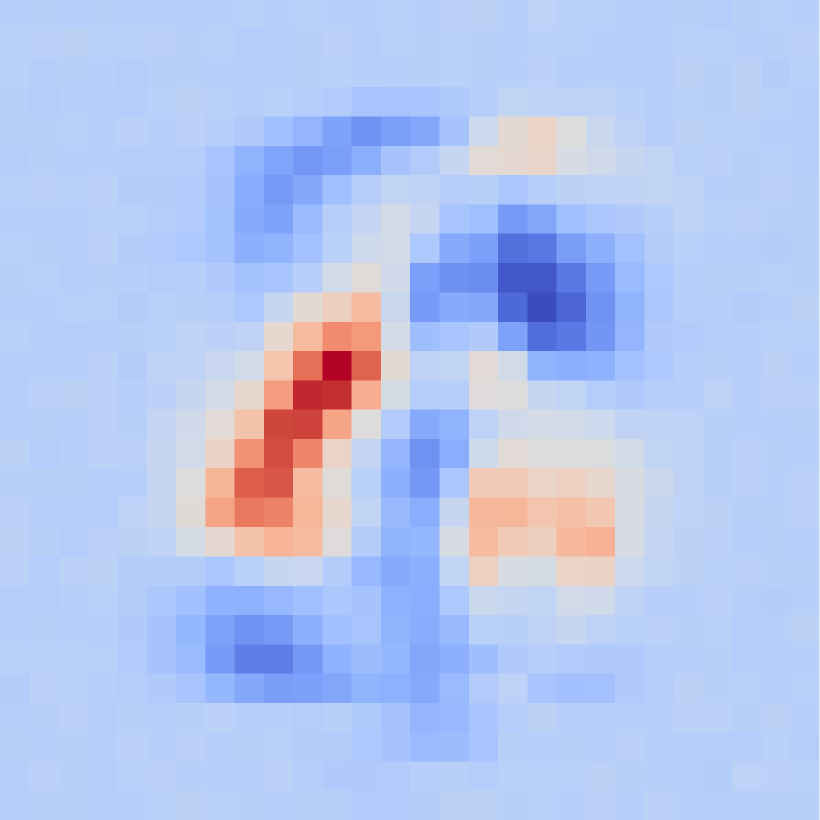} &
        \includegraphics[width=0.115\textwidth]{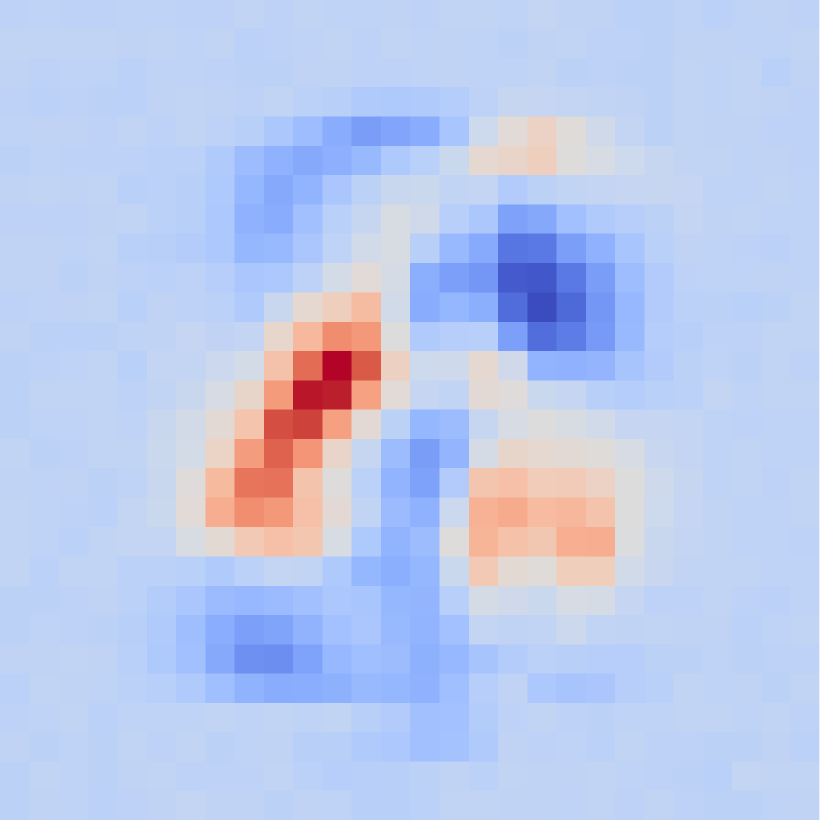} &
        \includegraphics[width=0.115\textwidth]{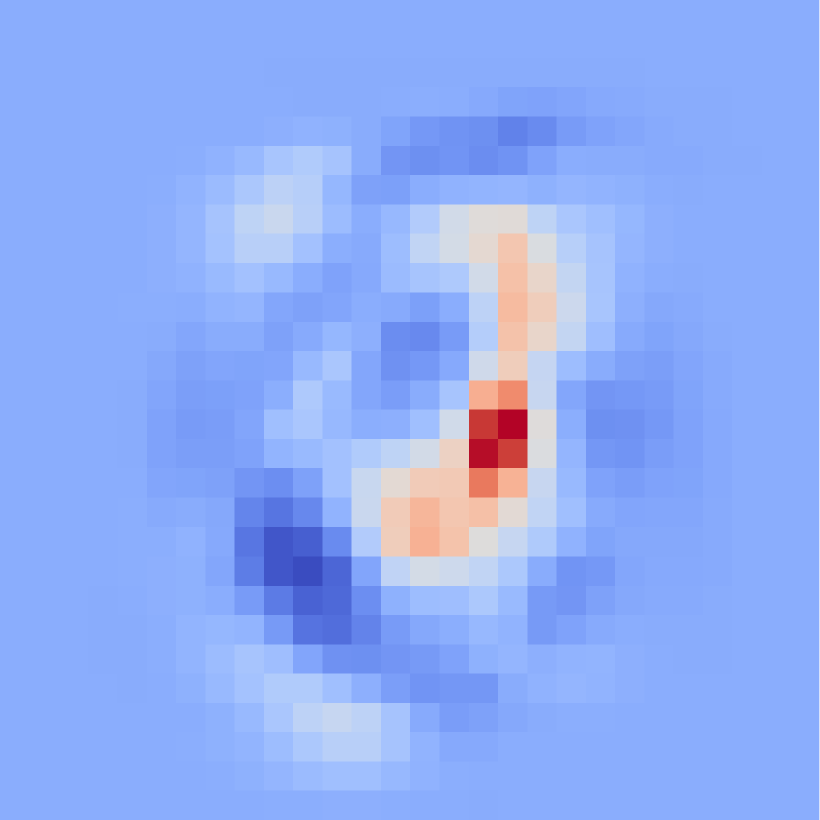} &
        \includegraphics[width=0.115\textwidth]{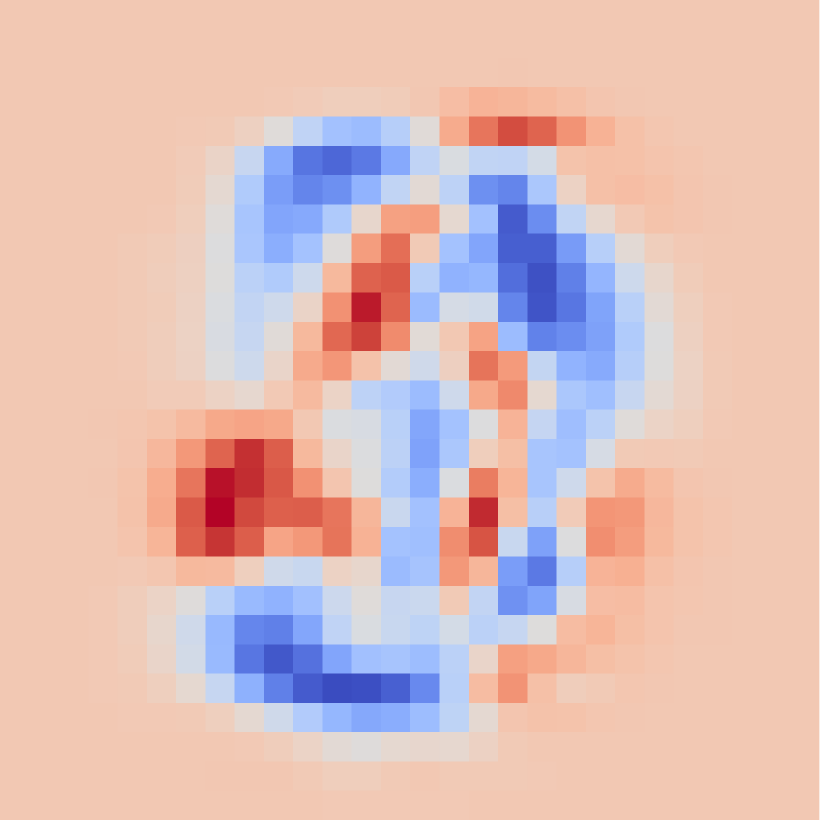} \\[3pt]

        \includegraphics[width=0.115\textwidth]{Experiments/typeface_mnist/sample_2.pdf} &
        \includegraphics[width=0.115\textwidth]{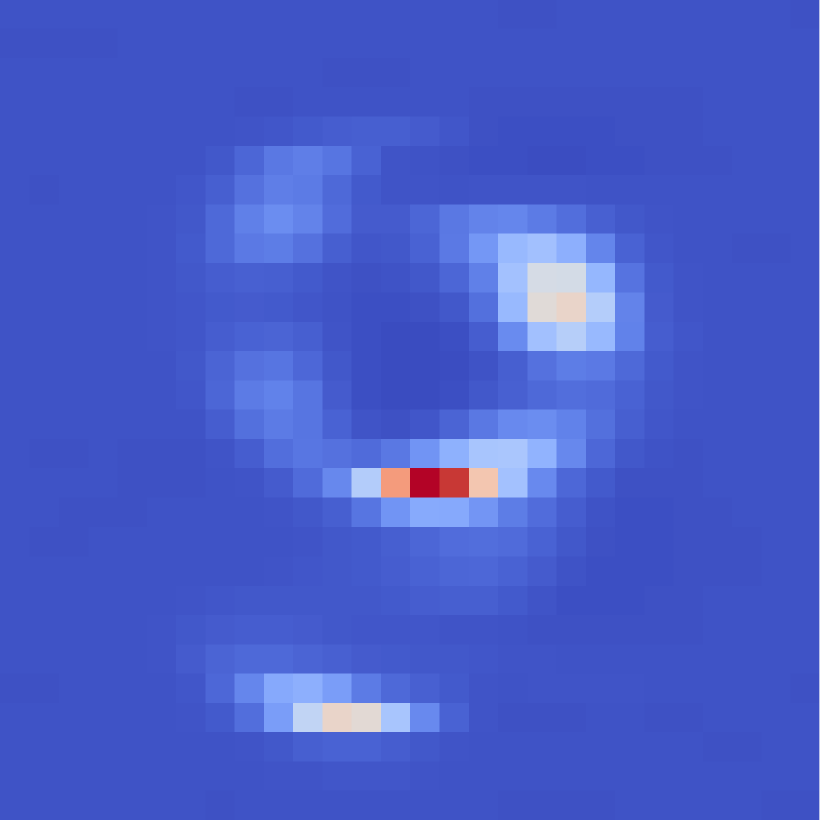} &
        \includegraphics[width=0.115\textwidth]{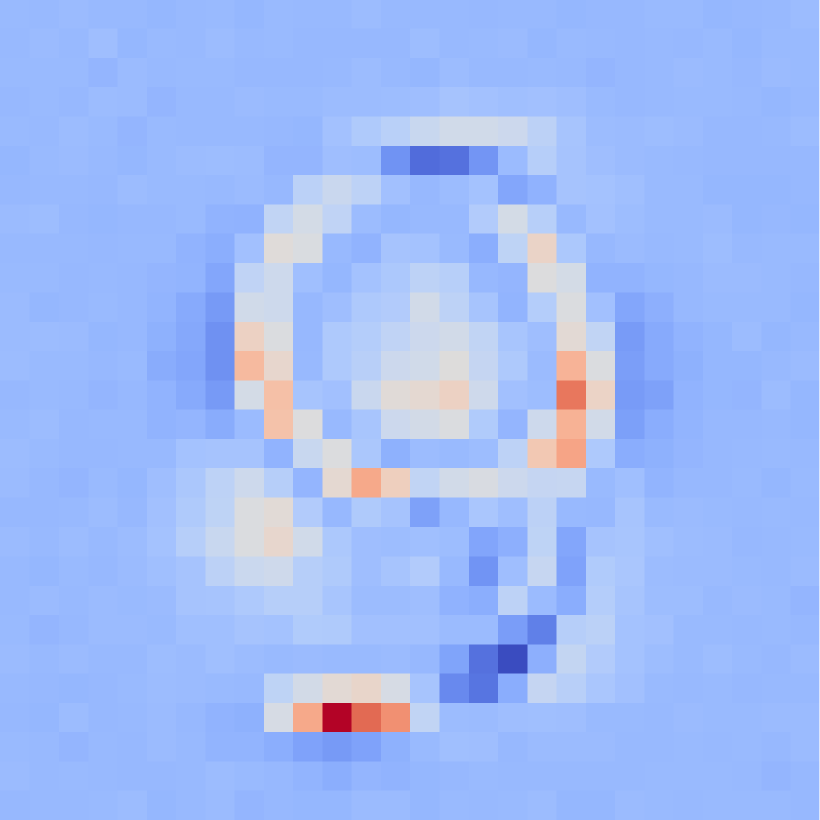} &
        \includegraphics[width=0.115\textwidth]{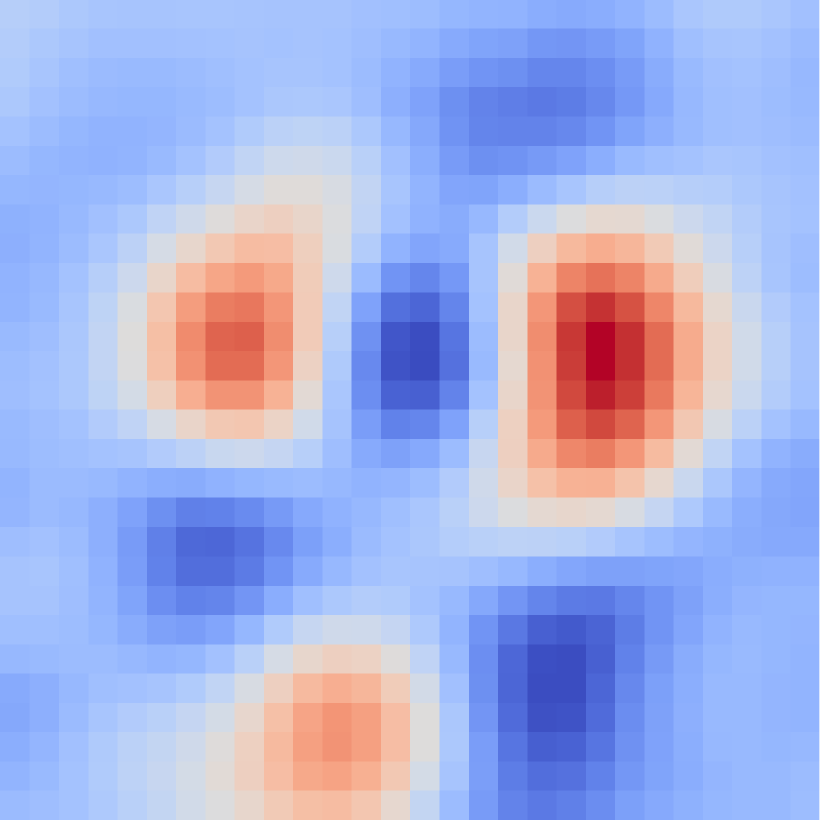} &
        \includegraphics[width=0.115\textwidth]{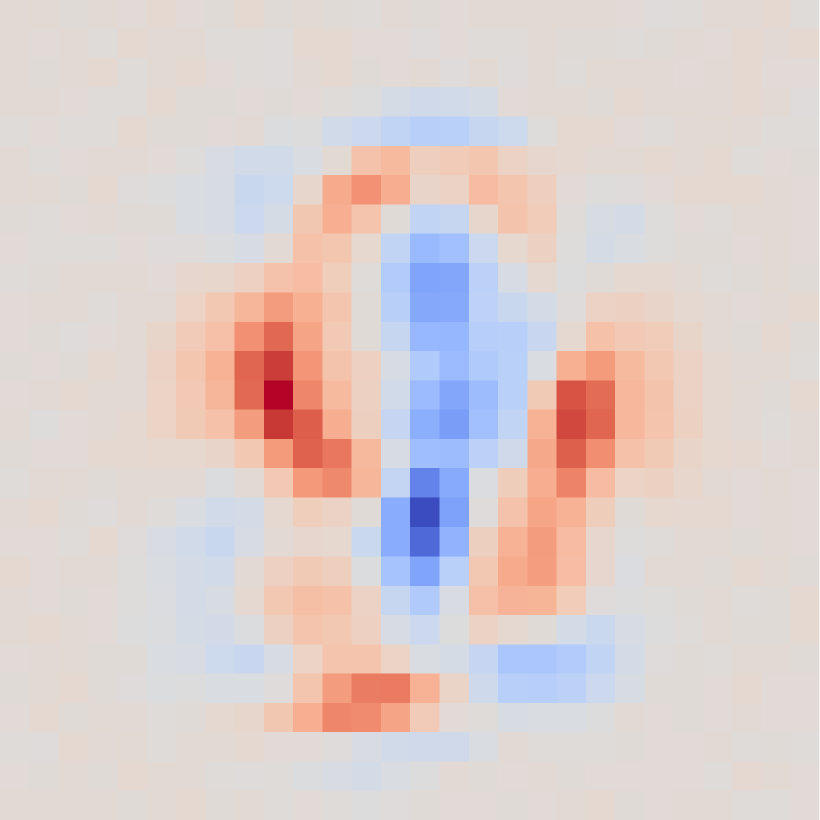} &
        \includegraphics[width=0.115\textwidth]{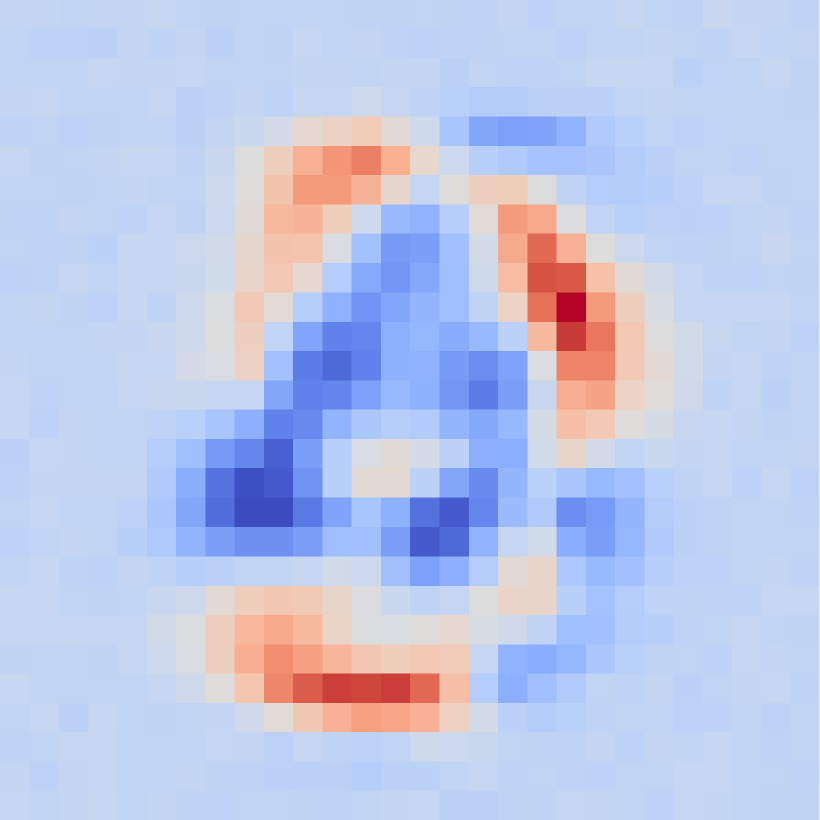} &
        \includegraphics[width=0.115\textwidth]{Experiments/typeface_mnist/XWP_2.pdf} &
        \includegraphics[width=0.115\textwidth]{Experiments/typeface_mnist/XWP_C_2.pdf} \\[3pt]

        \includegraphics[width=0.115\textwidth]{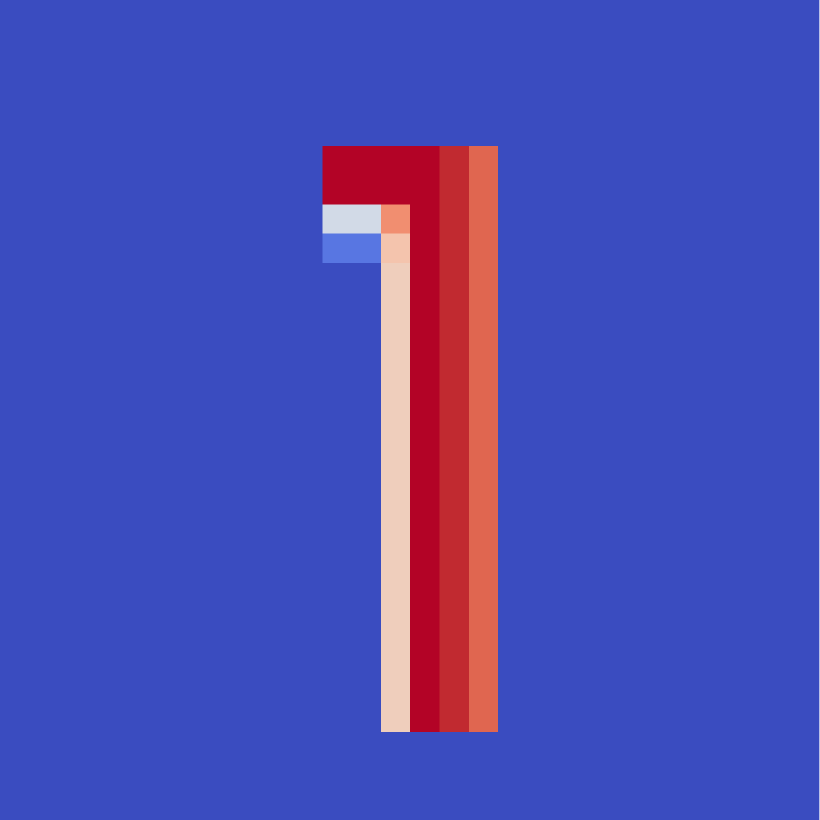} &
        \includegraphics[width=0.115\textwidth]{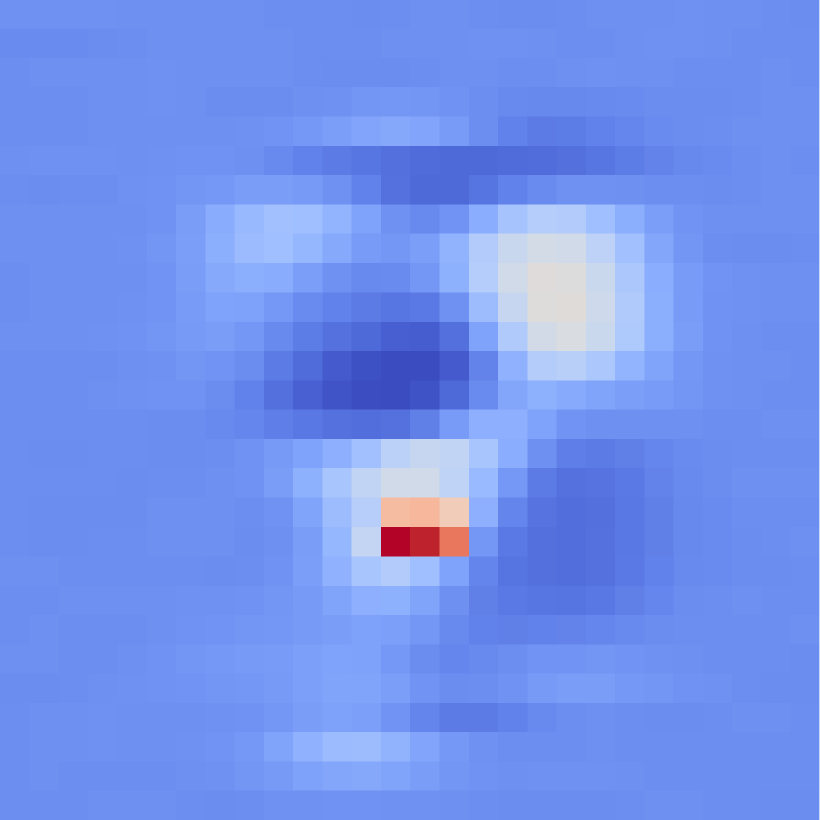} &
        \includegraphics[width=0.115\textwidth]{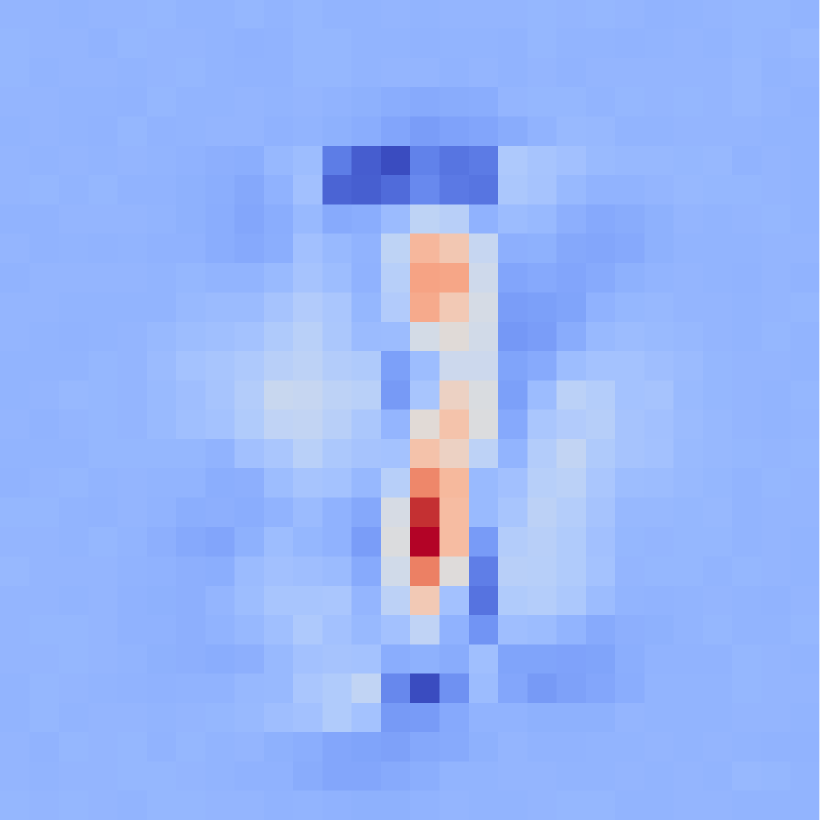} &
        \includegraphics[width=0.115\textwidth]{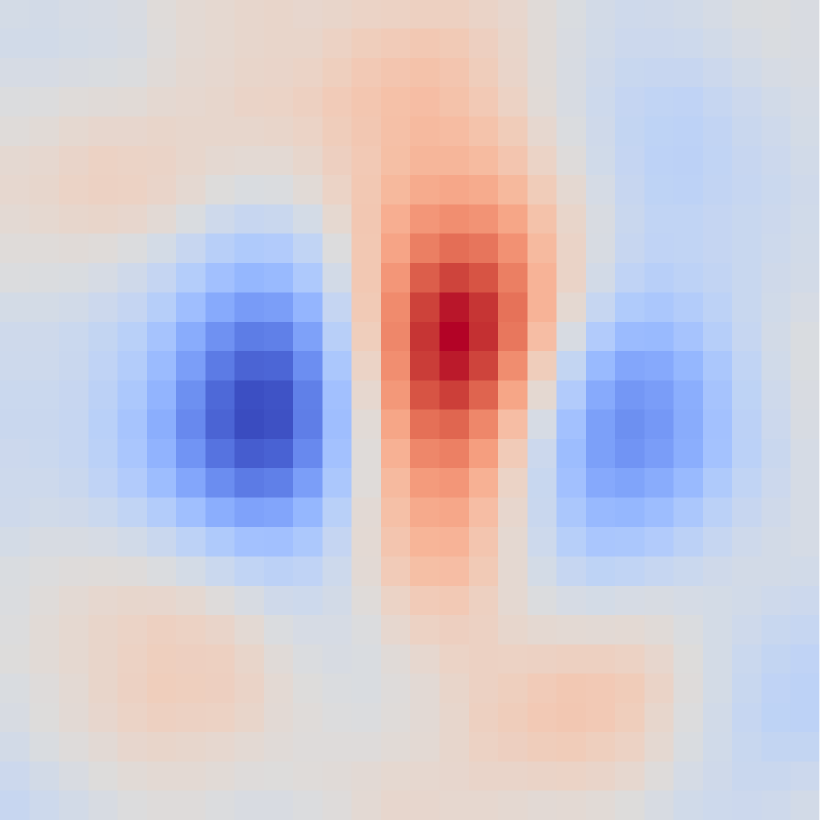} &
        \includegraphics[width=0.115\textwidth]{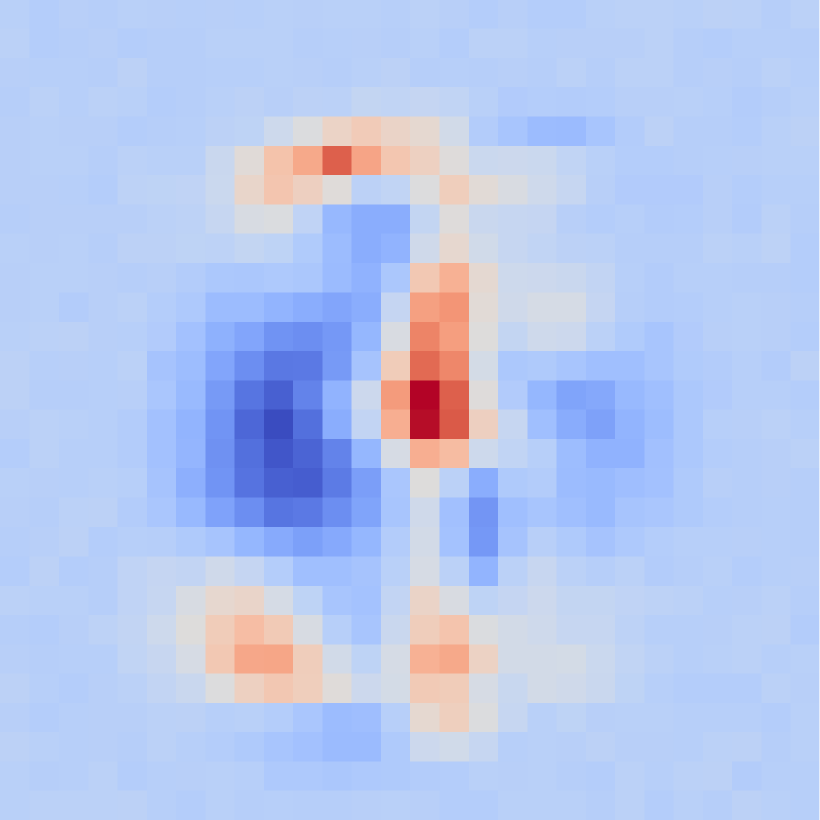} &
        \includegraphics[width=0.115\textwidth]{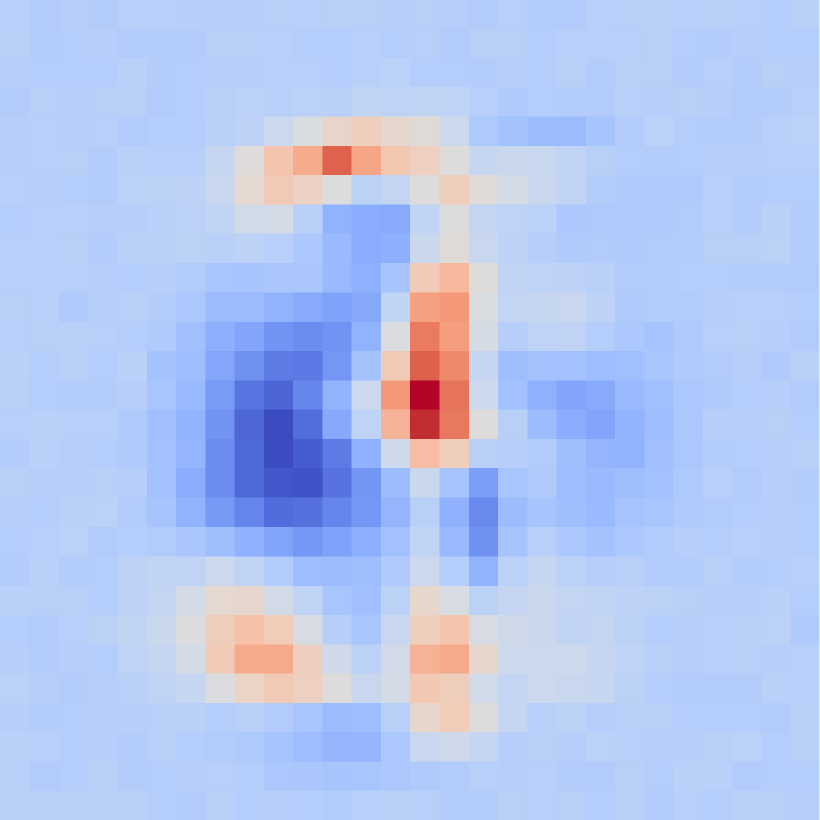} &
        \includegraphics[width=0.115\textwidth]{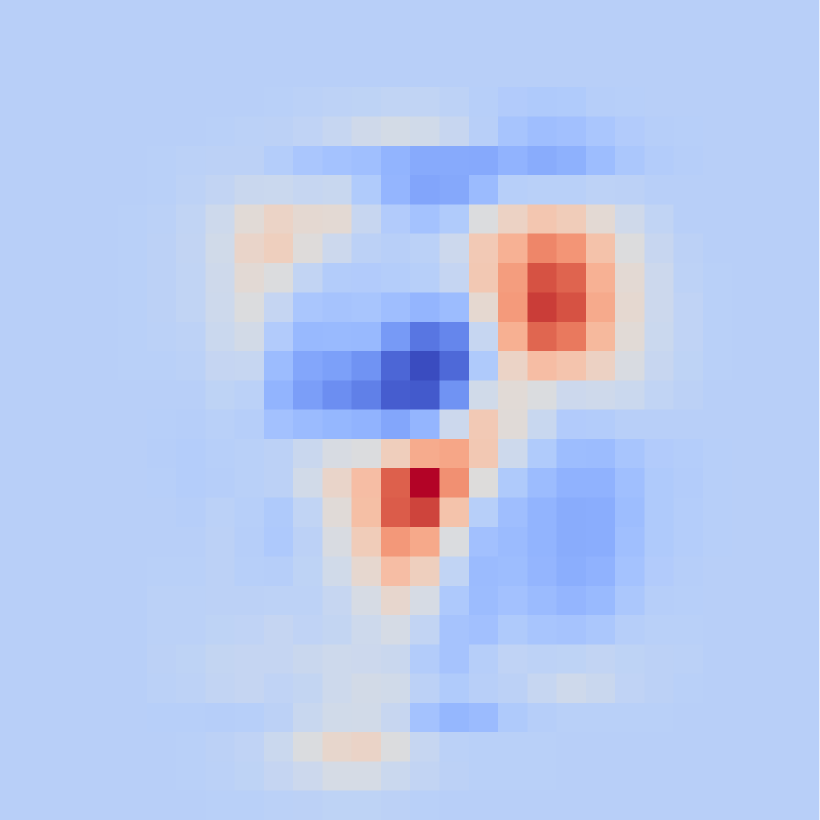} &
        \includegraphics[width=0.115\textwidth]{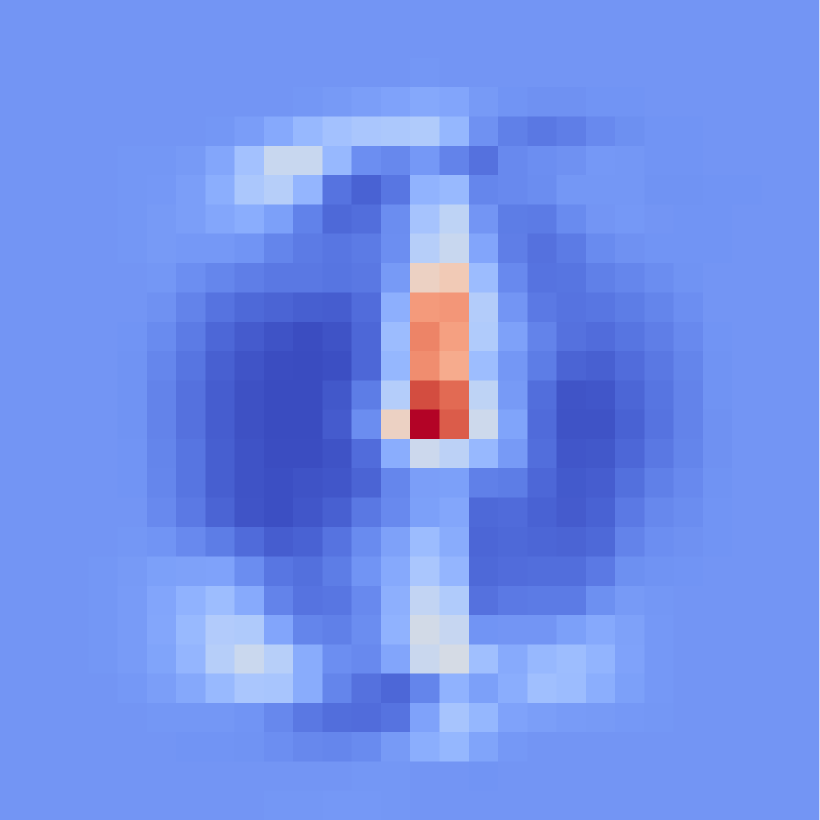} \\[3pt]

        \includegraphics[width=0.115\textwidth]{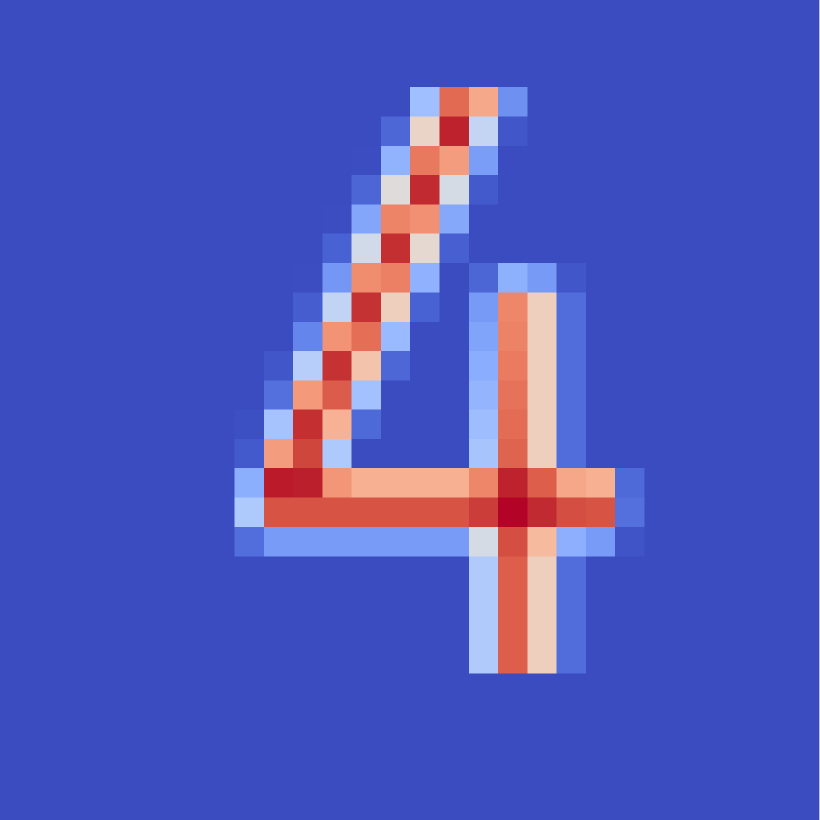} &
        \includegraphics[width=0.115\textwidth]{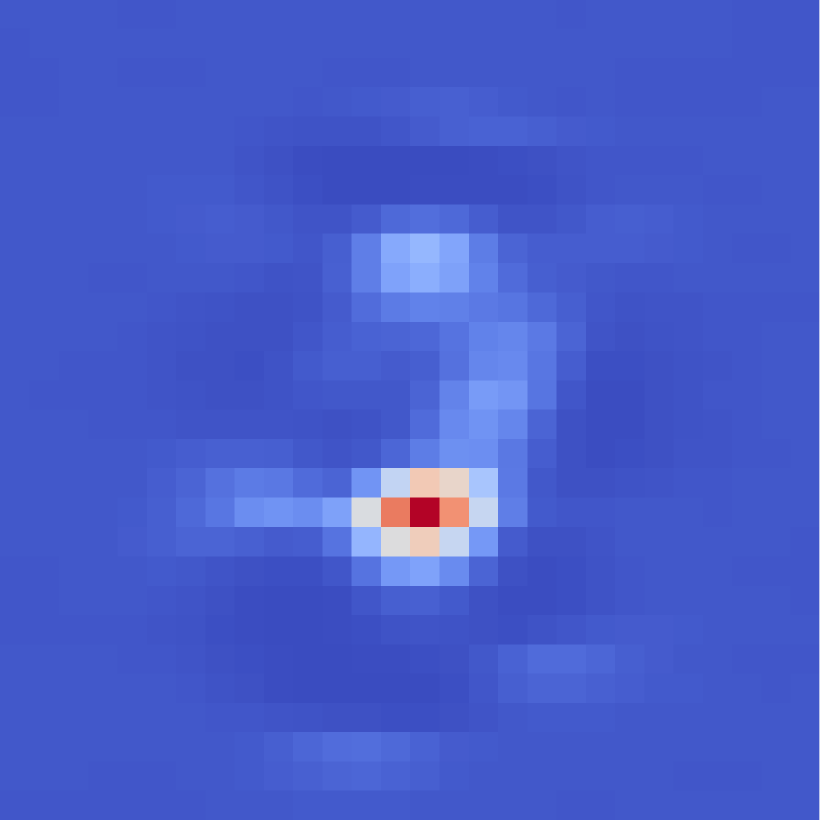} &
        \includegraphics[width=0.115\textwidth]{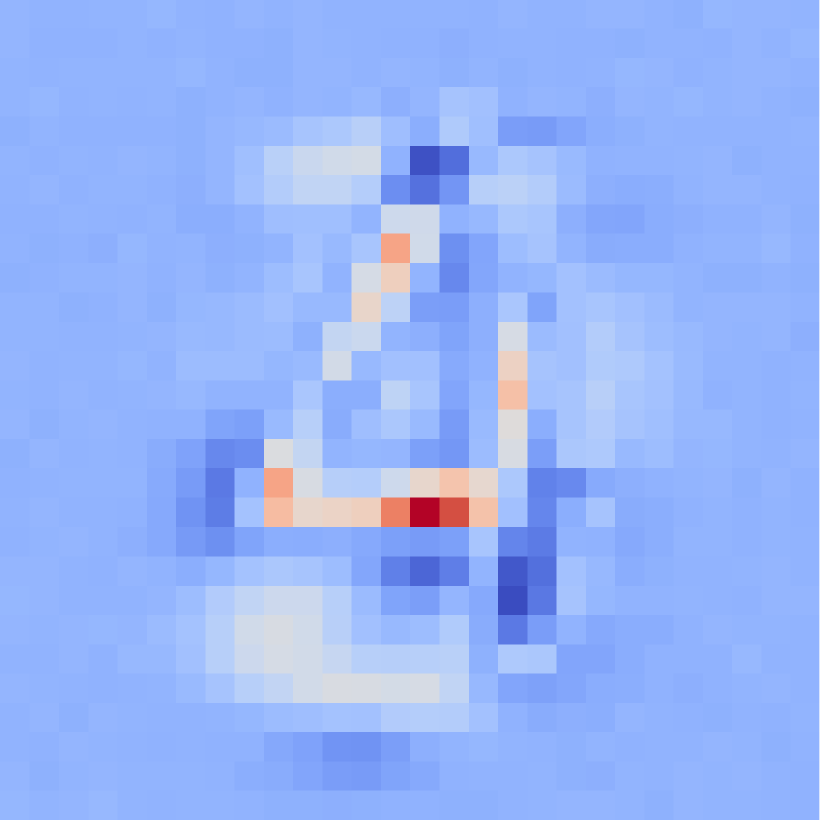} &
        \includegraphics[width=0.115\textwidth]{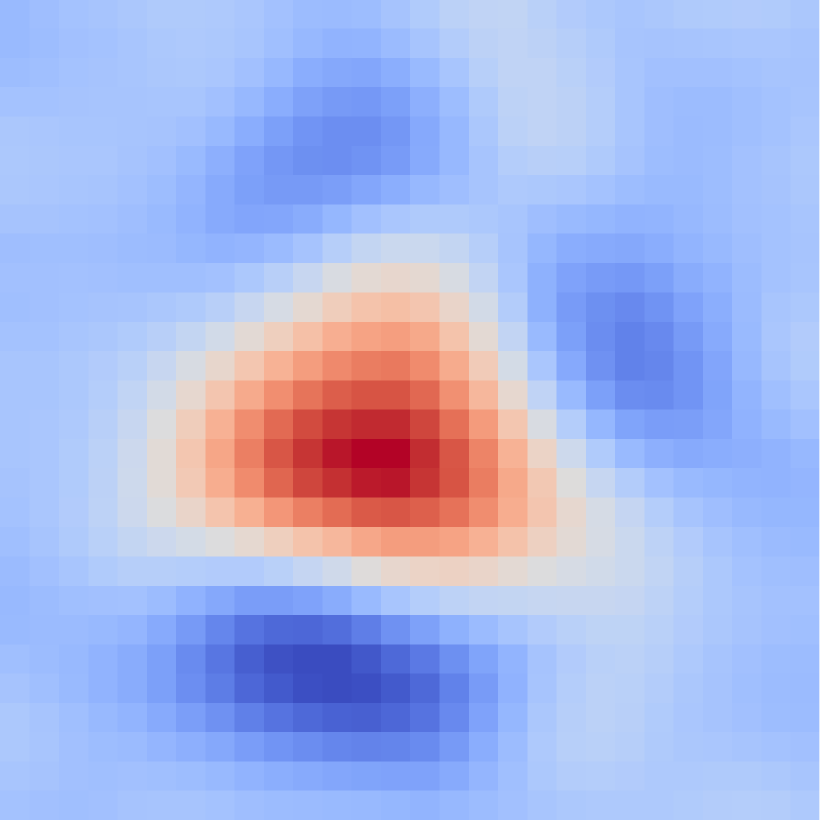} &
        \includegraphics[width=0.115\textwidth]{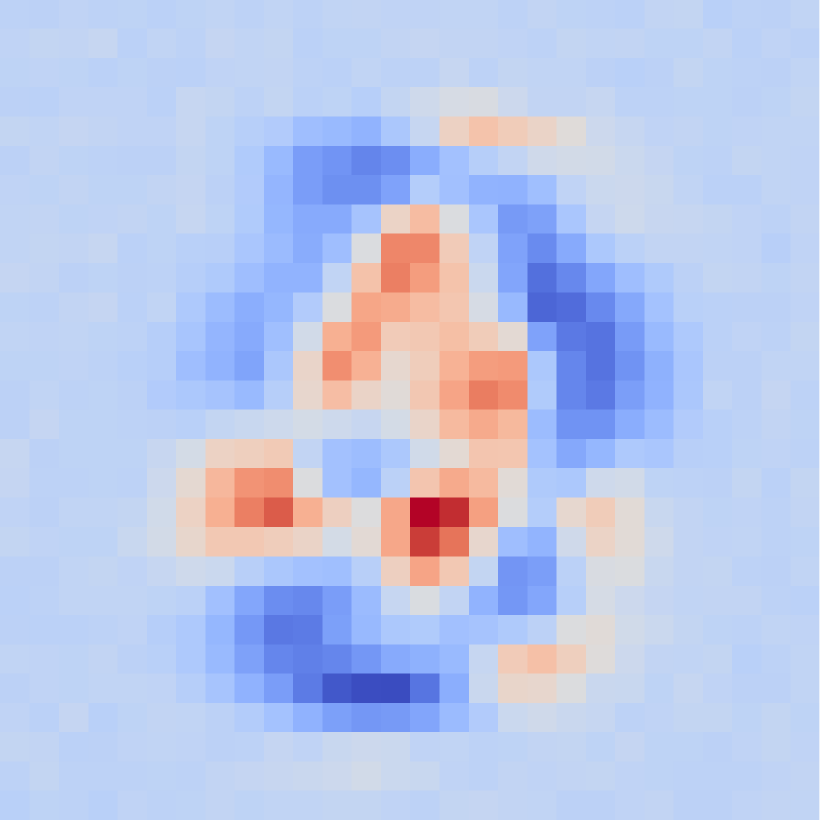} &
        \includegraphics[width=0.115\textwidth]{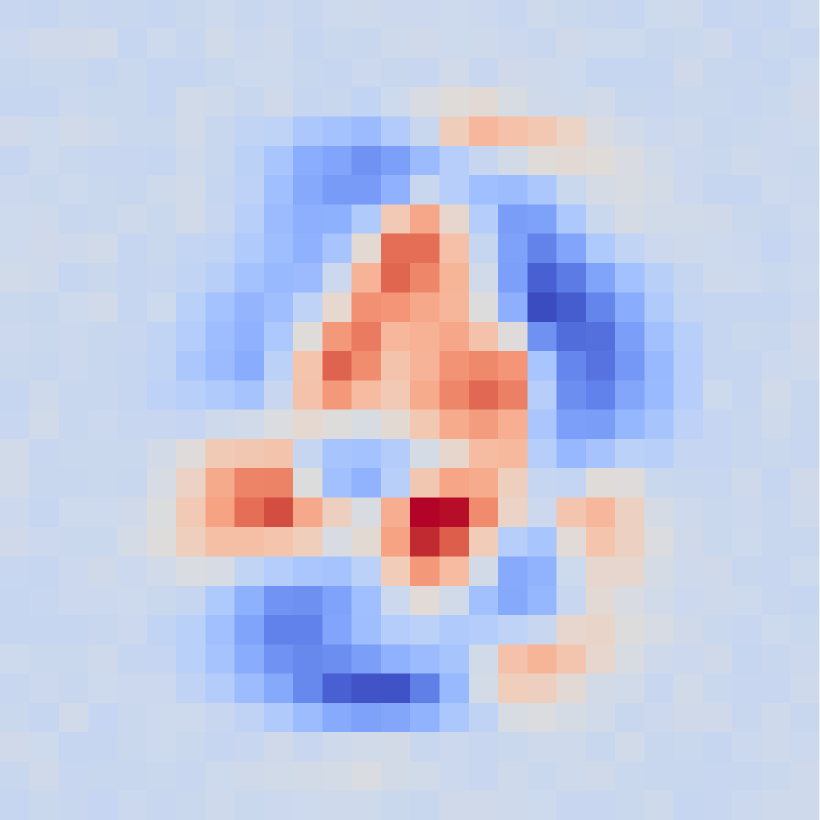} &
        \includegraphics[width=0.115\textwidth]{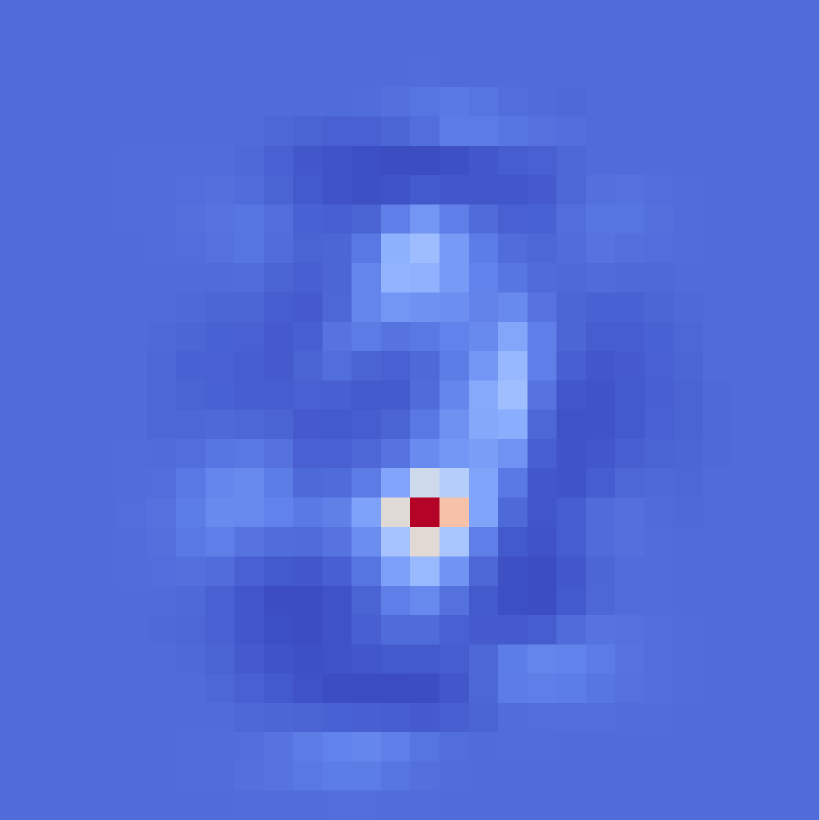} &
        \includegraphics[width=0.115\textwidth]{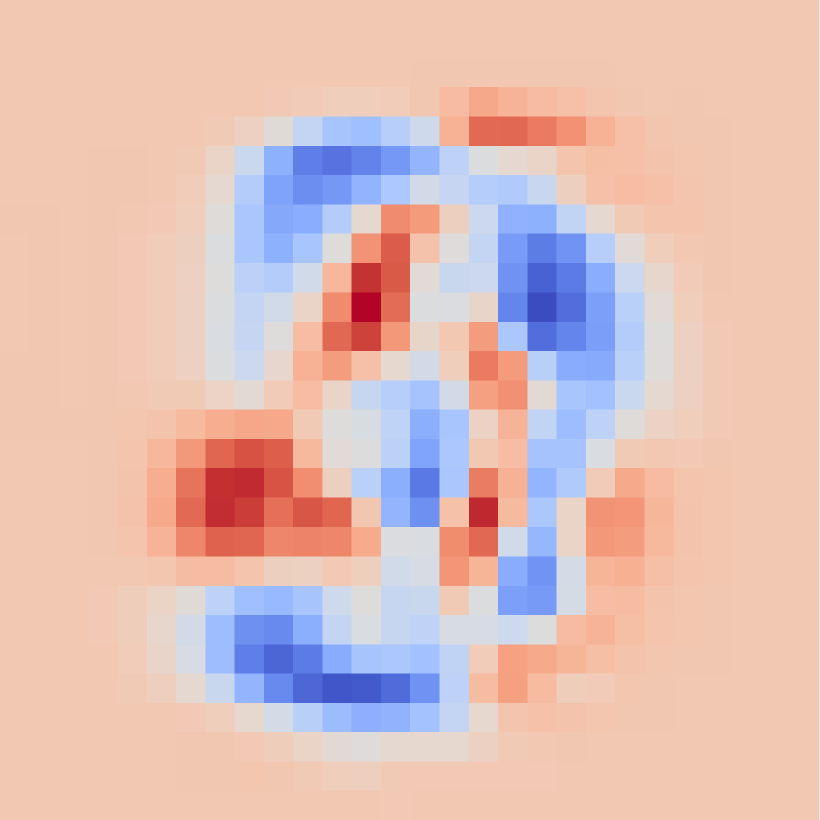} \\[3pt]

        \includegraphics[width=0.115\textwidth]{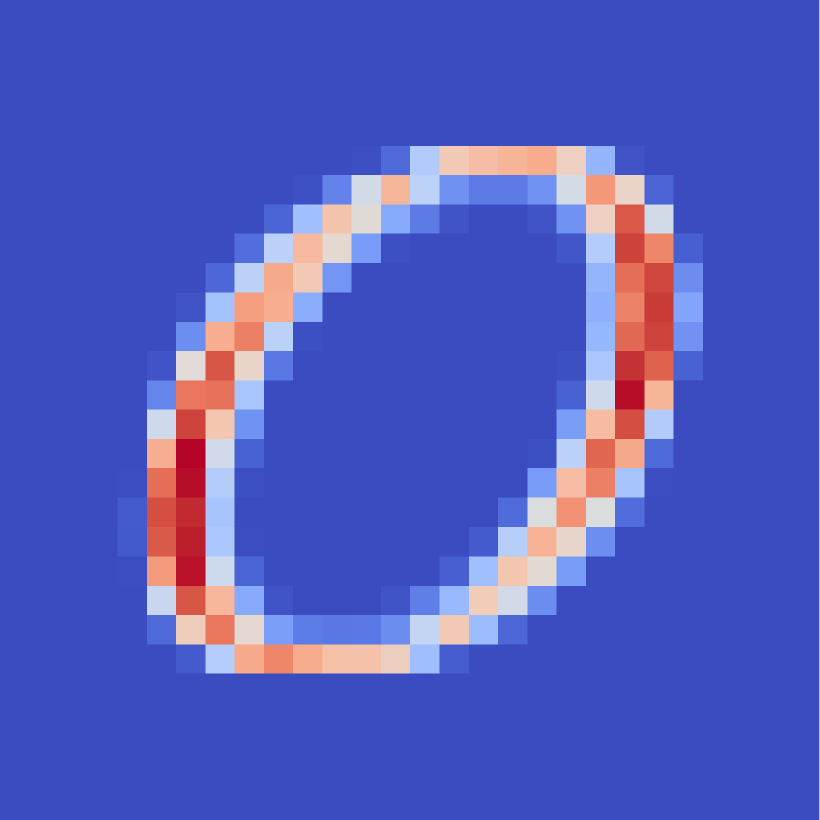} &
        \includegraphics[width=0.115\textwidth]{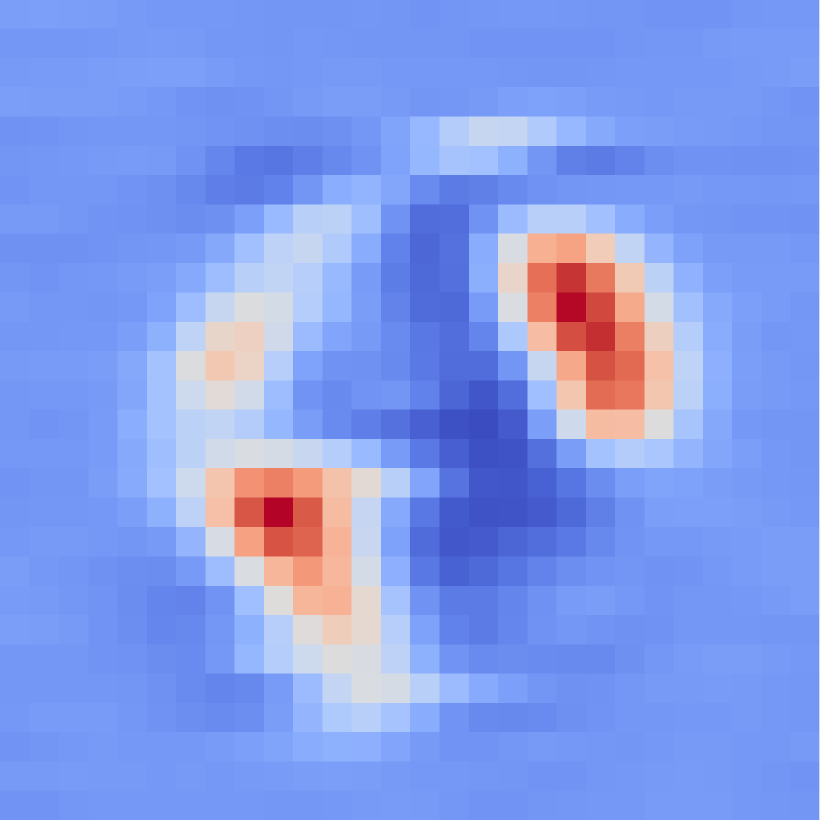} &
        \includegraphics[width=0.115\textwidth]{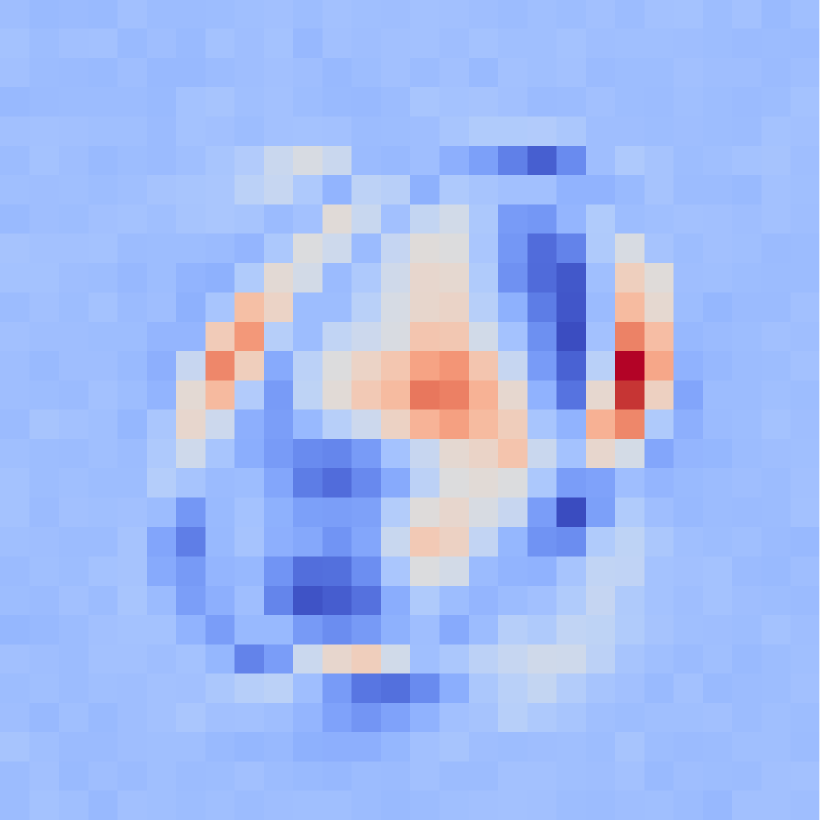} &
        \includegraphics[width=0.115\textwidth]{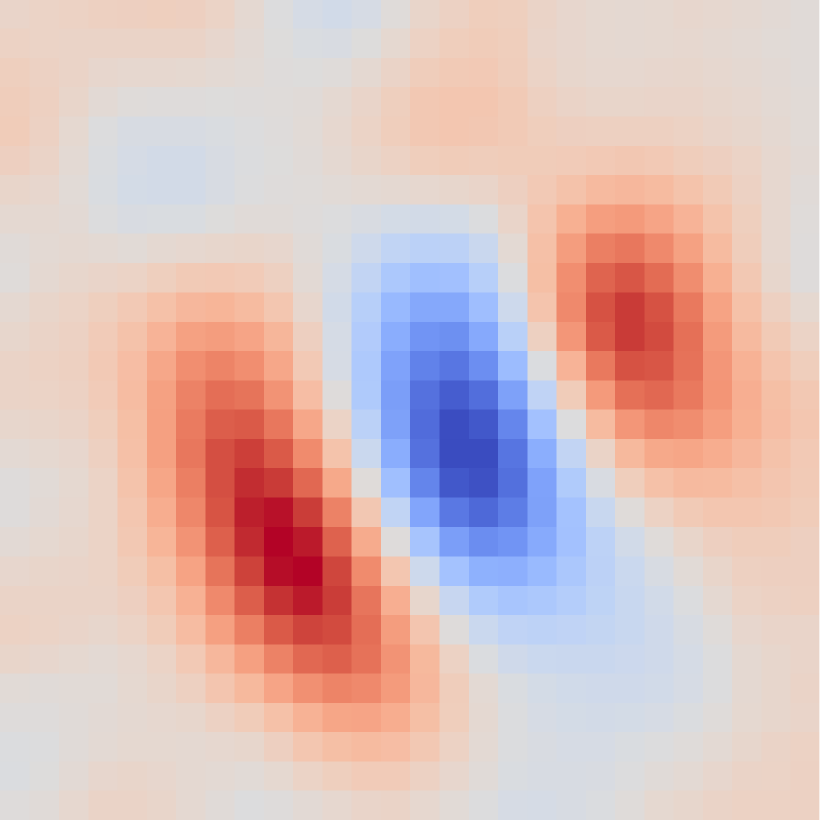} &
        \includegraphics[width=0.115\textwidth]{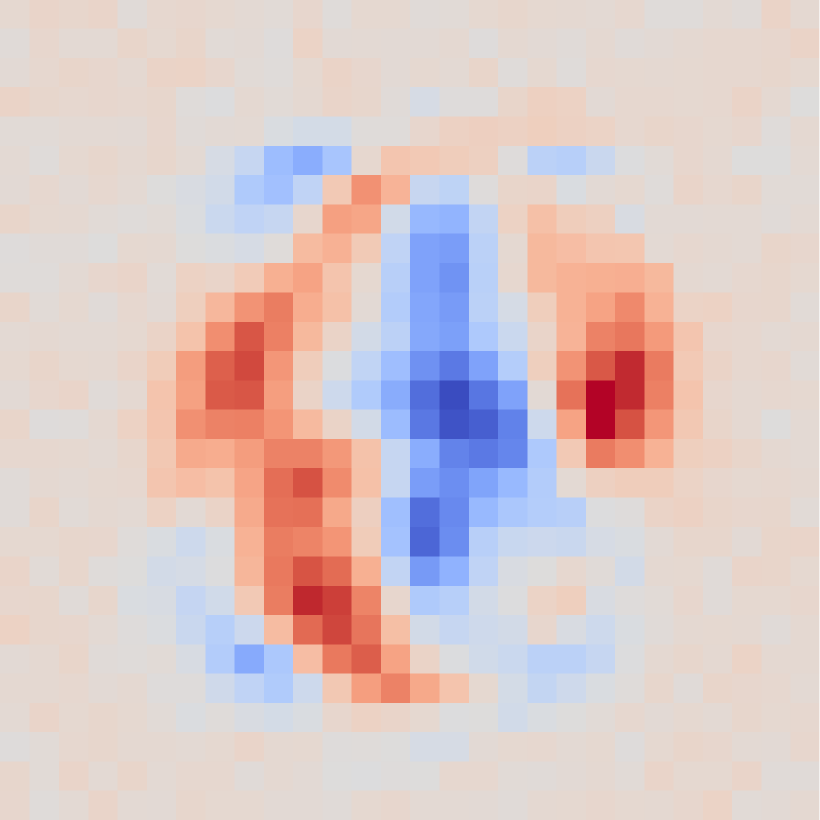} &
        \includegraphics[width=0.115\textwidth]{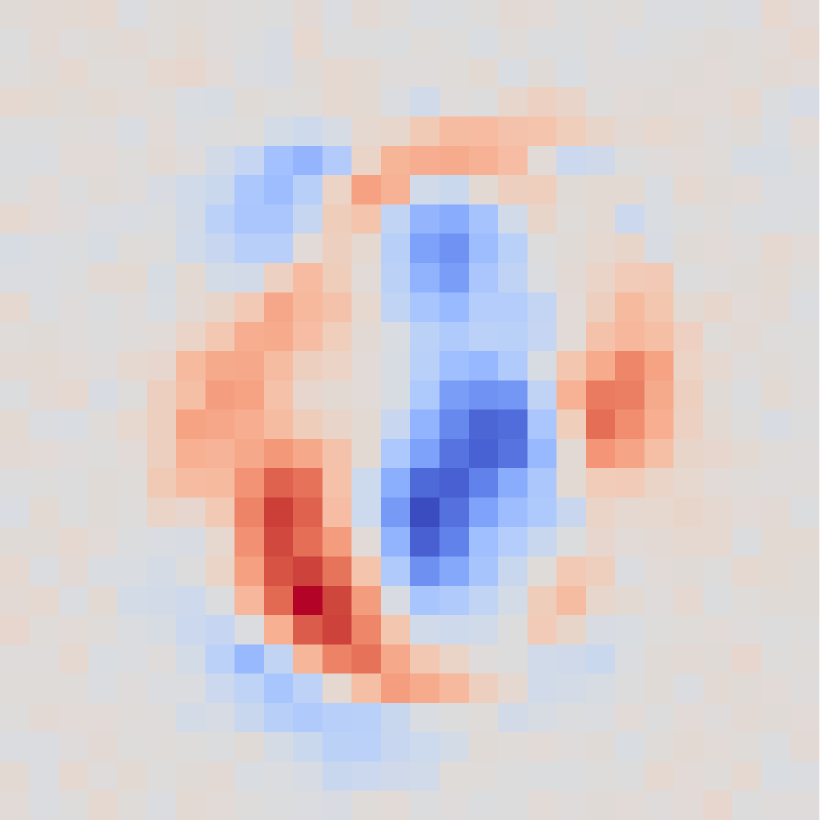} &
        \includegraphics[width=0.115\textwidth]{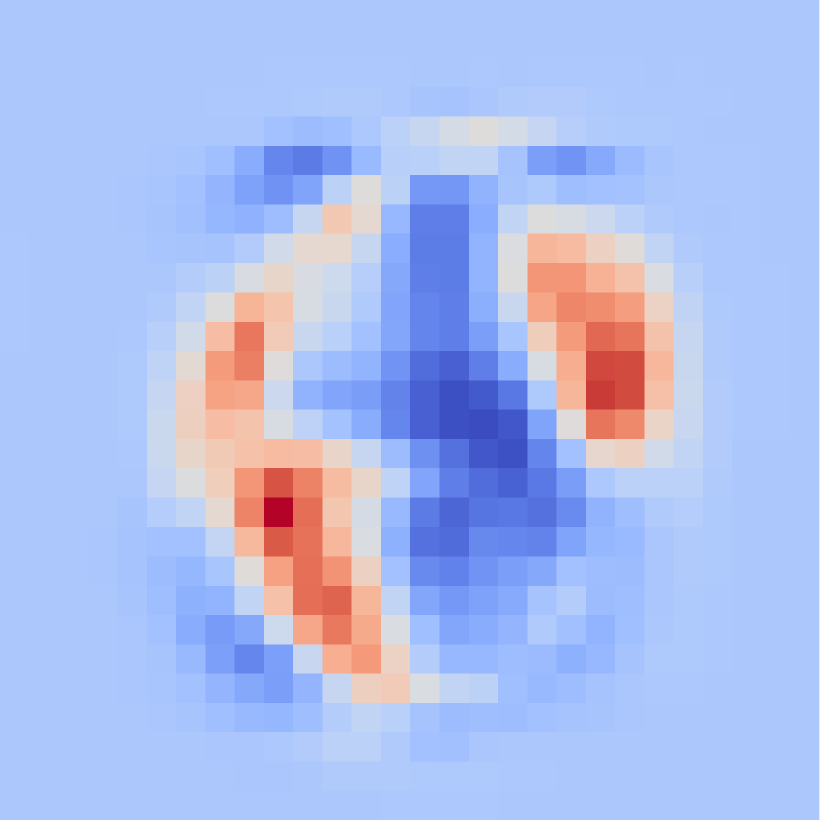} &
        \includegraphics[width=0.115\textwidth]{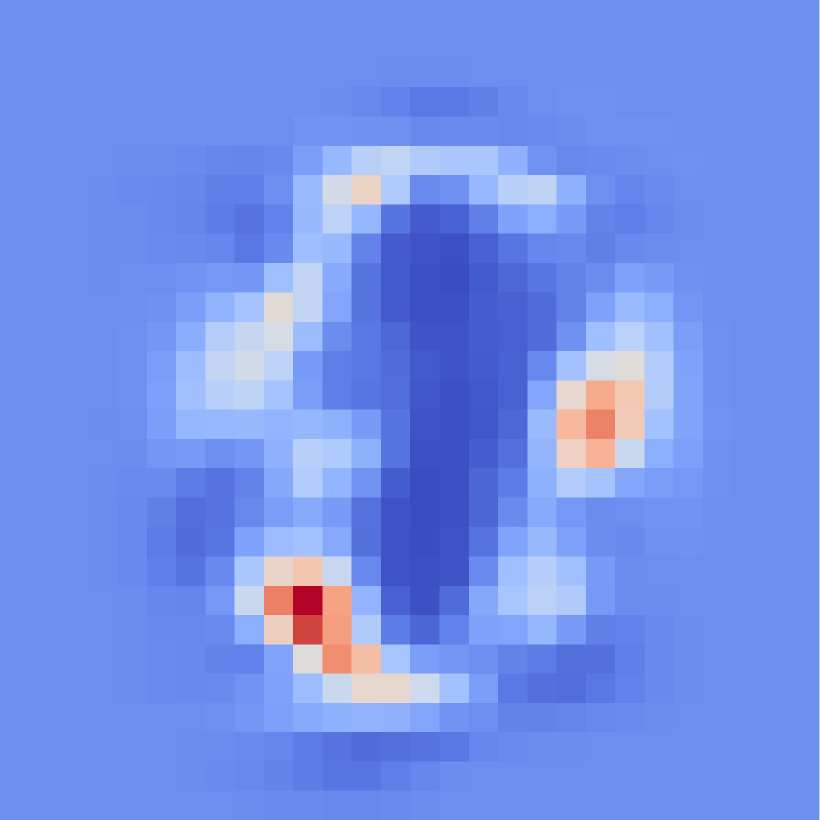} \\[3pt]

        \includegraphics[width=0.115\textwidth]{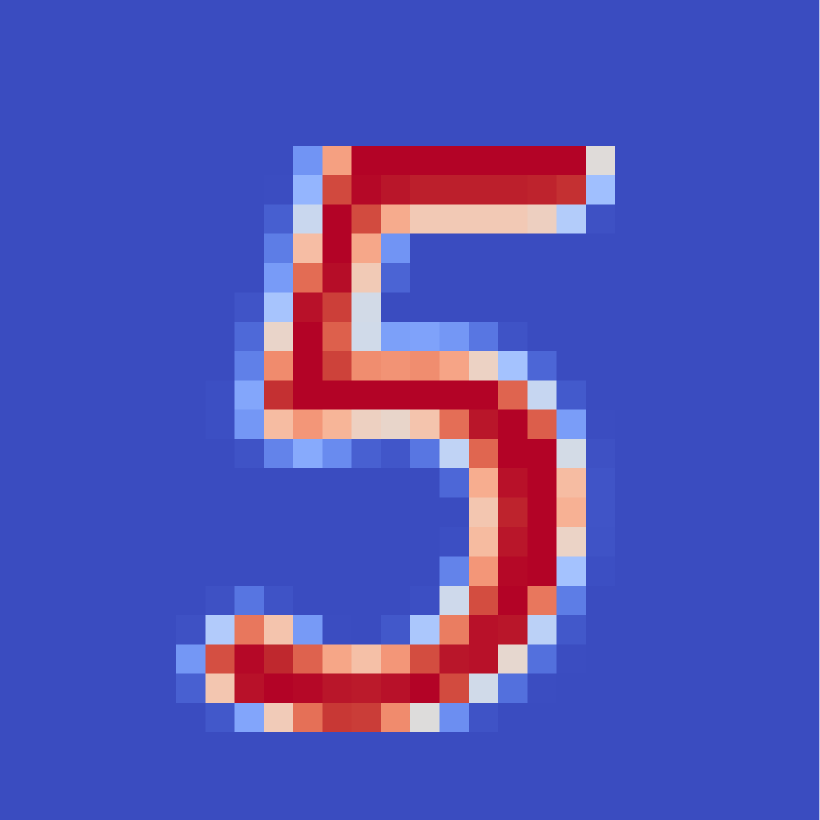} &
        \includegraphics[width=0.115\textwidth]{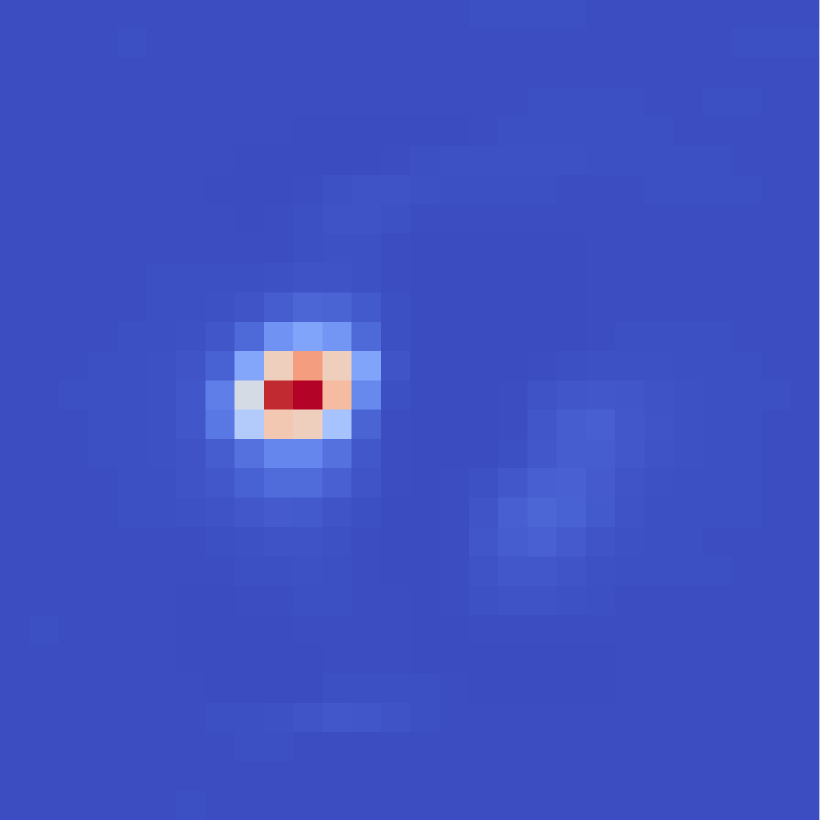} &
        \includegraphics[width=0.115\textwidth]{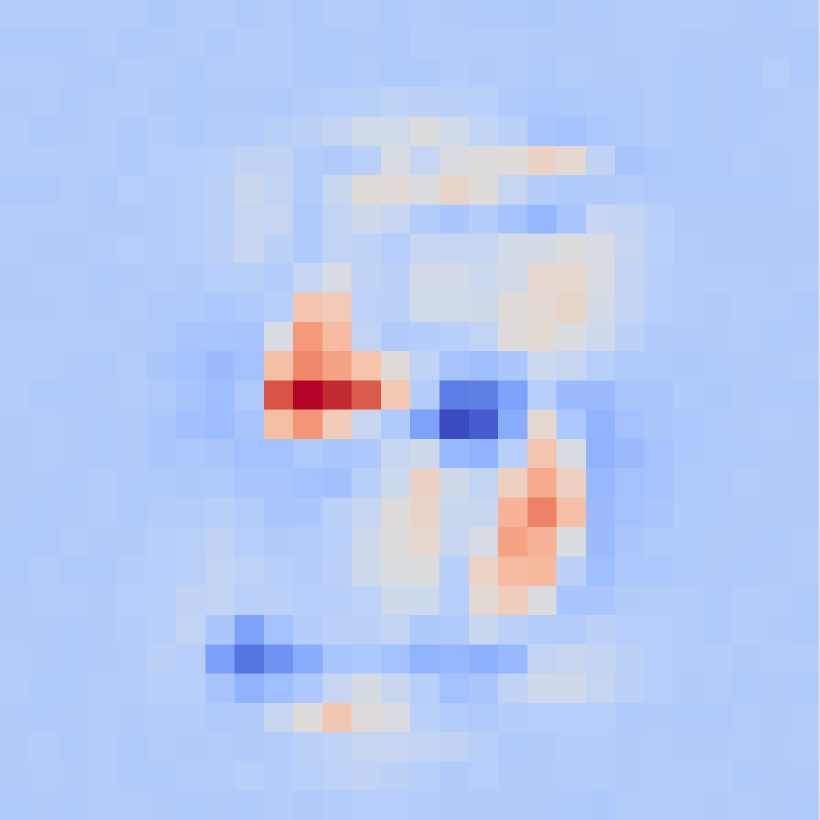} &
        \includegraphics[width=0.115\textwidth]{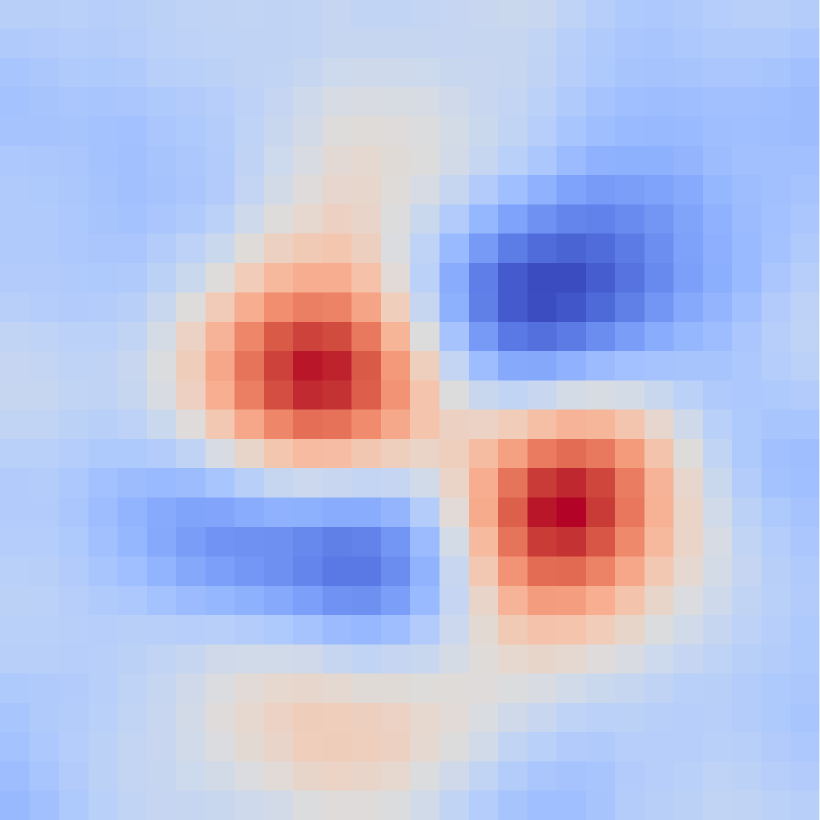} &
        \includegraphics[width=0.115\textwidth]{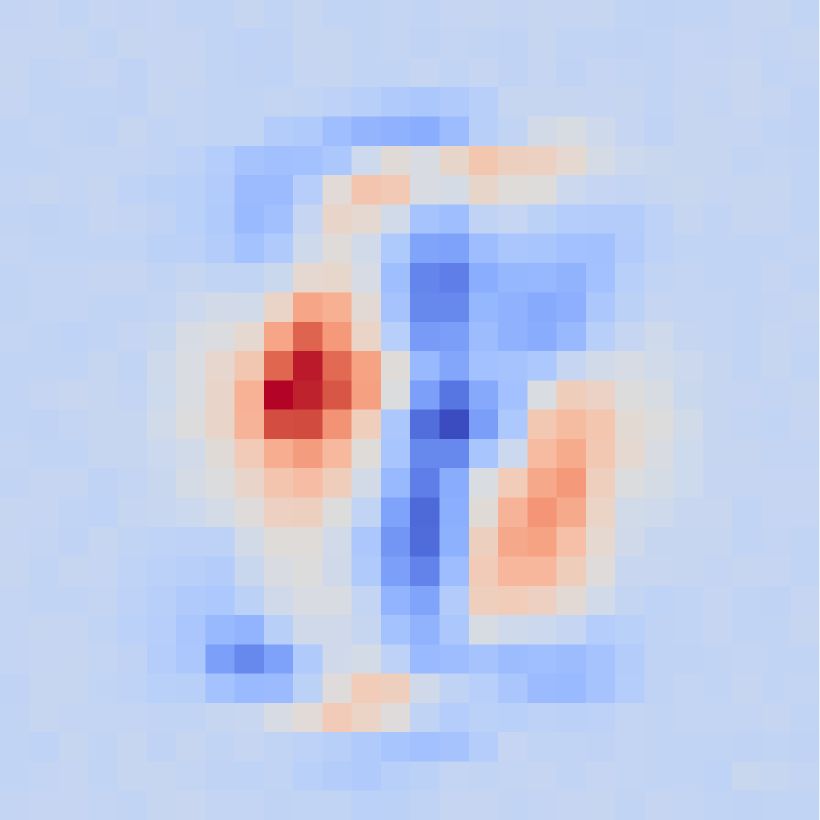} &
        \includegraphics[width=0.115\textwidth]{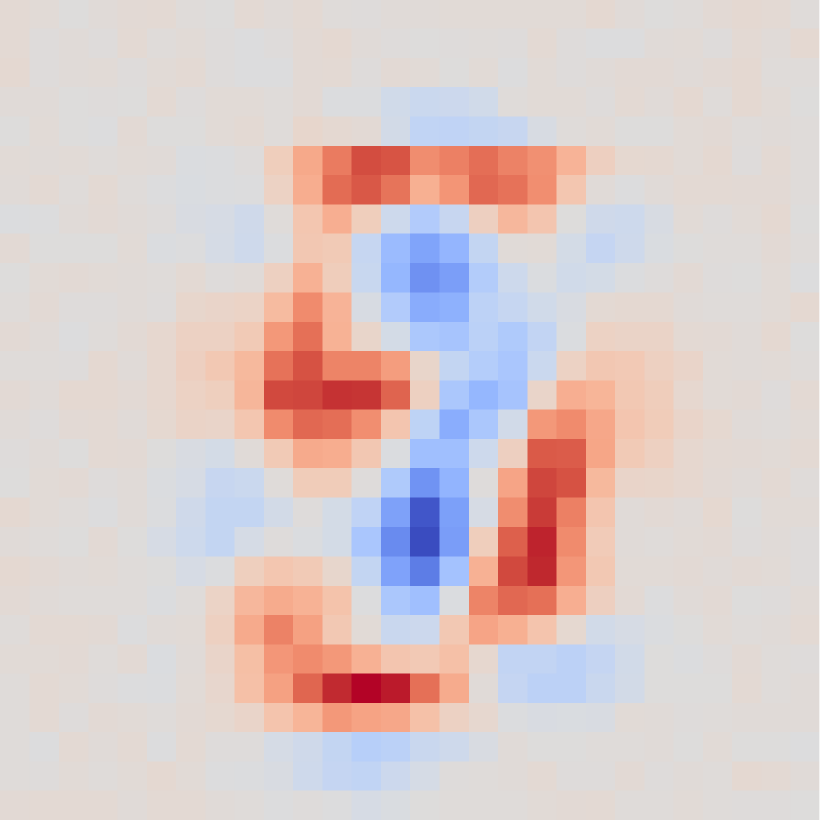} &
        \includegraphics[width=0.115\textwidth]{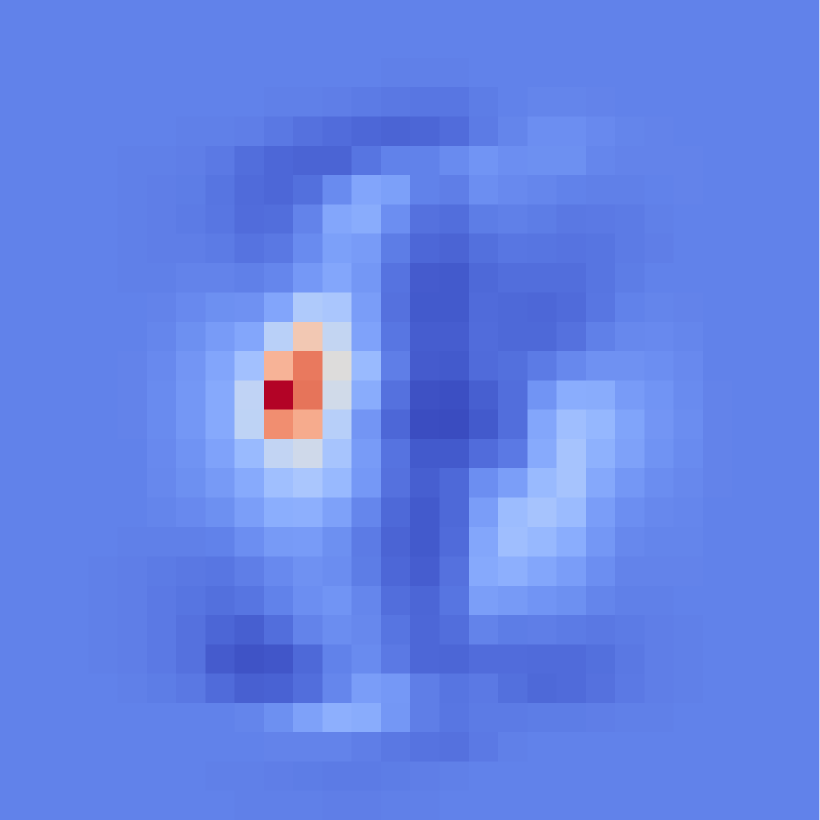} &
        \includegraphics[width=0.115\textwidth]{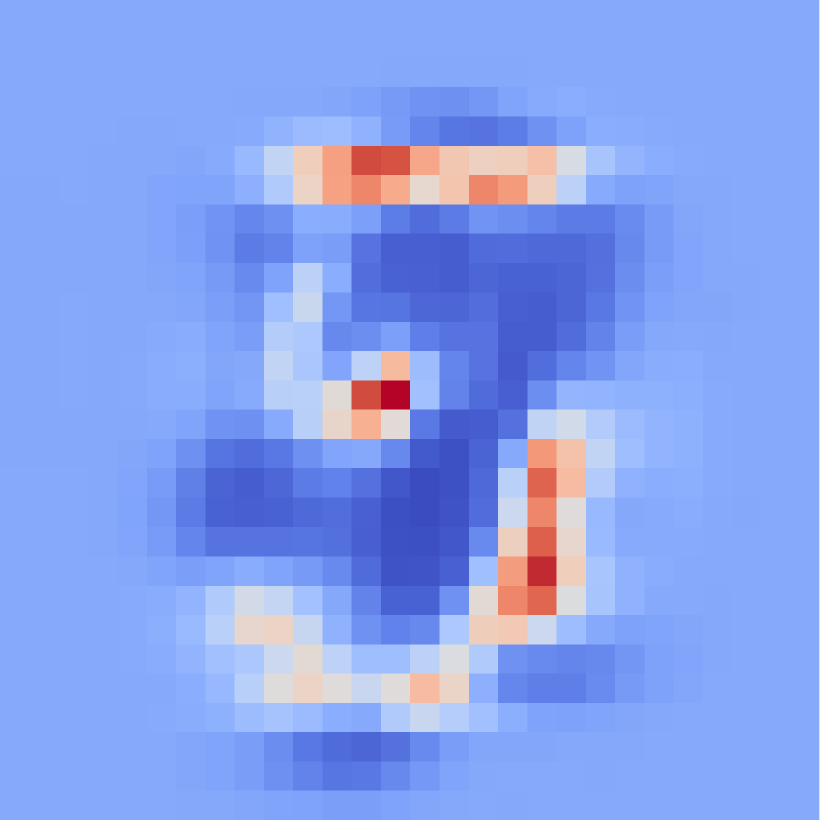} \\[3pt]

        \includegraphics[width=0.115\textwidth]{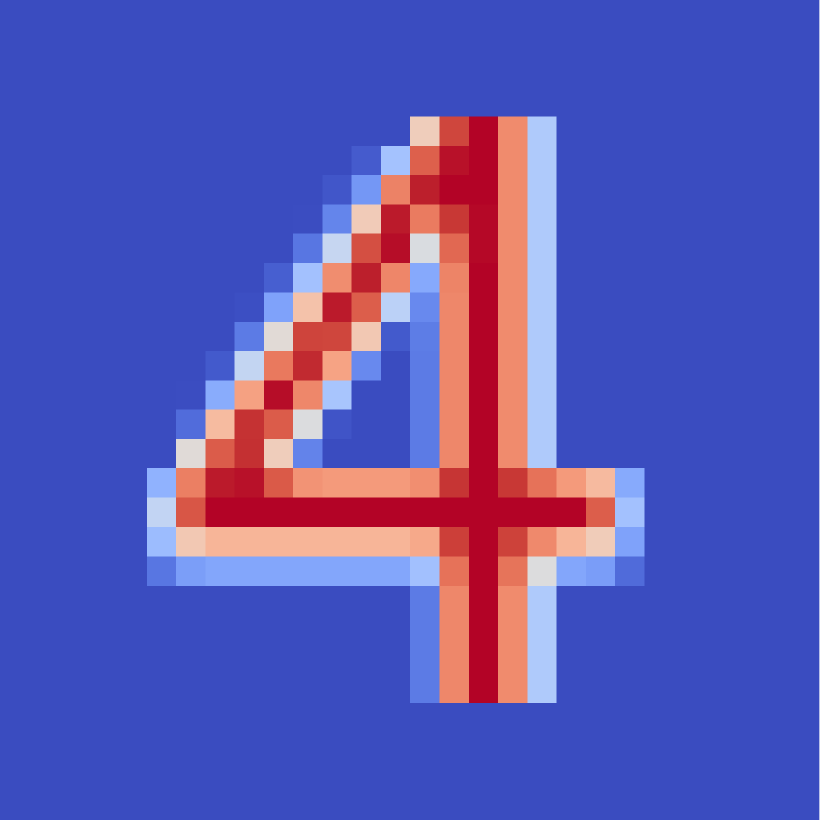} &
        \includegraphics[width=0.115\textwidth]{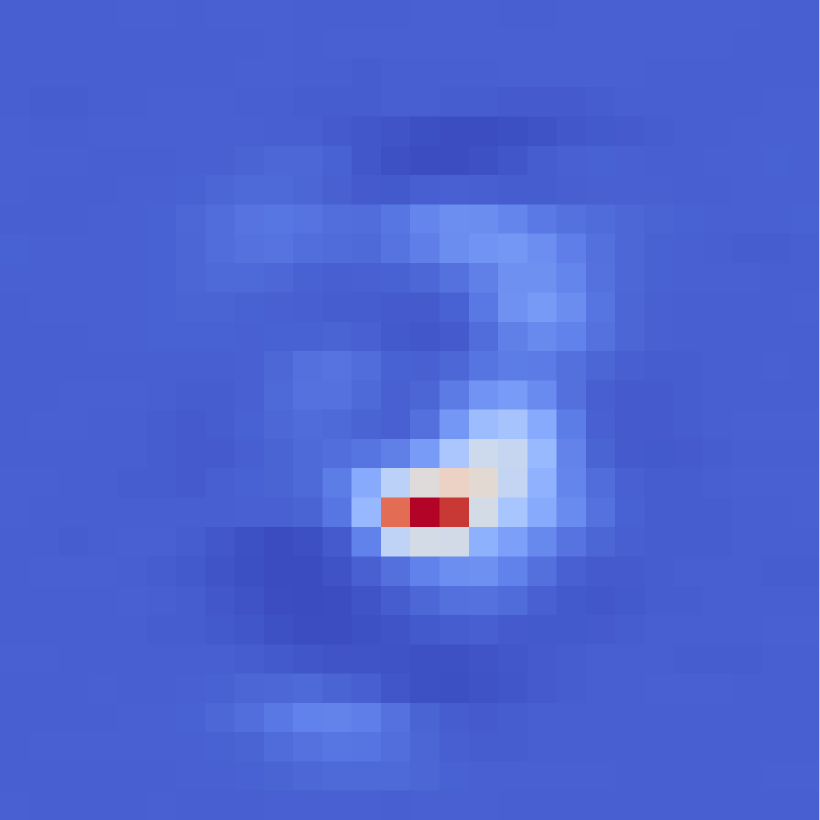} &
        \includegraphics[width=0.115\textwidth]{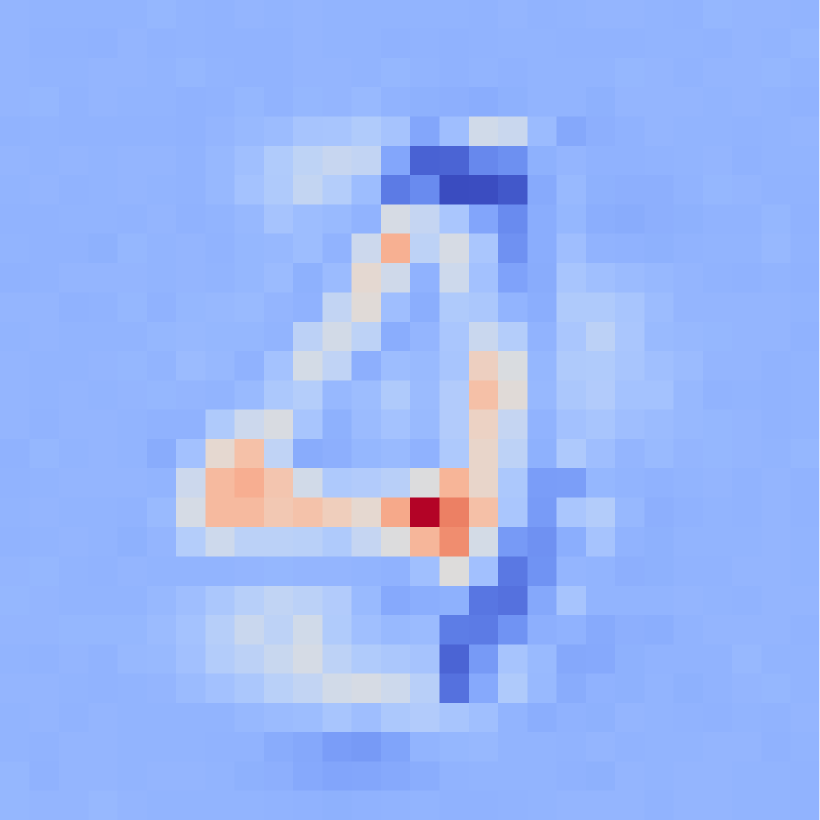} &
        \includegraphics[width=0.115\textwidth]{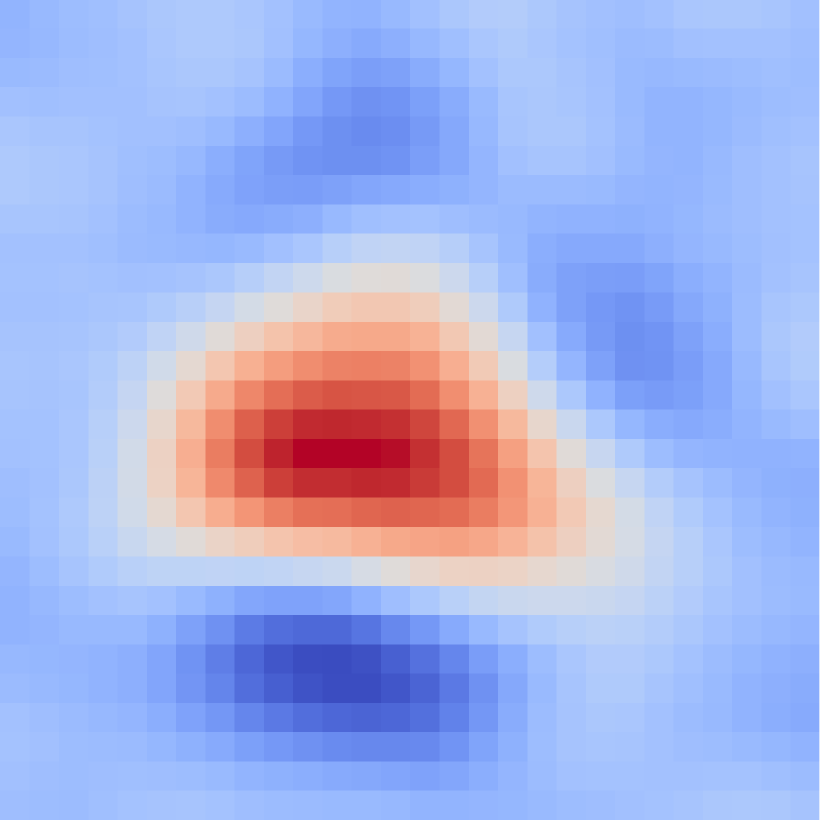} &
        \includegraphics[width=0.115\textwidth]{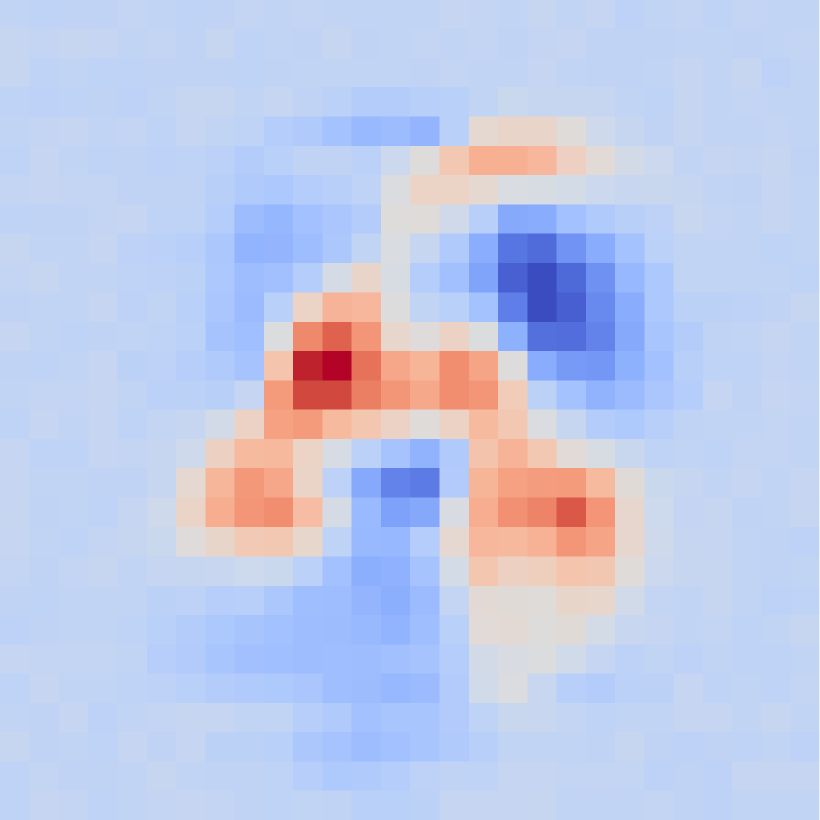} &
        \includegraphics[width=0.115\textwidth]{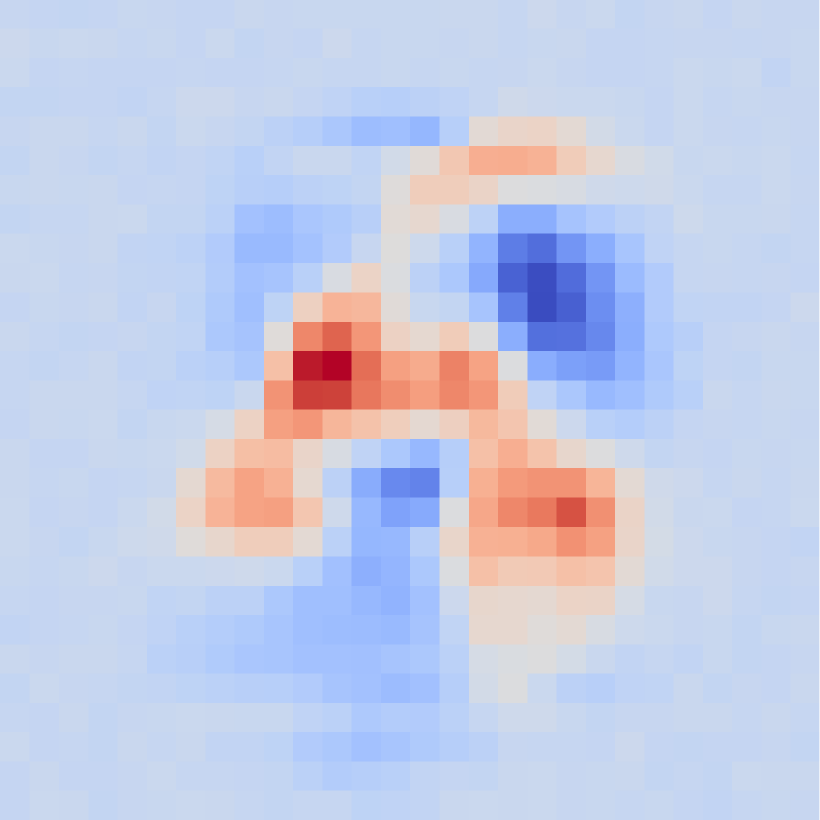} &
        \includegraphics[width=0.115\textwidth]{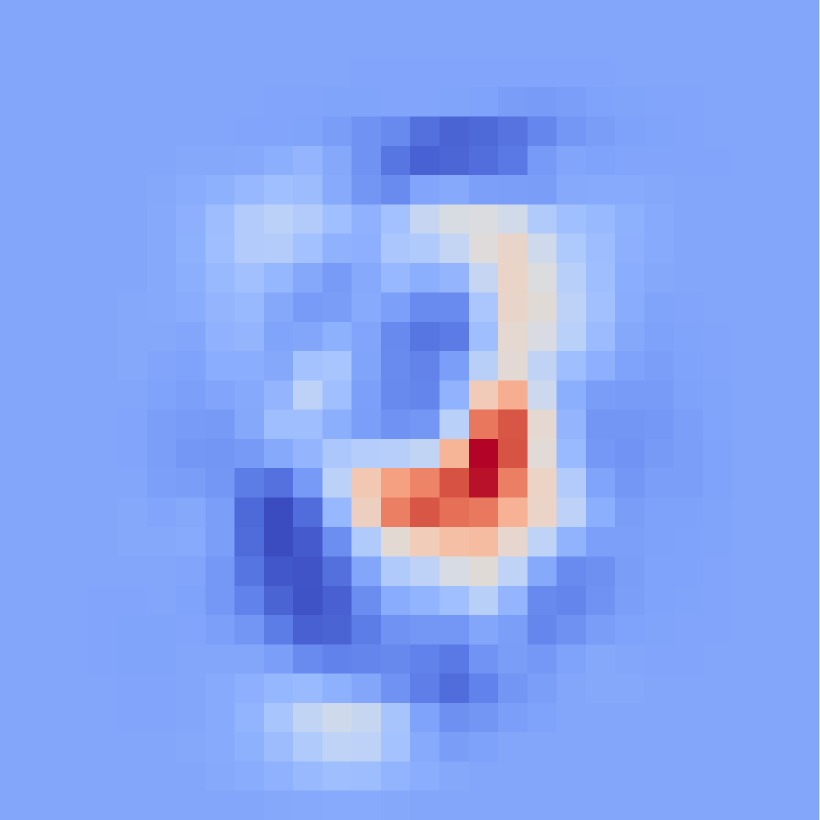} &
        \includegraphics[width=0.115\textwidth]{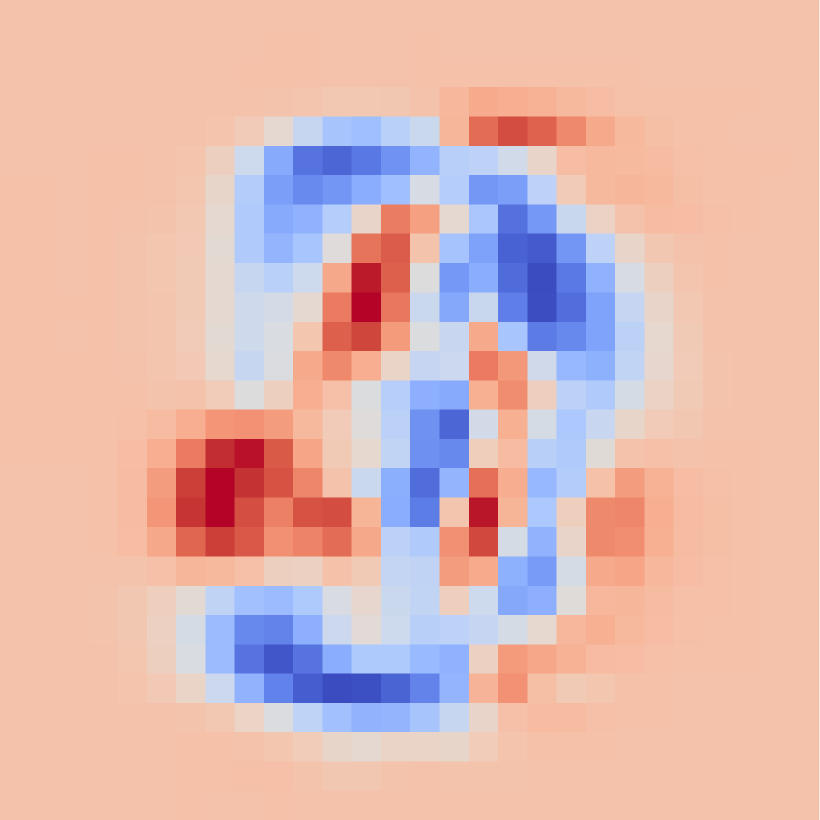} \\[3pt]

        \includegraphics[width=0.115\textwidth]{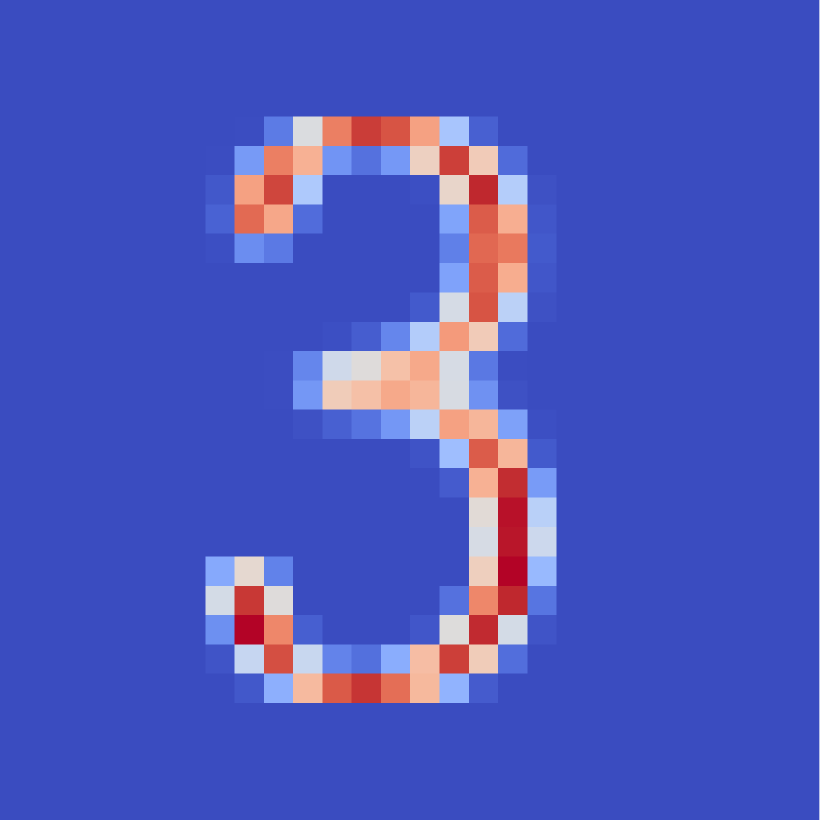} &
        \includegraphics[width=0.115\textwidth]{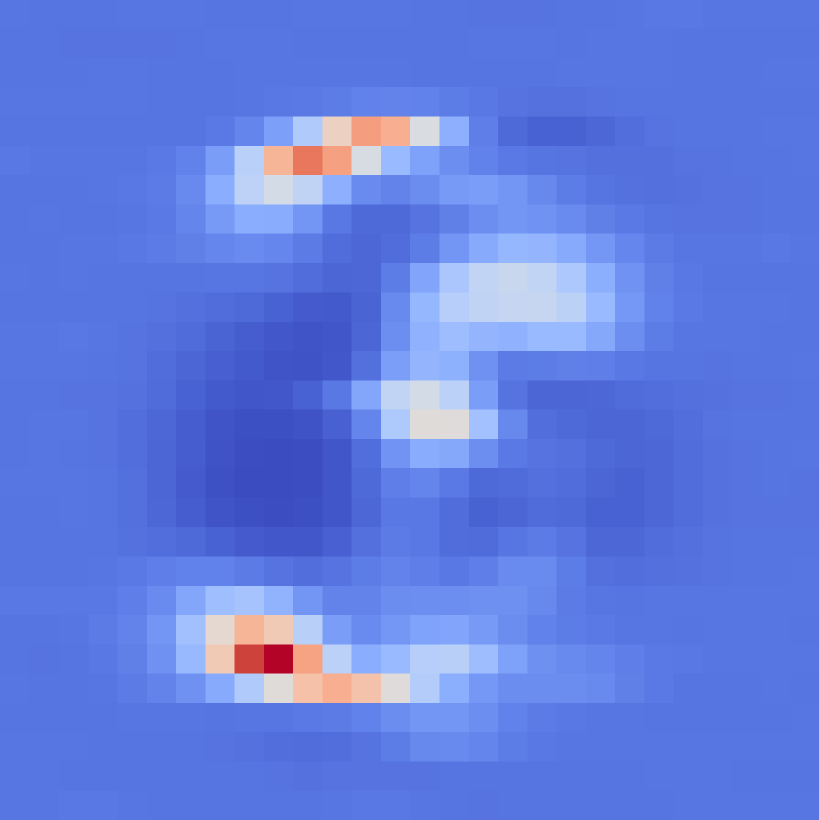} &
        \includegraphics[width=0.115\textwidth]{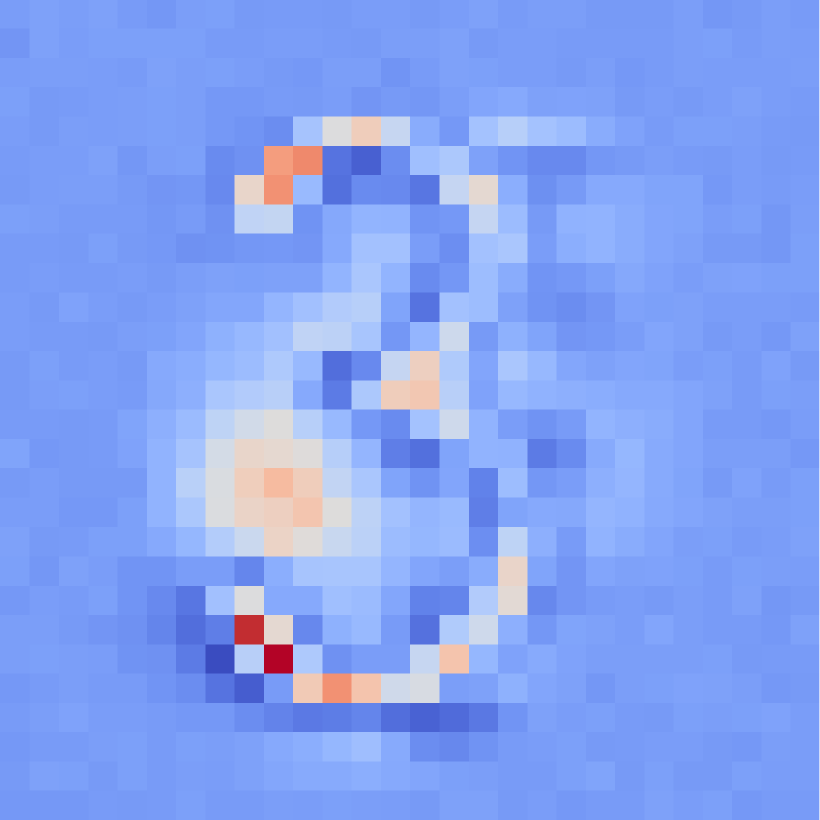} &
        \includegraphics[width=0.115\textwidth]{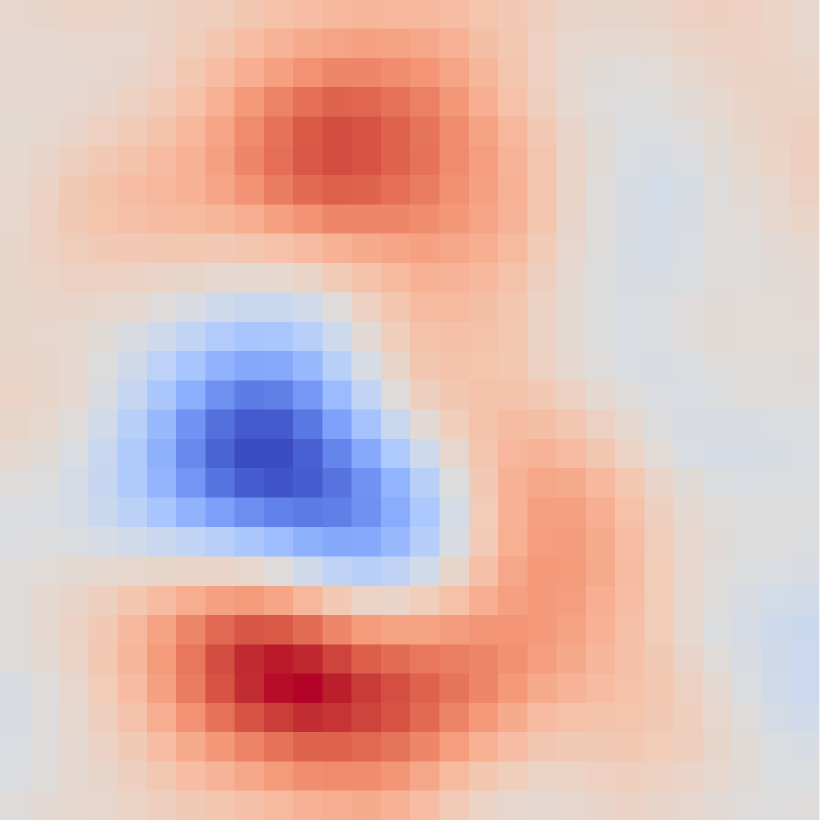} &
        \includegraphics[width=0.115\textwidth]{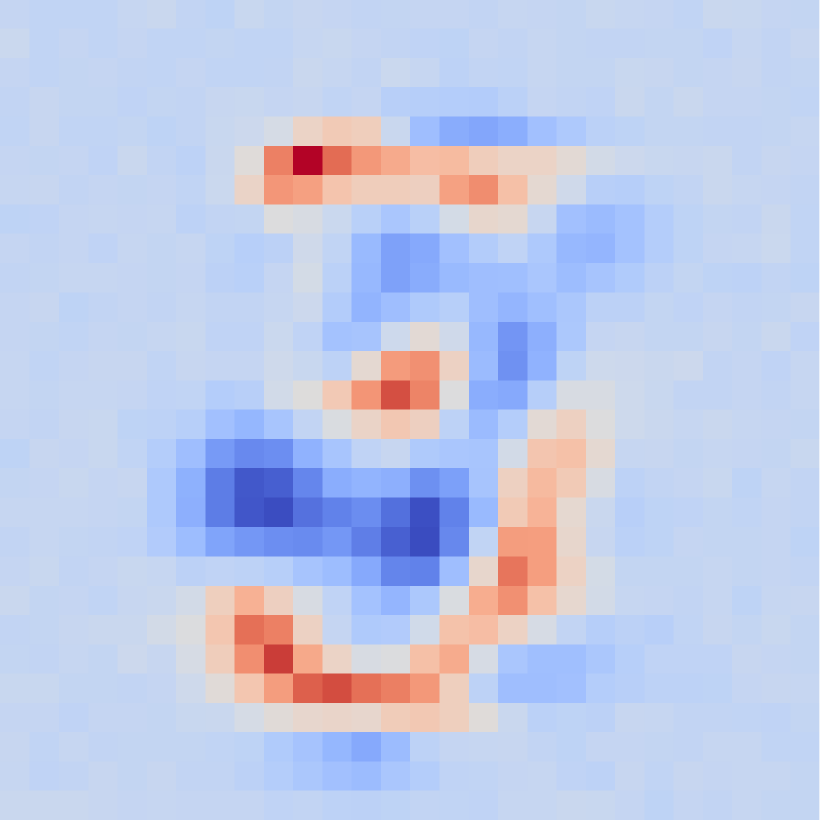} &
        \includegraphics[width=0.115\textwidth]{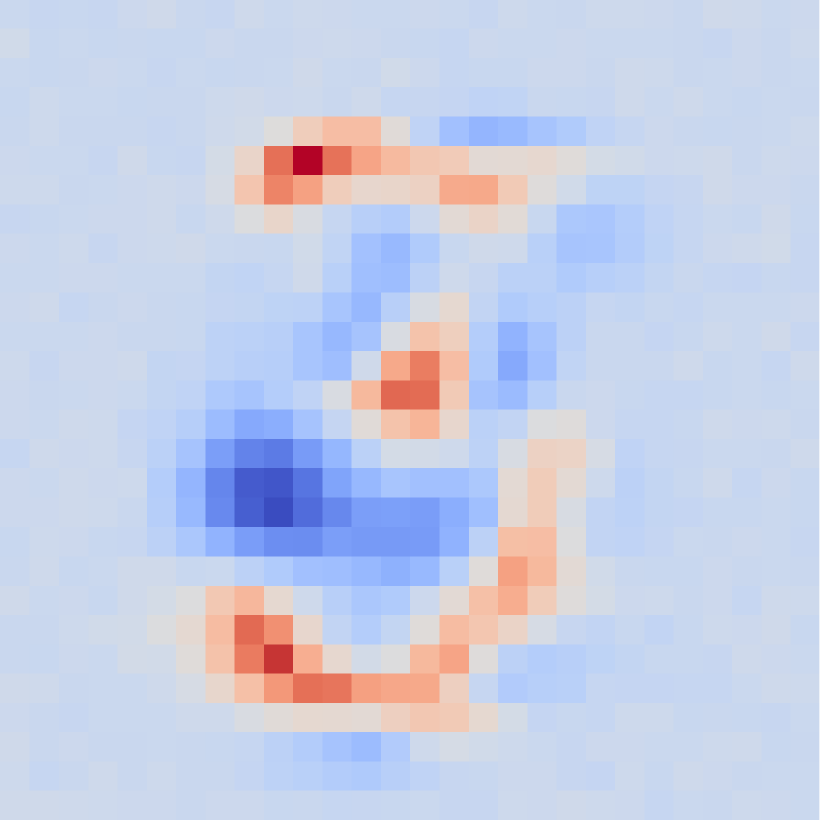} &
        \includegraphics[width=0.115\textwidth]{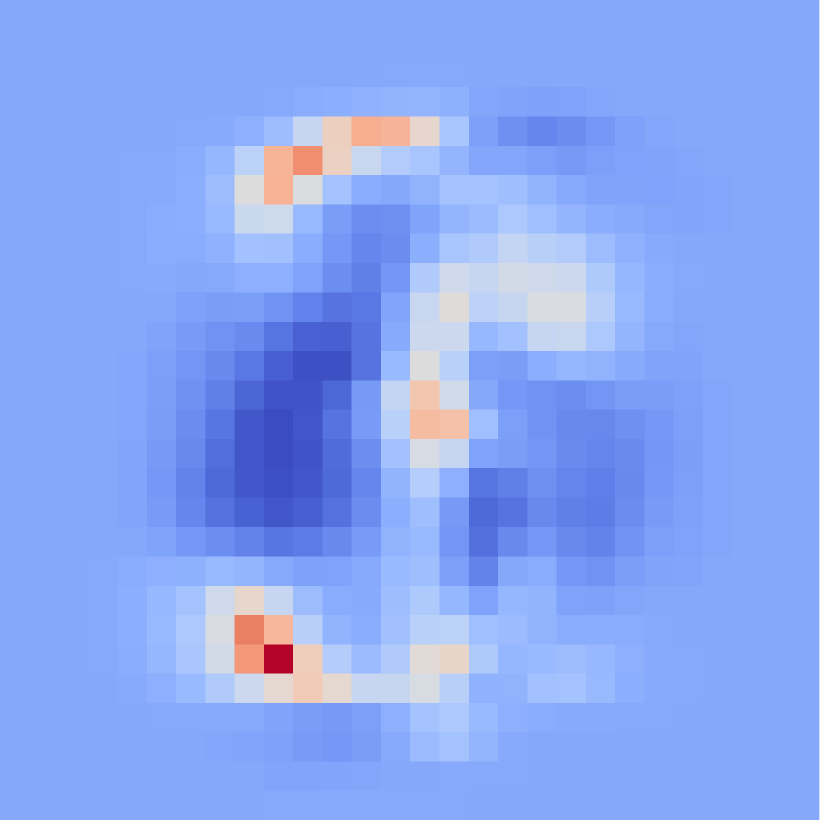} &
        \includegraphics[width=0.115\textwidth]{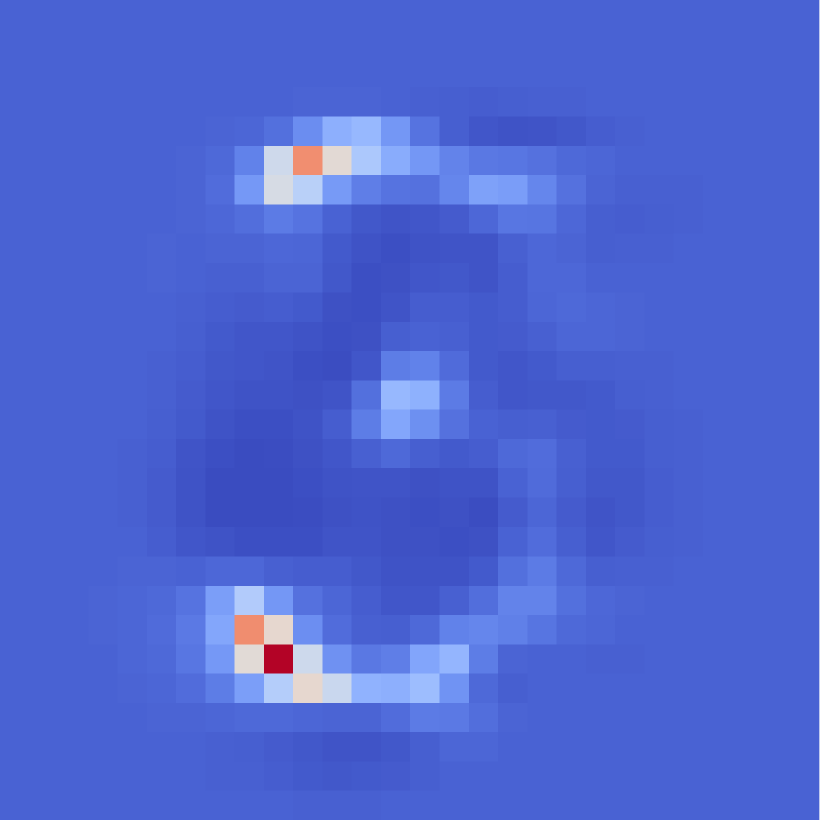} \\[3pt]
       
        \includegraphics[width=0.115\textwidth]{Experiments/typeface_mnist/sample_10.pdf} &
        \includegraphics[width=0.115\textwidth]{Experiments/typeface_mnist/OCCL_10.pdf} &
        \includegraphics[width=0.115\textwidth]{Experiments/typeface_mnist/SHAP_10.pdf} &
        \includegraphics[width=0.115\textwidth]{Experiments/typeface_mnist/RISE_10.pdf} &
        \includegraphics[width=0.115\textwidth]{Experiments/typeface_mnist/IG_10.pdf} &
        \includegraphics[width=0.115\textwidth]{Experiments/typeface_mnist/LRP_10.pdf} &
        \includegraphics[width=0.115\textwidth]{Experiments/typeface_mnist/XWP_10.pdf} &
        \includegraphics[width=0.115\textwidth]{Experiments/typeface_mnist/XWP_C_10.pdf} \\[3pt]
    \end{tabular}

    \caption{The importance scores of different attribution methods for the Typeface MNIST datasets. XWP\textsuperscript{c}
    recognizes the input signal with greater confidence XWP, while the latter produces more focused heatmaps across different digits.}
    \label{fig:tmnist}
    \Description{A grid of 9x8 importance scores of different attribution methods for the Typeface MNIST datasets. Starting
    from the leftmost column, the image sample is presented, followed by different attribution methods: Occlusion, Shapley Values, Integrated Gradients, 
    LRP, XWP and XWP\textsuperscript{c}. Heatmaps indicate that XWP\textsuperscript{c} recognizes the input signal with greater confidence XWP, while 
    the latter produces more focused heatmaps across different digits.}

\end{figure*}
\begin{figure*}[t]
    \centering
    \setlength{\tabcolsep}{2pt}  

    \begin{tabular}{c c c c c c c c}
        \textbf{Sample} & \textbf{OCCL} & \textbf{SHAP} & \textbf{RISE} & \textbf{IG} &
        \textbf{LRP} & \textbf{XWP} & \textbf{XWP\textsuperscript{c}} \\[3pt]

        \includegraphics[width=0.115\textwidth]{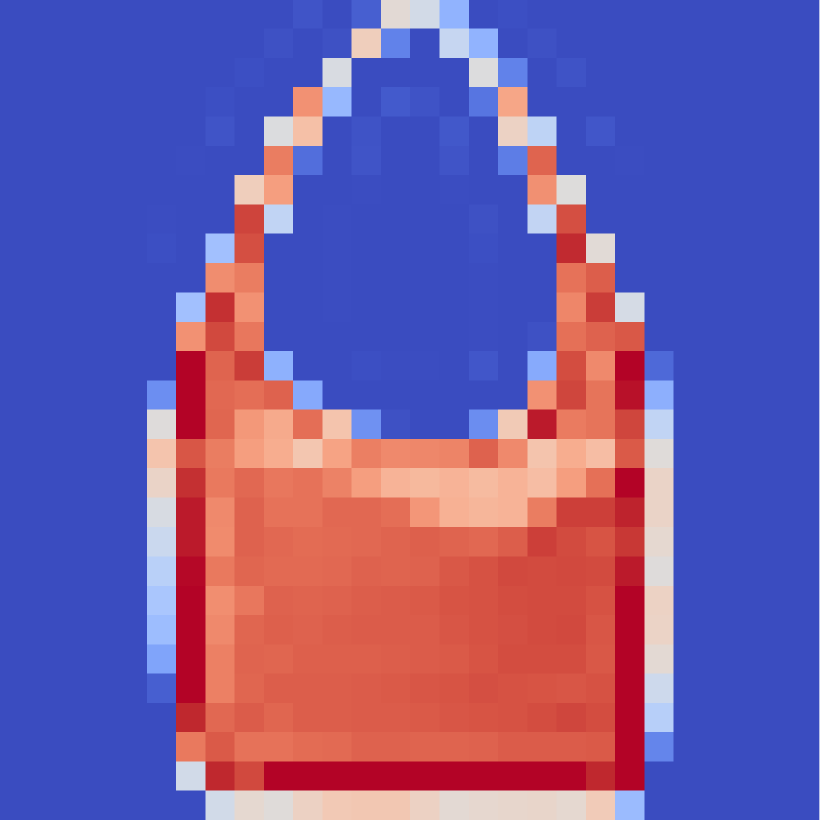} &
        \includegraphics[width=0.115\textwidth]{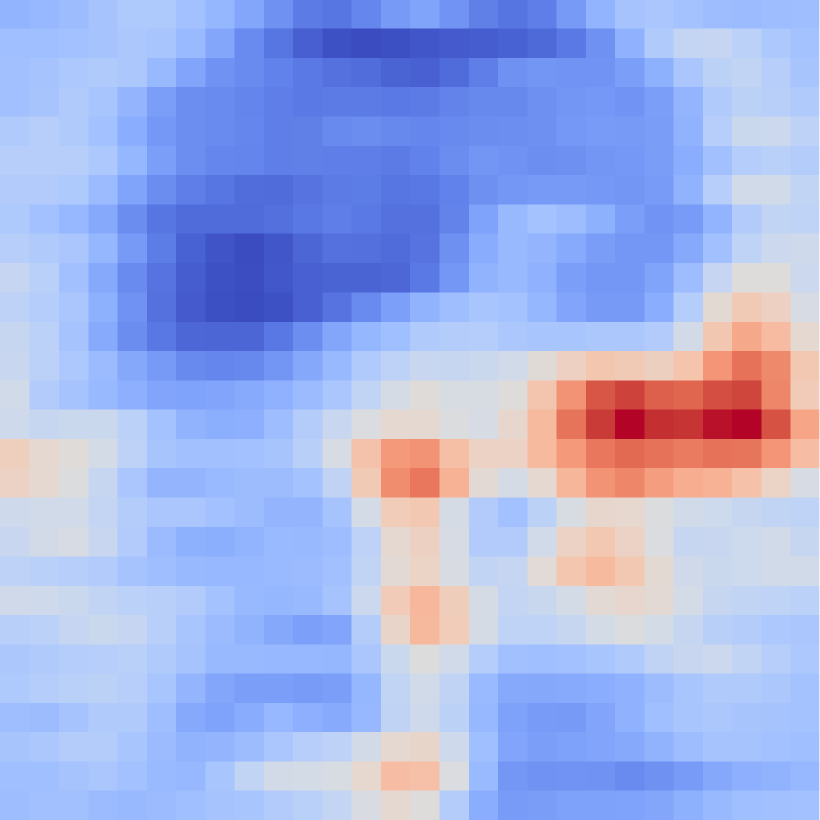} &
        \includegraphics[width=0.115\textwidth]{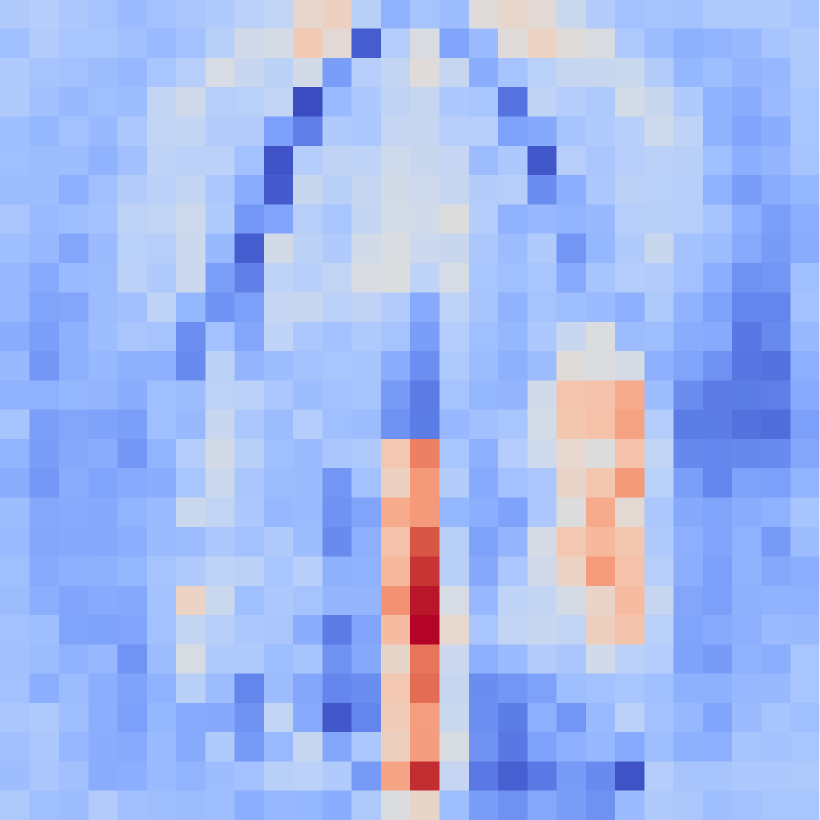} &
        \includegraphics[width=0.115\textwidth]{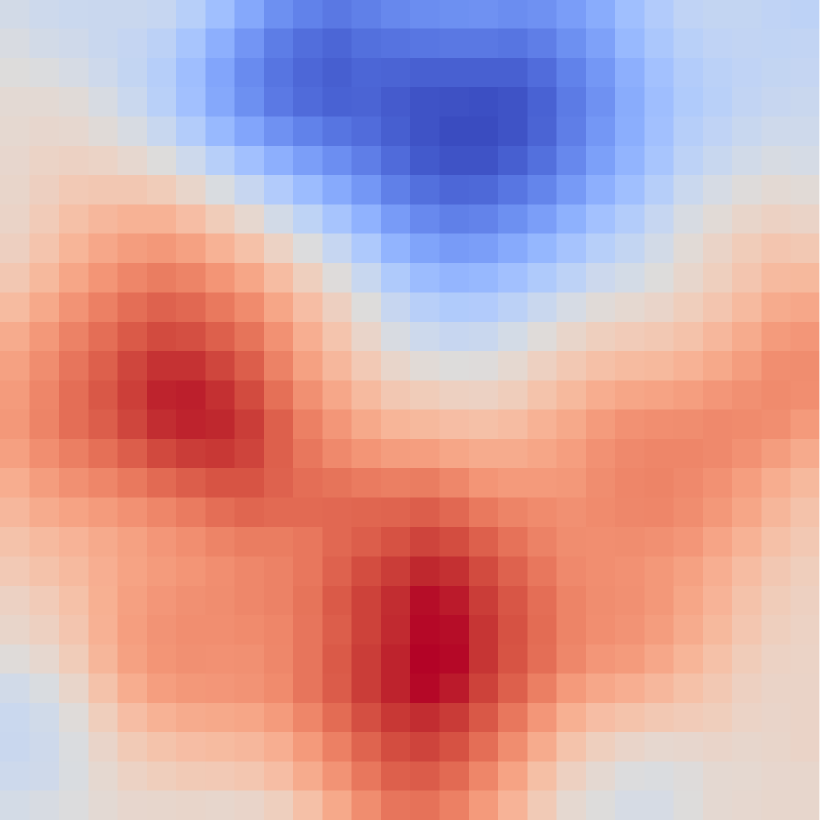} &
        \includegraphics[width=0.115\textwidth]{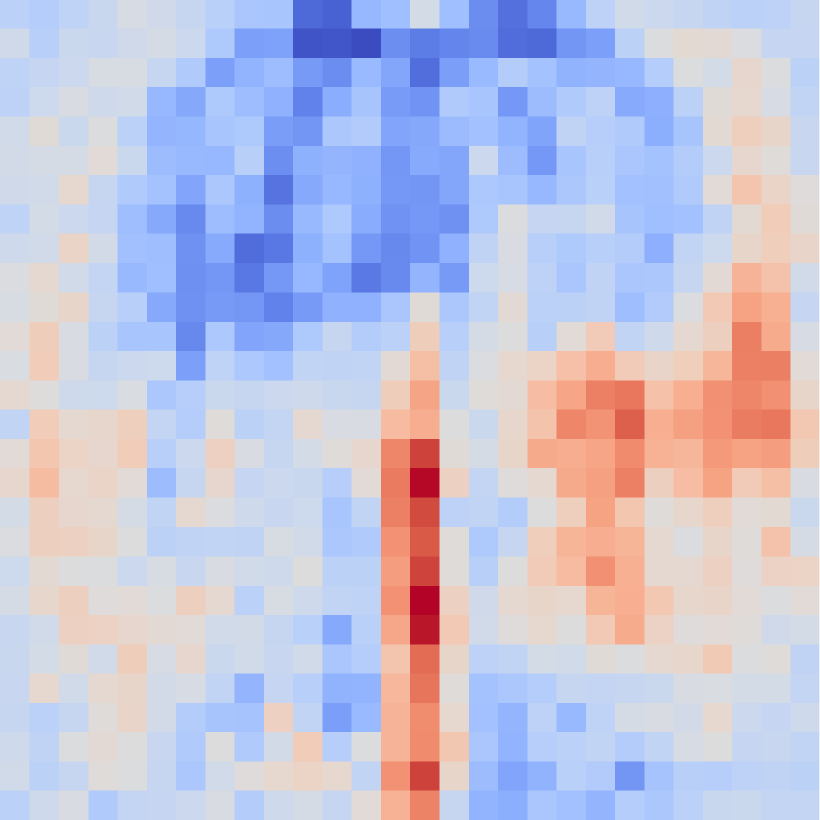} &
        \includegraphics[width=0.115\textwidth]{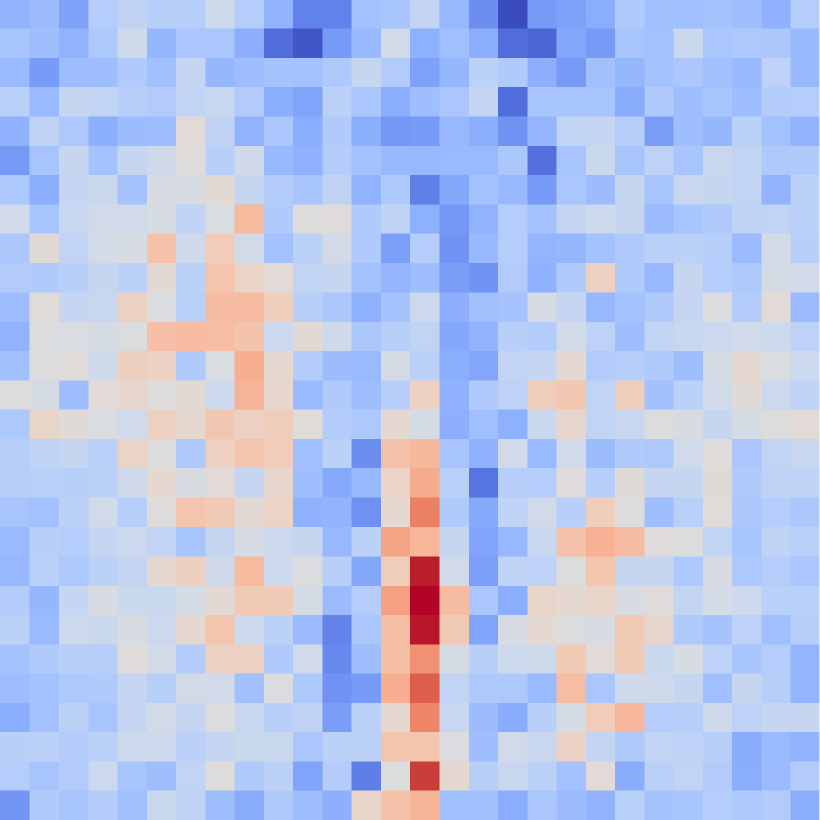} &
        \includegraphics[width=0.115\textwidth]{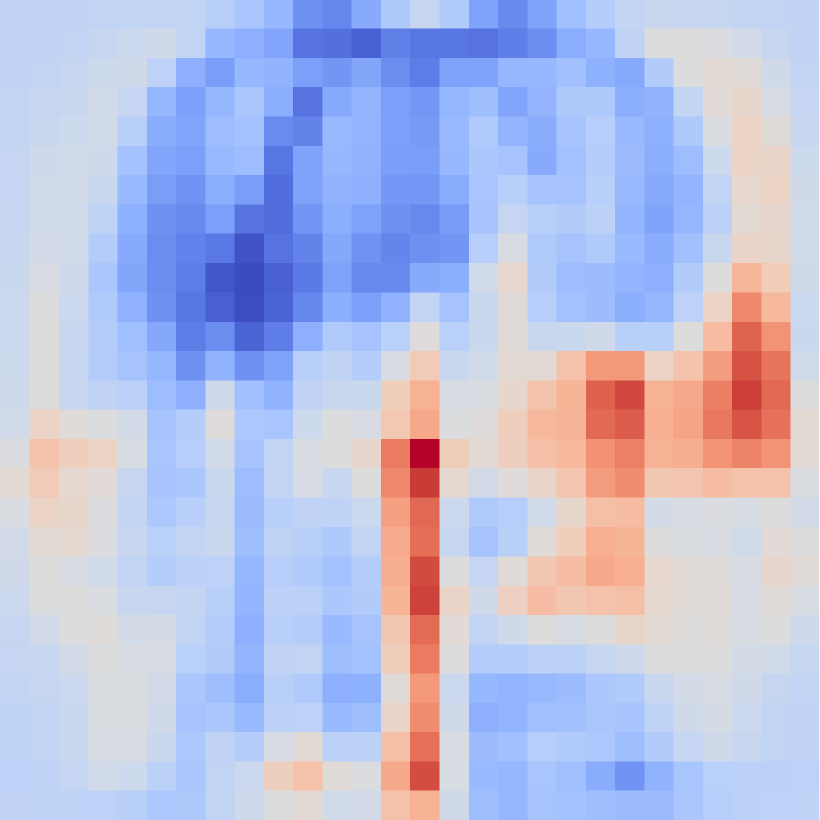} &
        \includegraphics[width=0.115\textwidth]{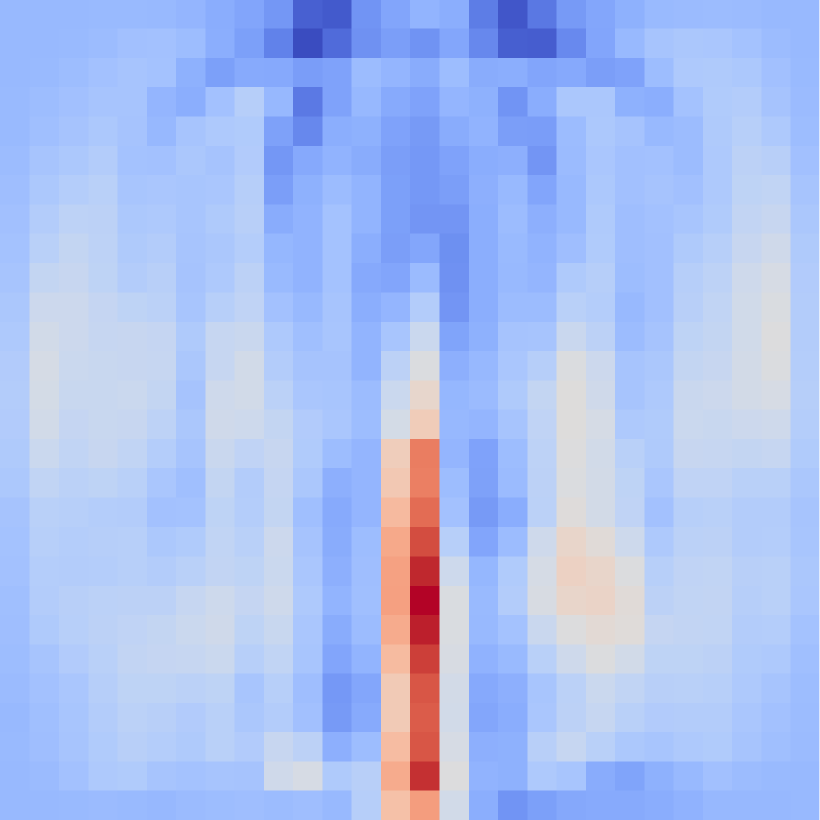} \\[3pt]

        \includegraphics[width=0.115\textwidth]{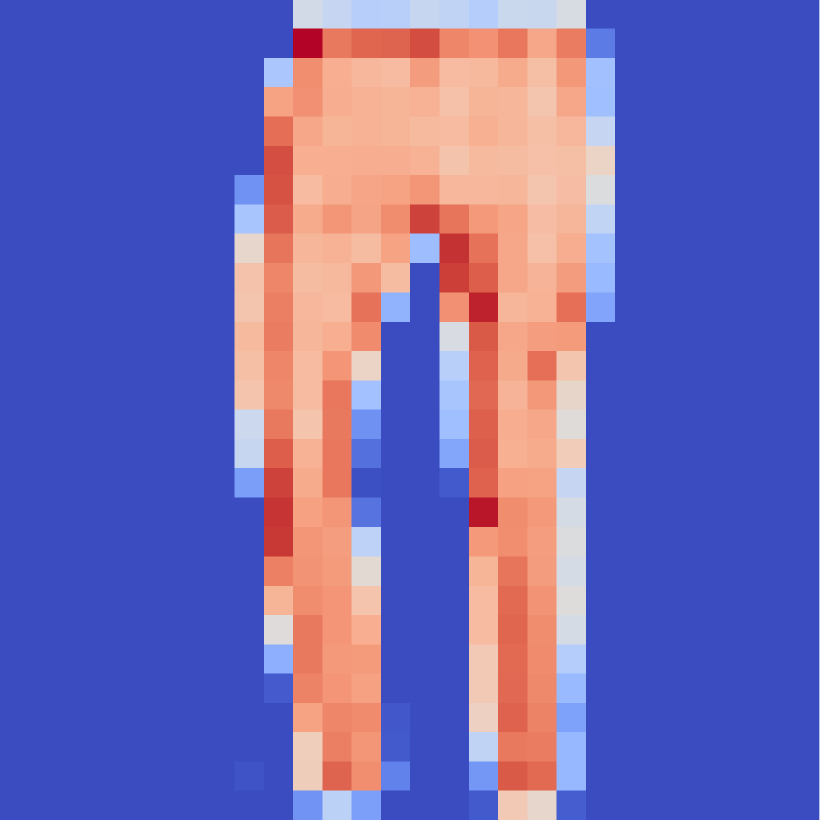} &
        \includegraphics[width=0.115\textwidth]{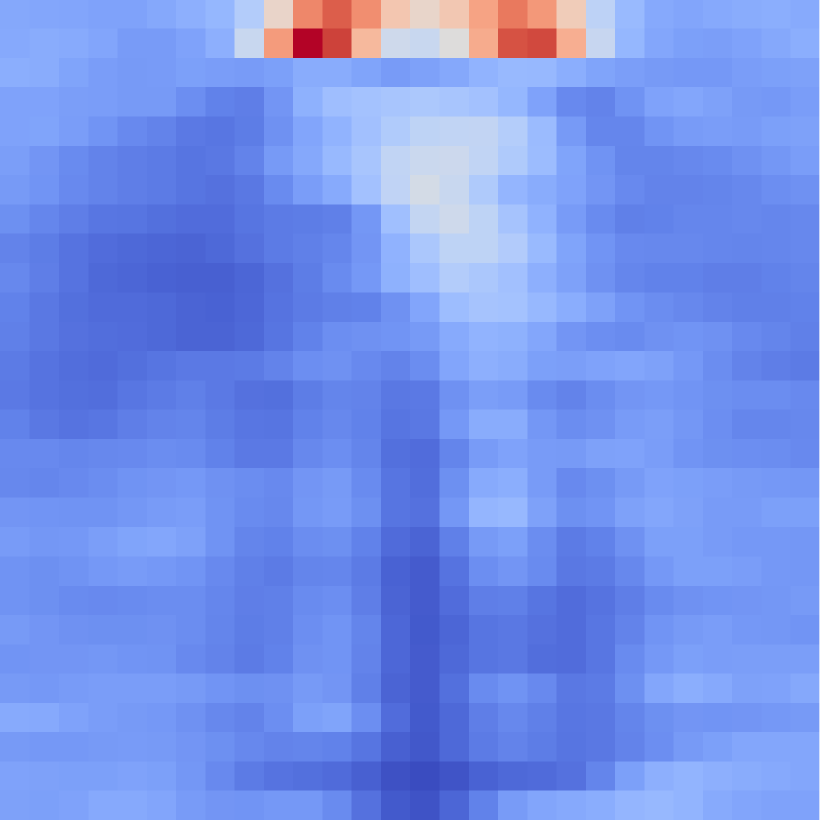} &
        \includegraphics[width=0.115\textwidth]{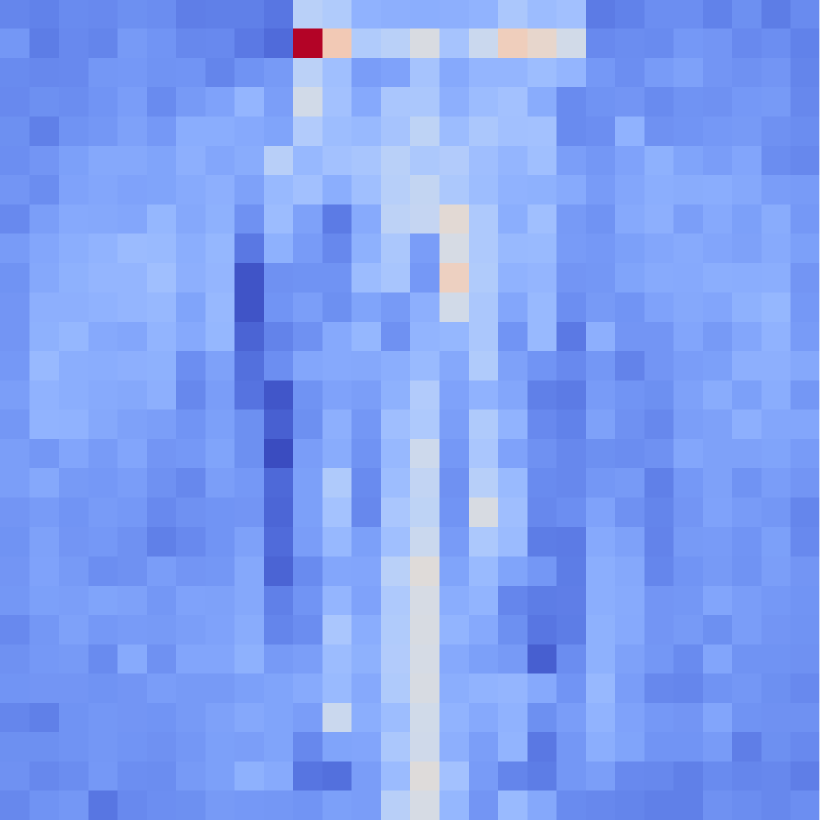} &
        \includegraphics[width=0.115\textwidth]{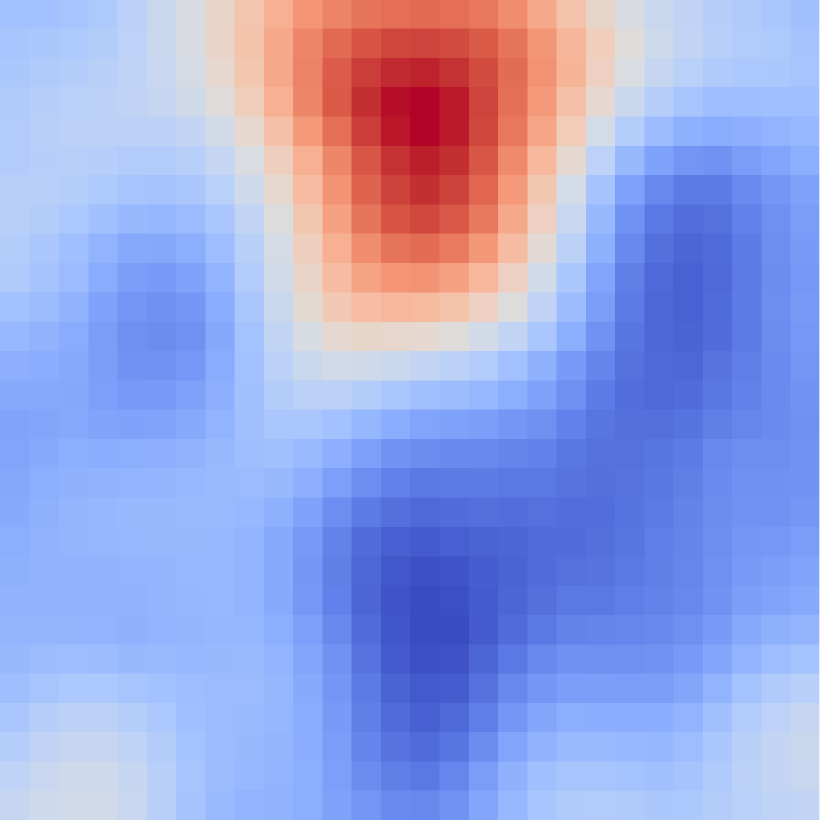} &
        \includegraphics[width=0.115\textwidth]{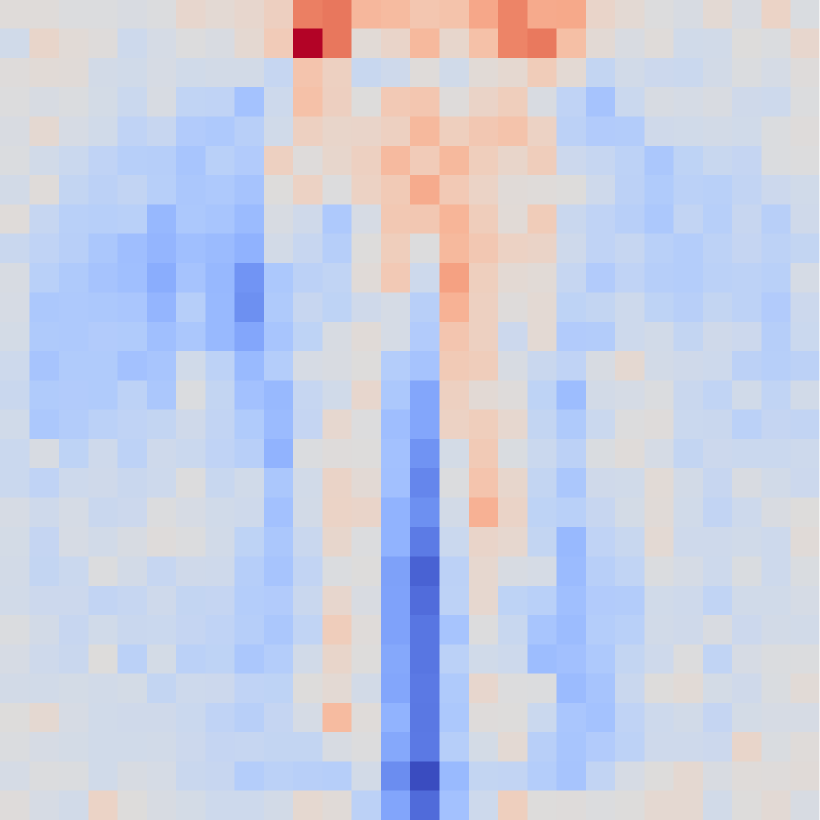} &
        \includegraphics[width=0.115\textwidth]{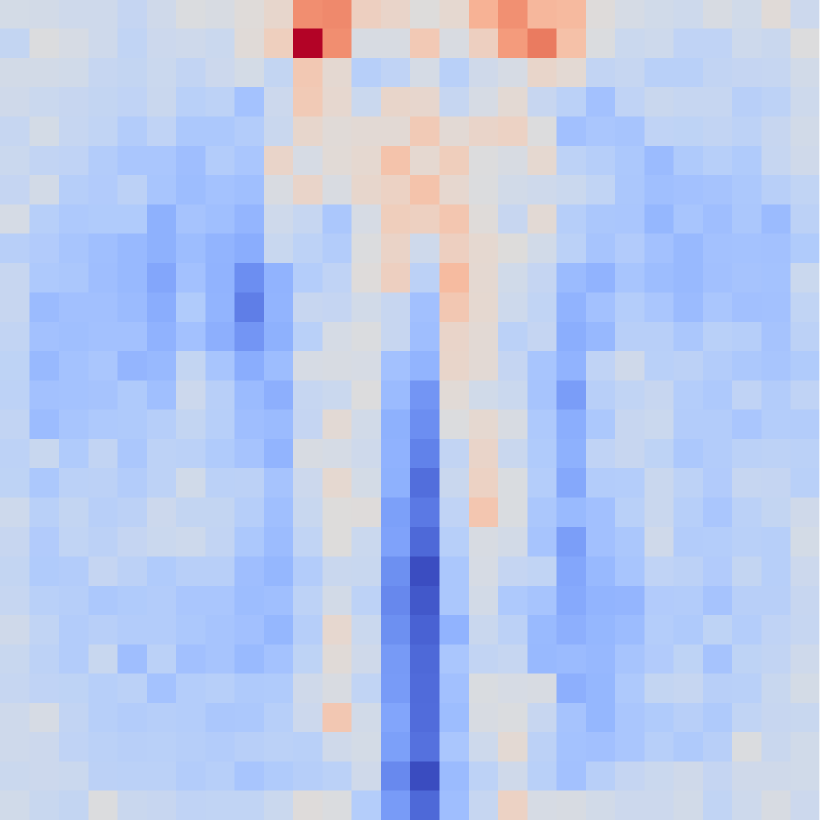} &
        \includegraphics[width=0.115\textwidth]{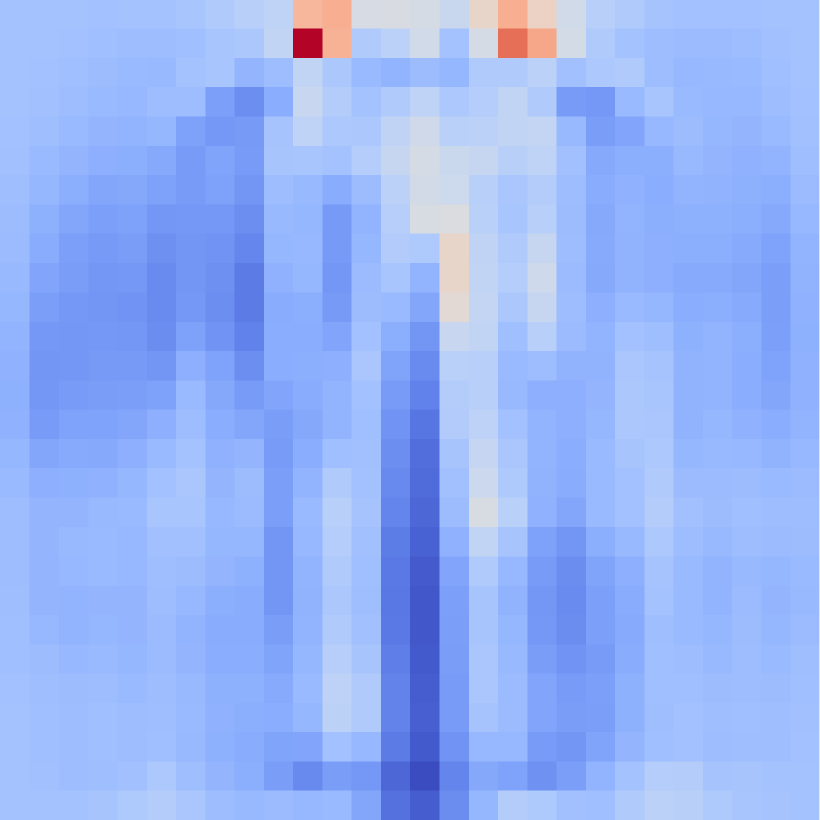} &
        \includegraphics[width=0.115\textwidth]{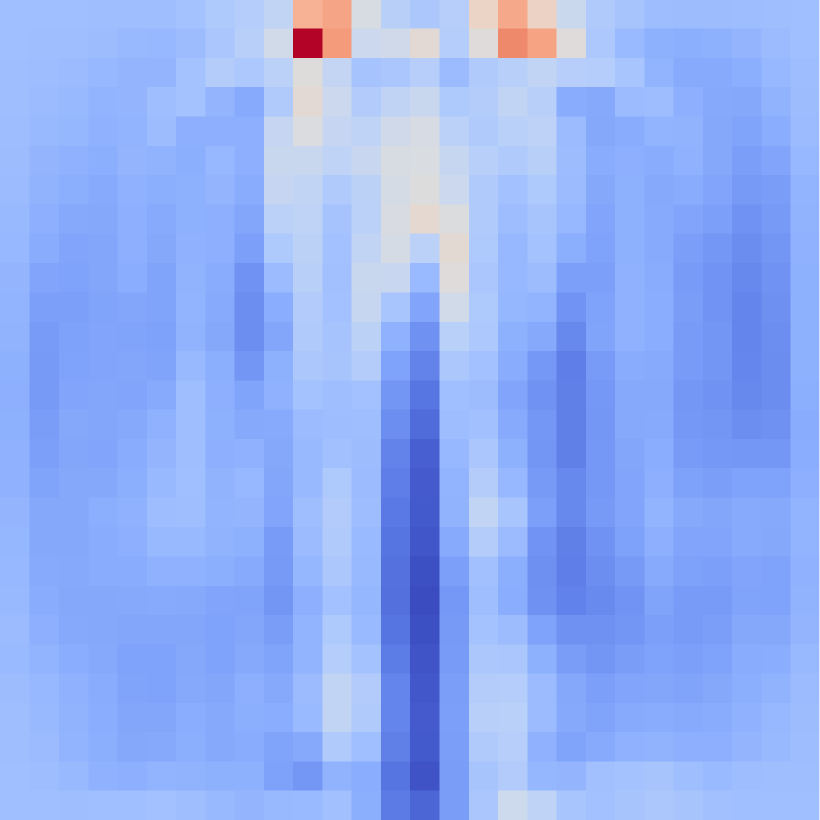} \\[3pt]

        \includegraphics[width=0.115\textwidth]{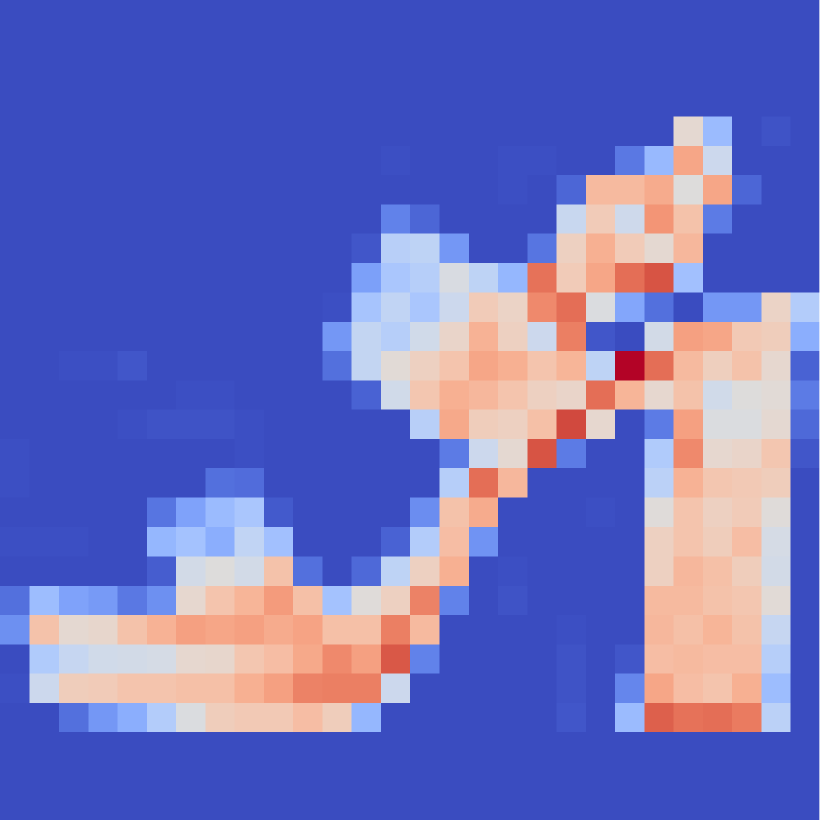} &
        \includegraphics[width=0.115\textwidth]{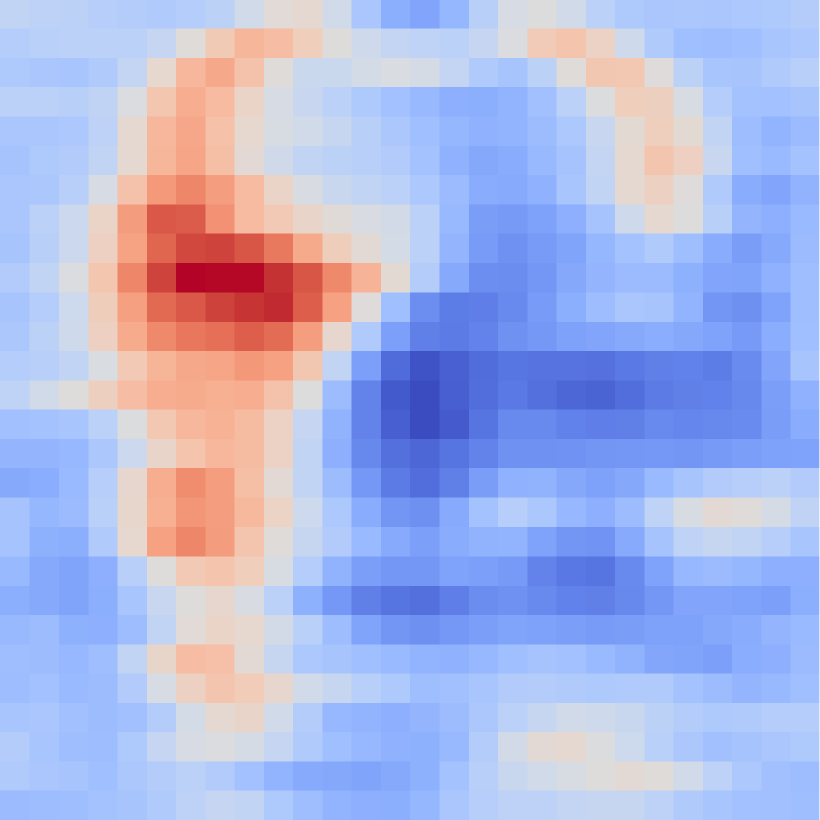} &
        \includegraphics[width=0.115\textwidth]{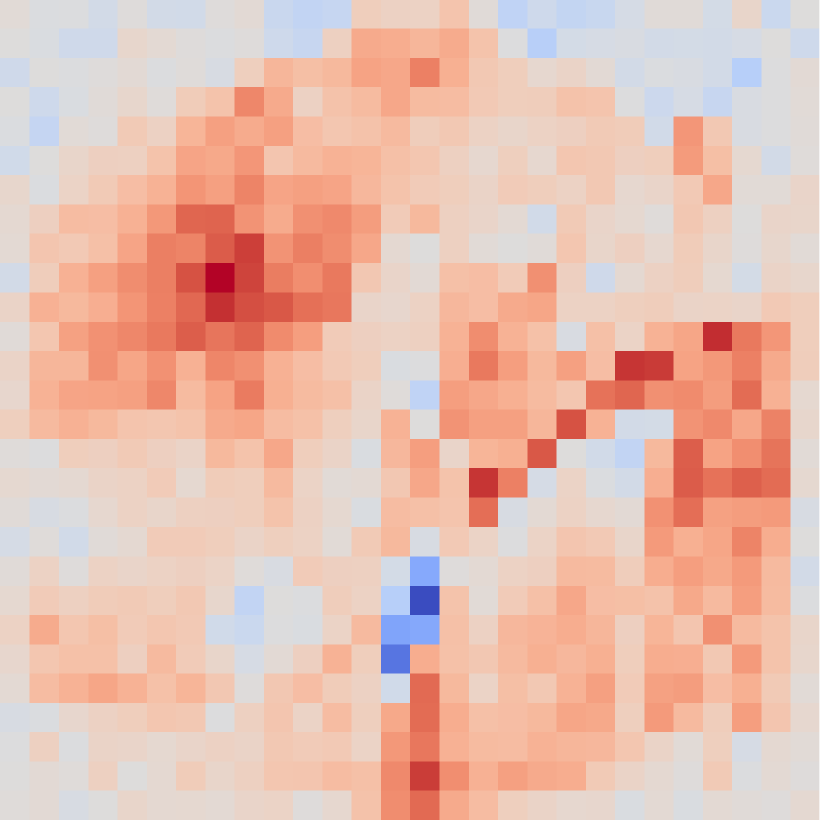} &
        \includegraphics[width=0.115\textwidth]{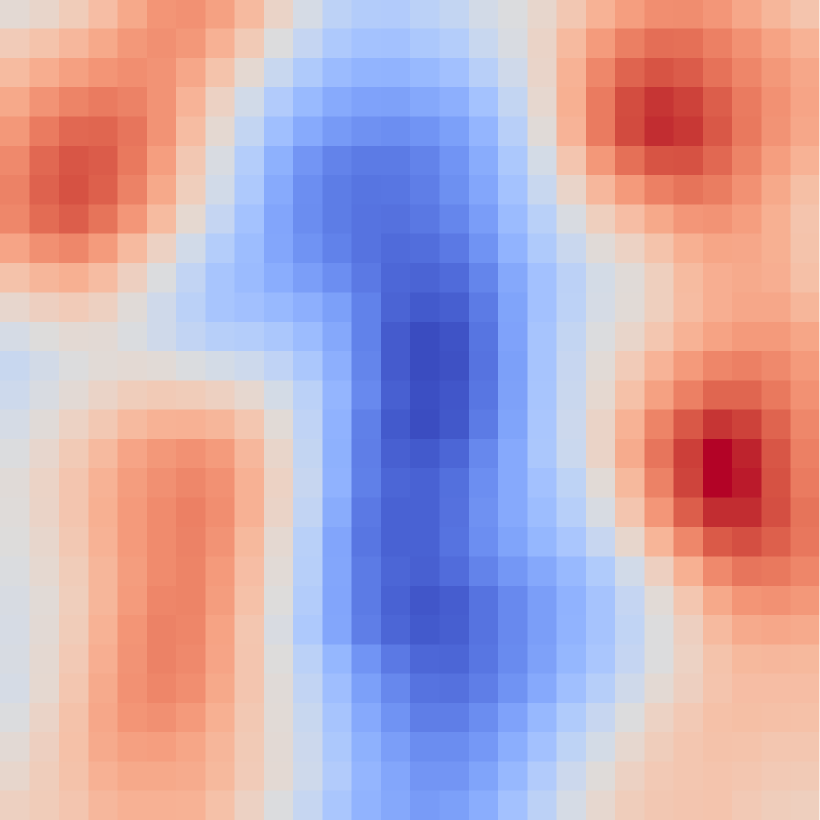} &
        \includegraphics[width=0.115\textwidth]{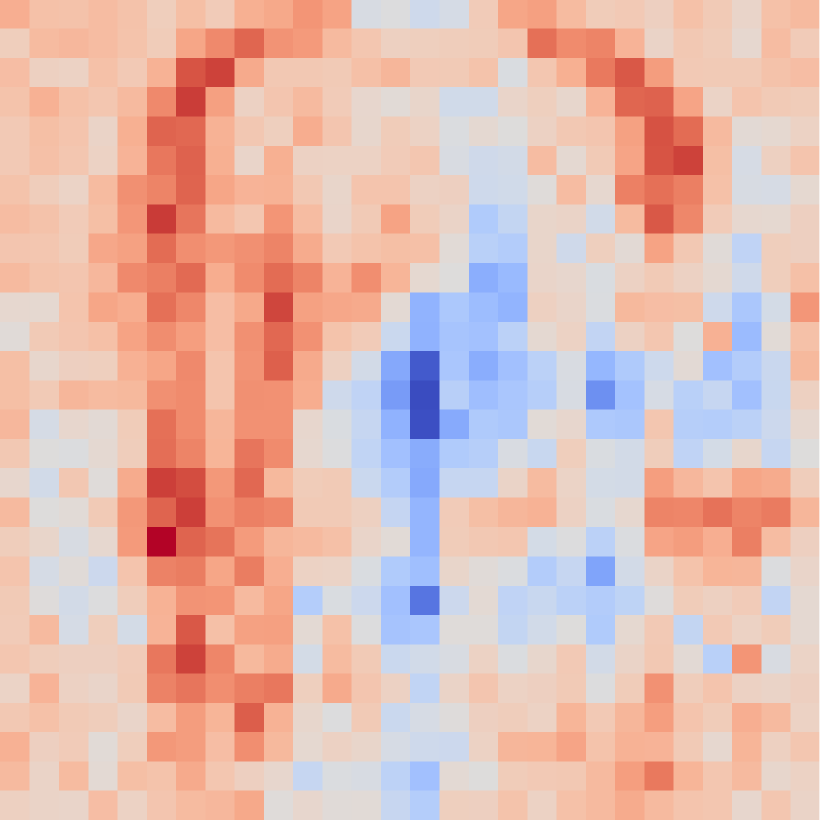} &
        \includegraphics[width=0.115\textwidth]{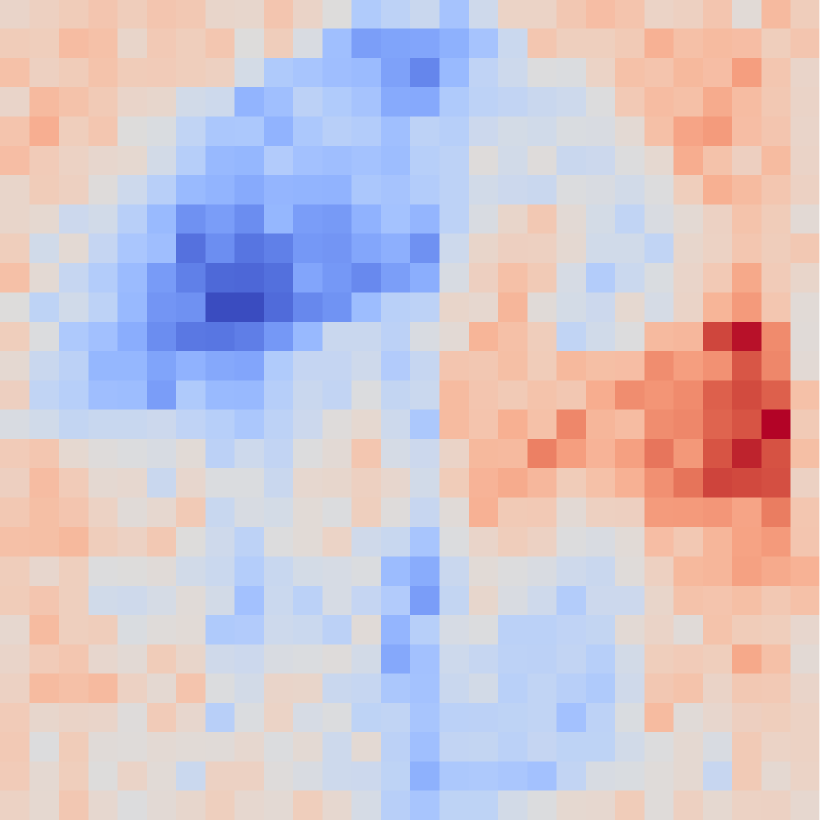} &
        \includegraphics[width=0.115\textwidth]{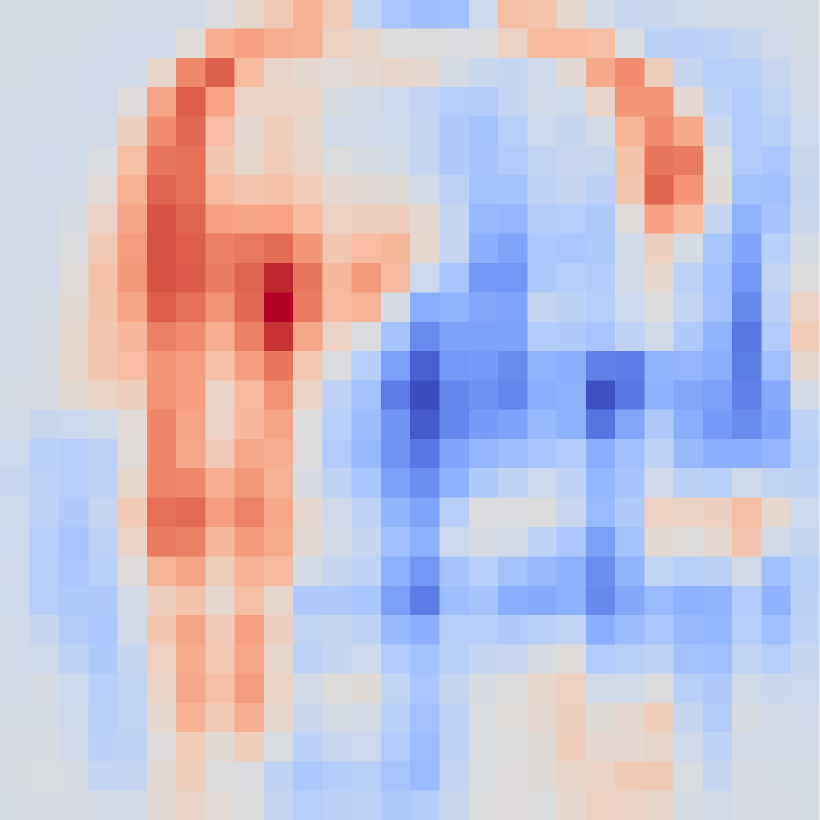} &
        \includegraphics[width=0.115\textwidth]{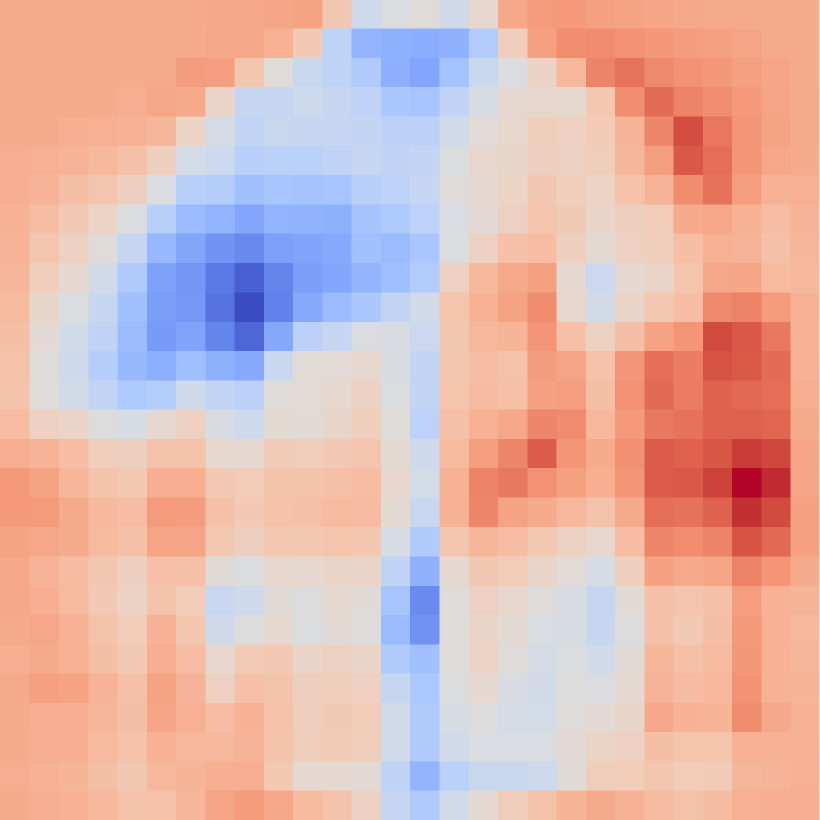} \\[3pt]

        \includegraphics[width=0.115\textwidth]{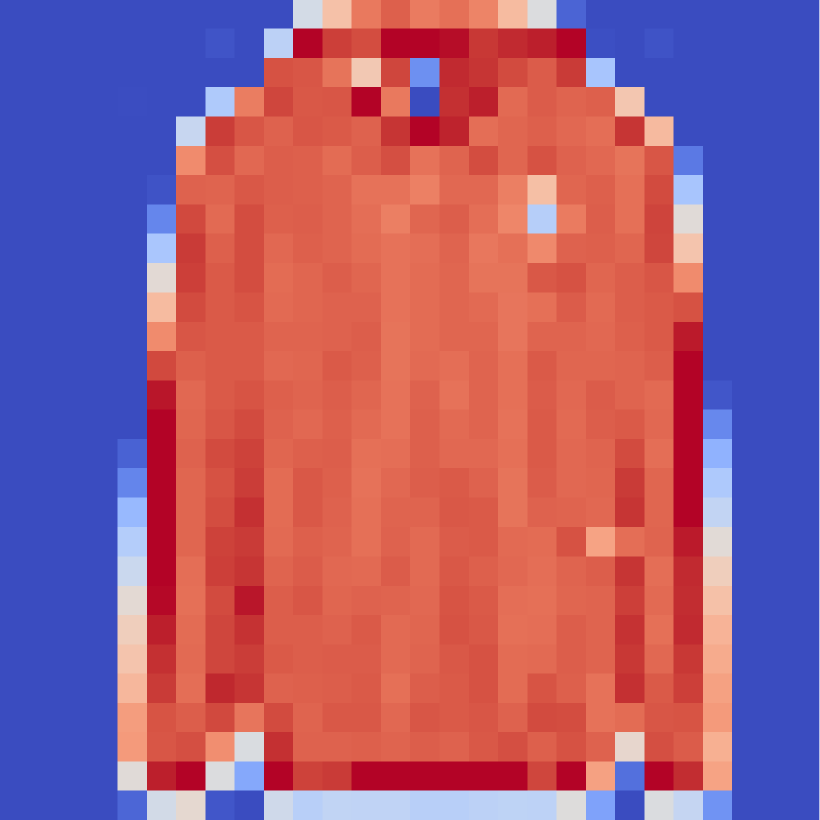} &
        \includegraphics[width=0.115\textwidth]{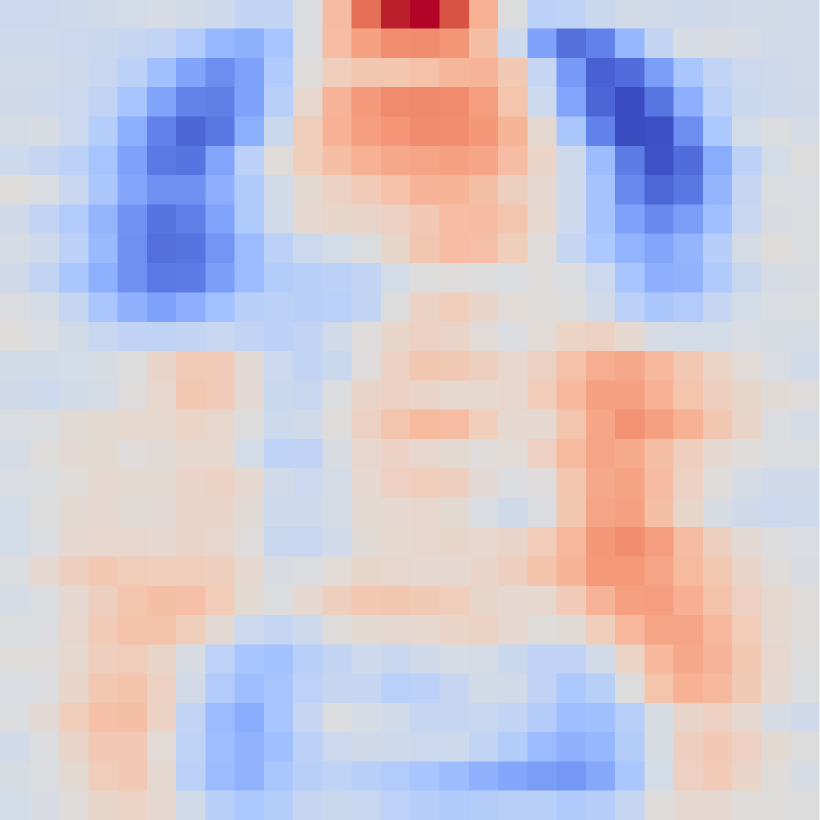} &
        \includegraphics[width=0.115\textwidth]{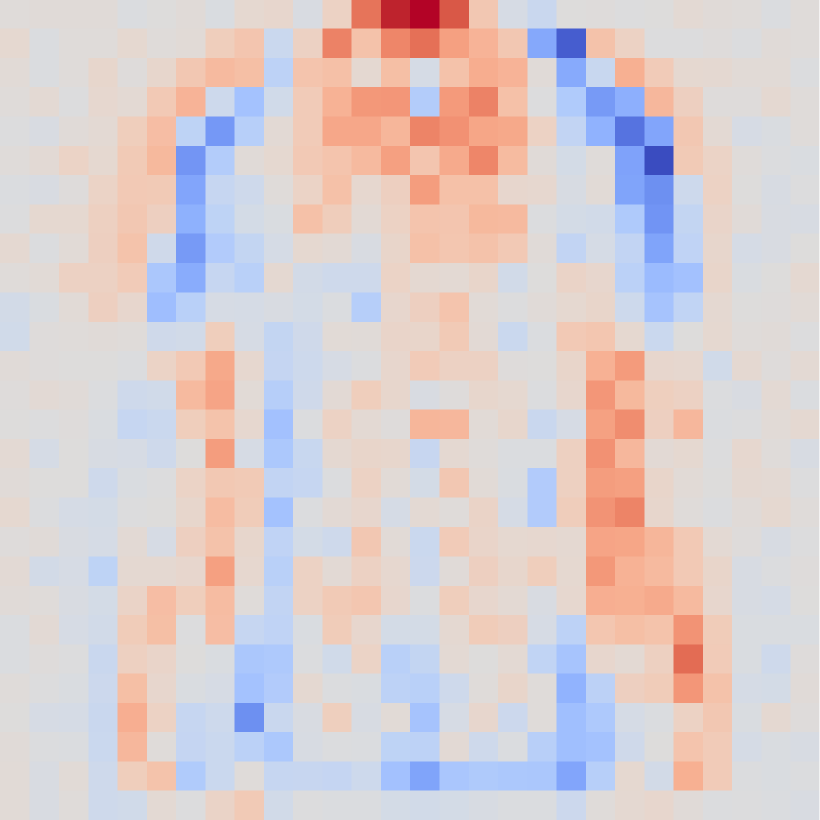} &
        \includegraphics[width=0.115\textwidth]{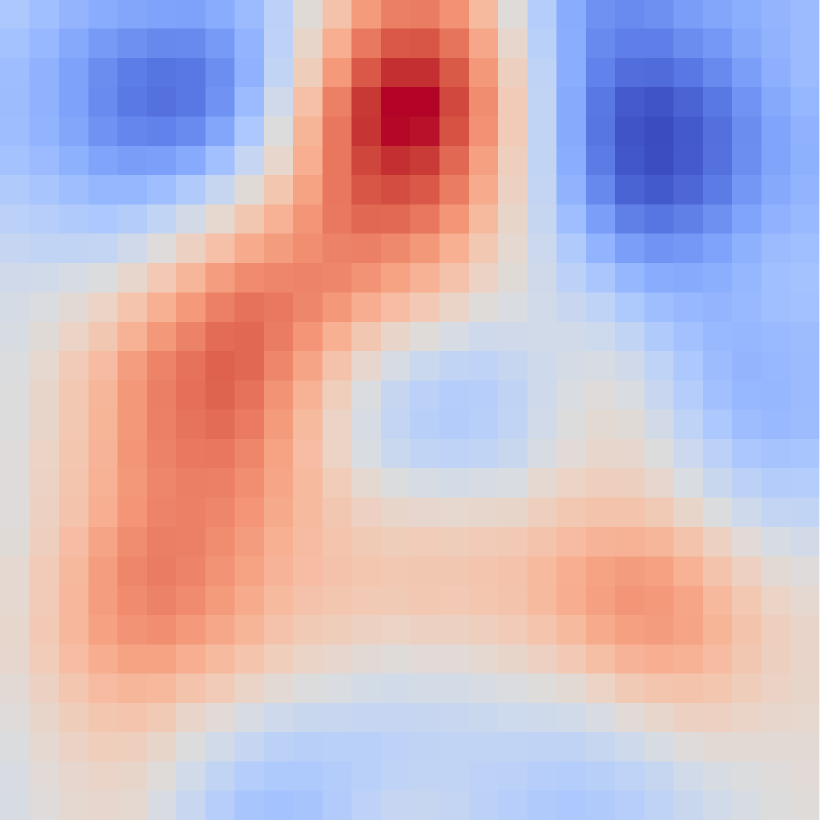} &
        \includegraphics[width=0.115\textwidth]{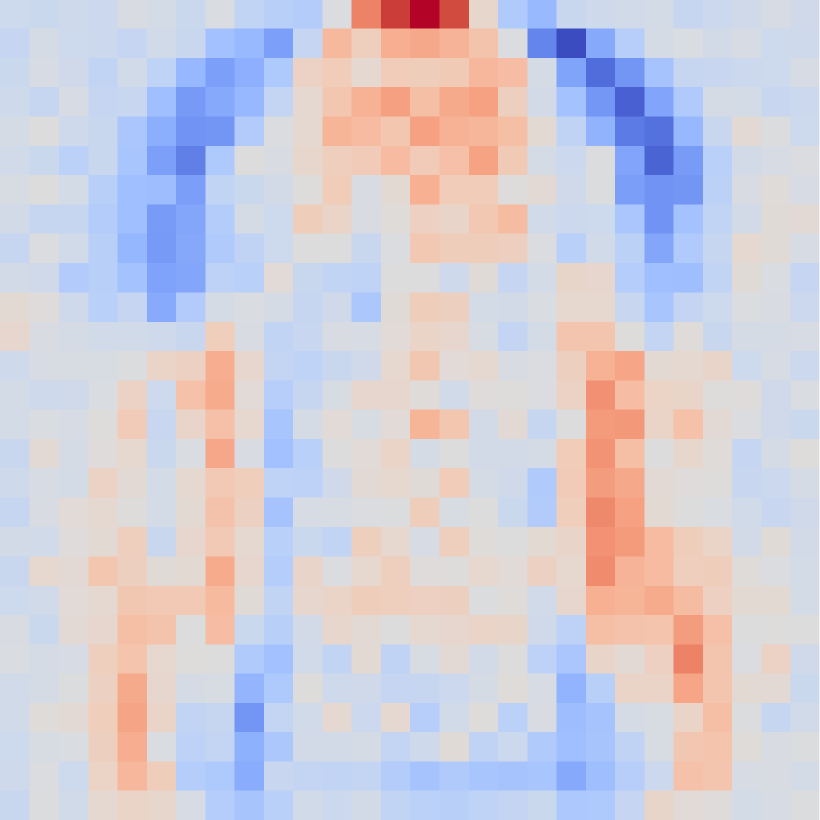} &
        \includegraphics[width=0.115\textwidth]{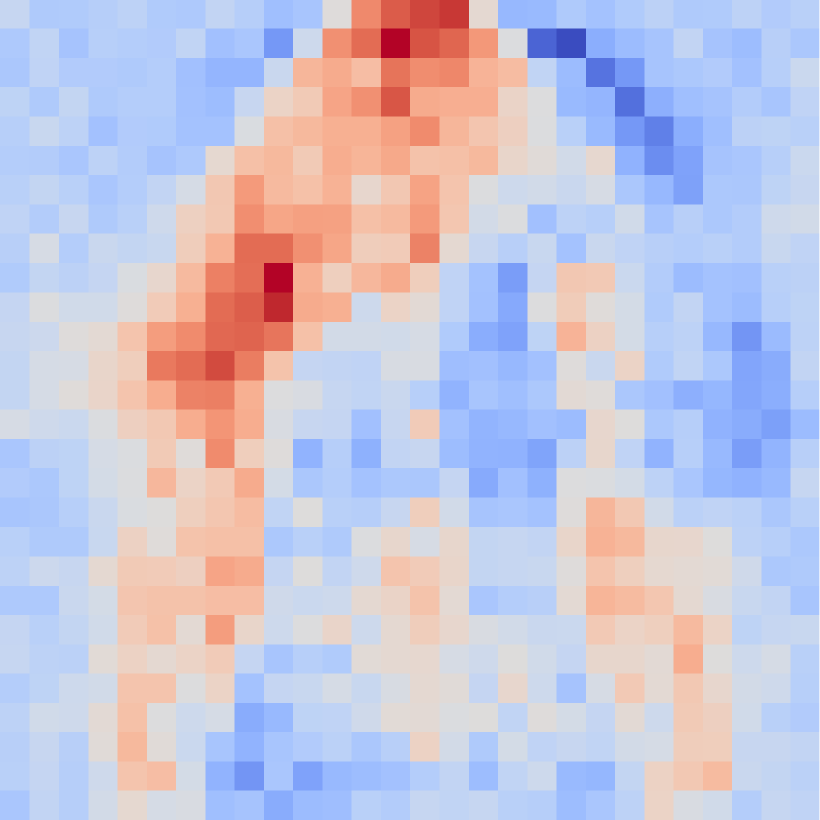} &
        \includegraphics[width=0.115\textwidth]{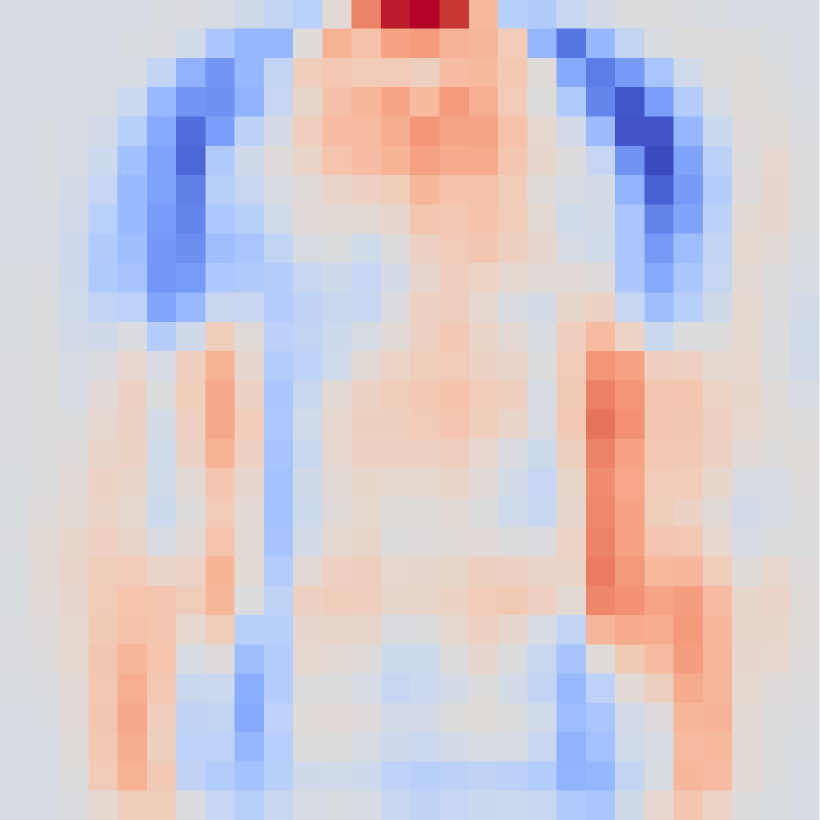} &
        \includegraphics[width=0.115\textwidth]{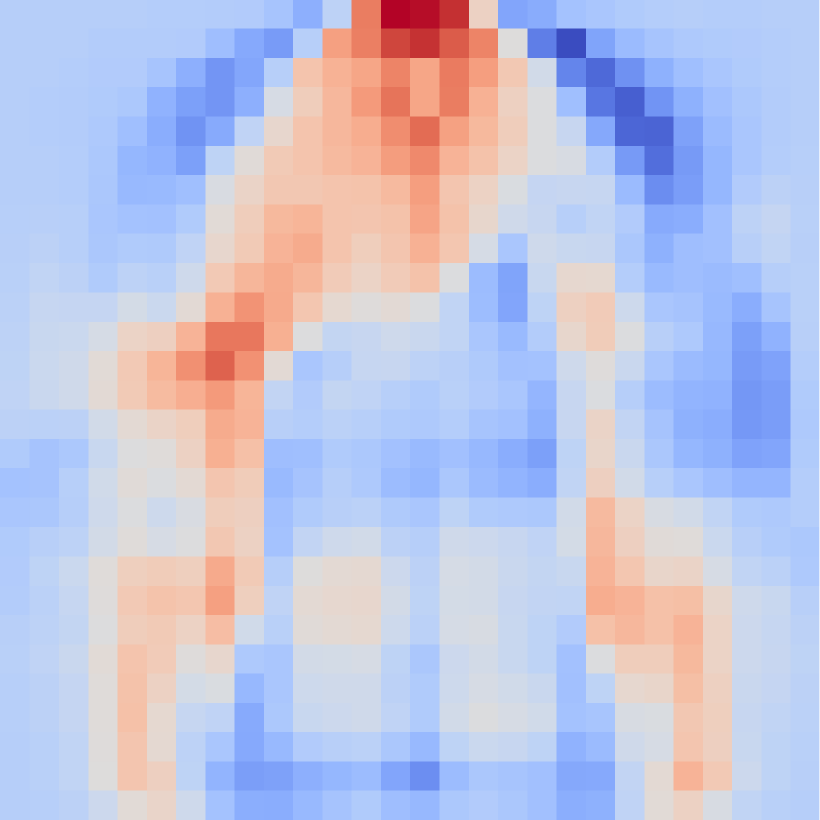} \\[3pt]

        \includegraphics[width=0.115\textwidth]{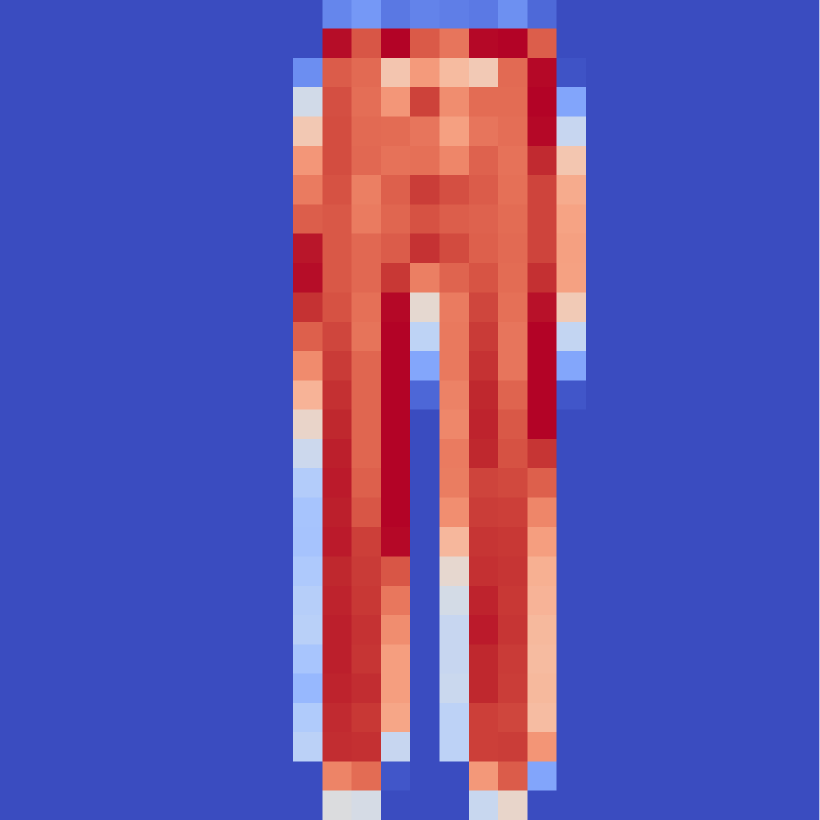} &
        \includegraphics[width=0.115\textwidth]{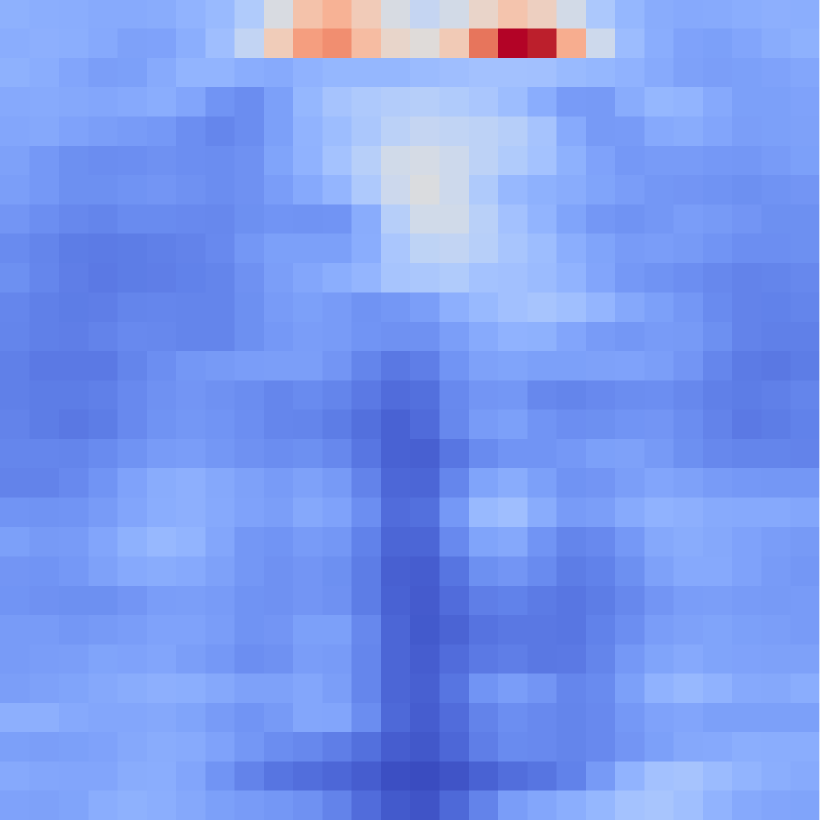} &
        \includegraphics[width=0.115\textwidth]{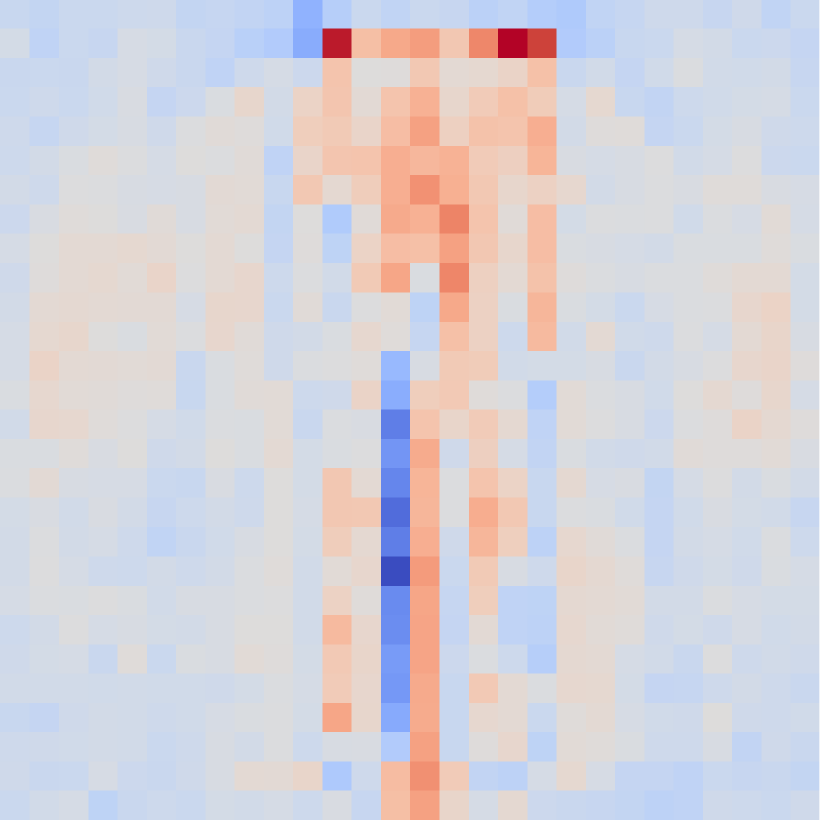} &
        \includegraphics[width=0.115\textwidth]{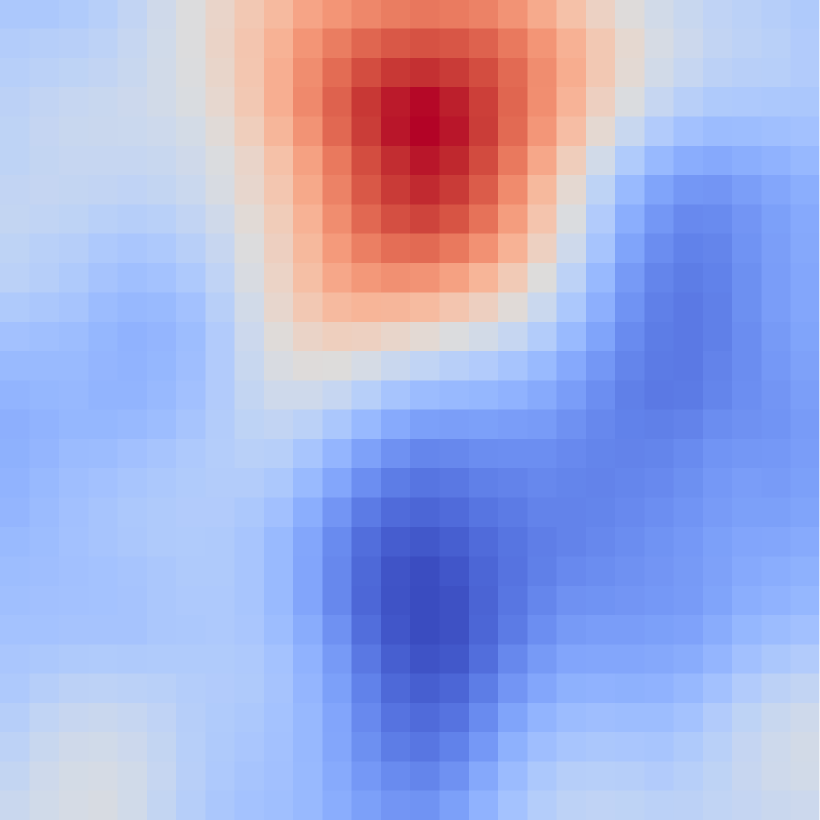} &
        \includegraphics[width=0.115\textwidth]{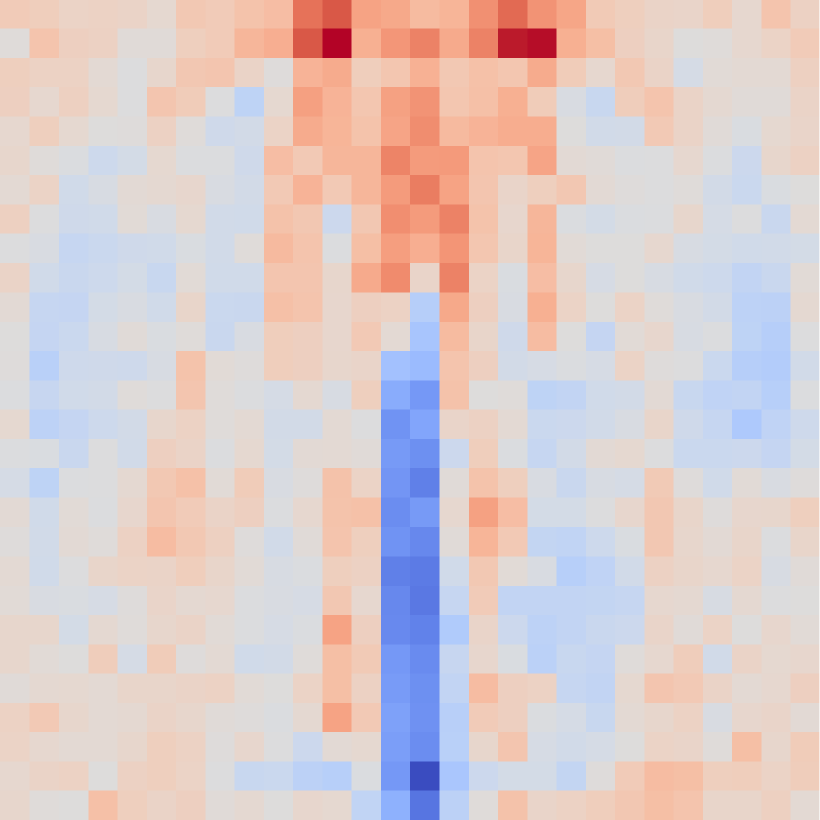} &
        \includegraphics[width=0.115\textwidth]{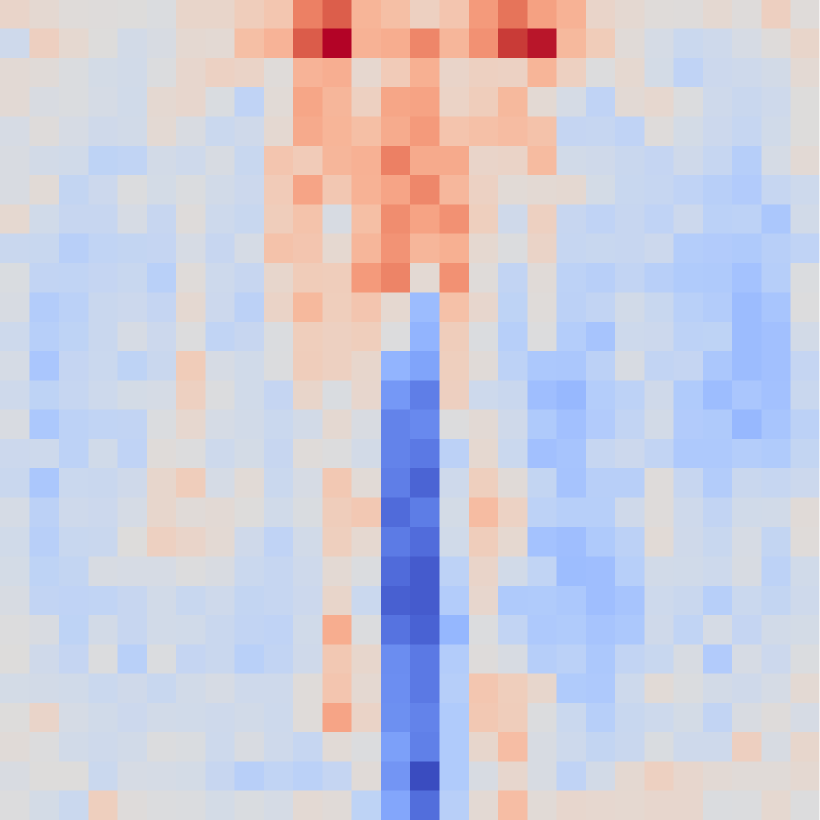} &
        \includegraphics[width=0.115\textwidth]{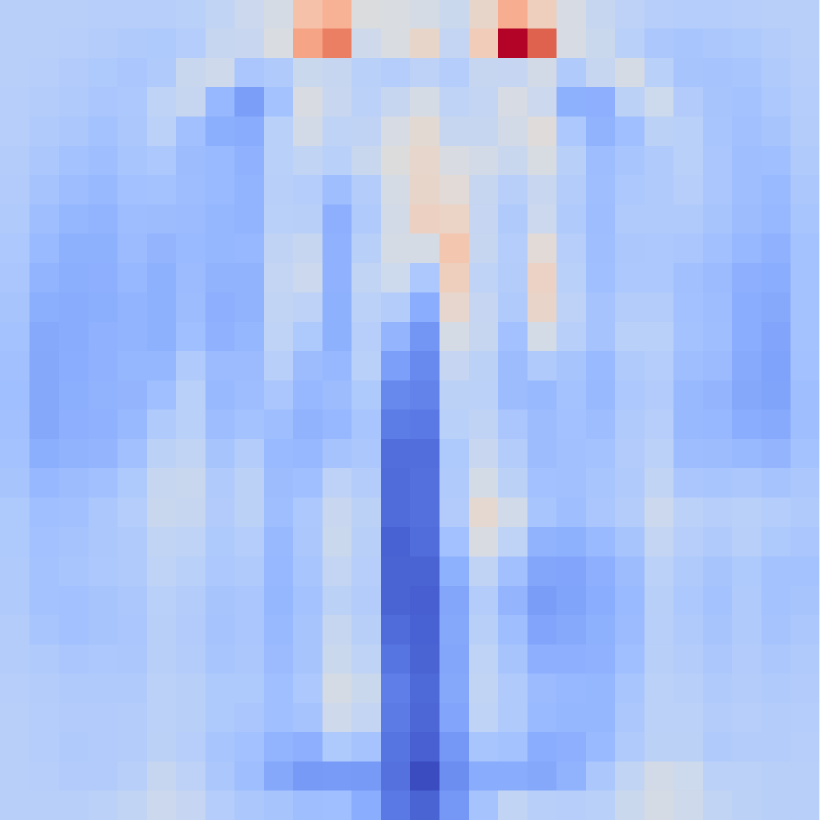} &
        \includegraphics[width=0.115\textwidth]{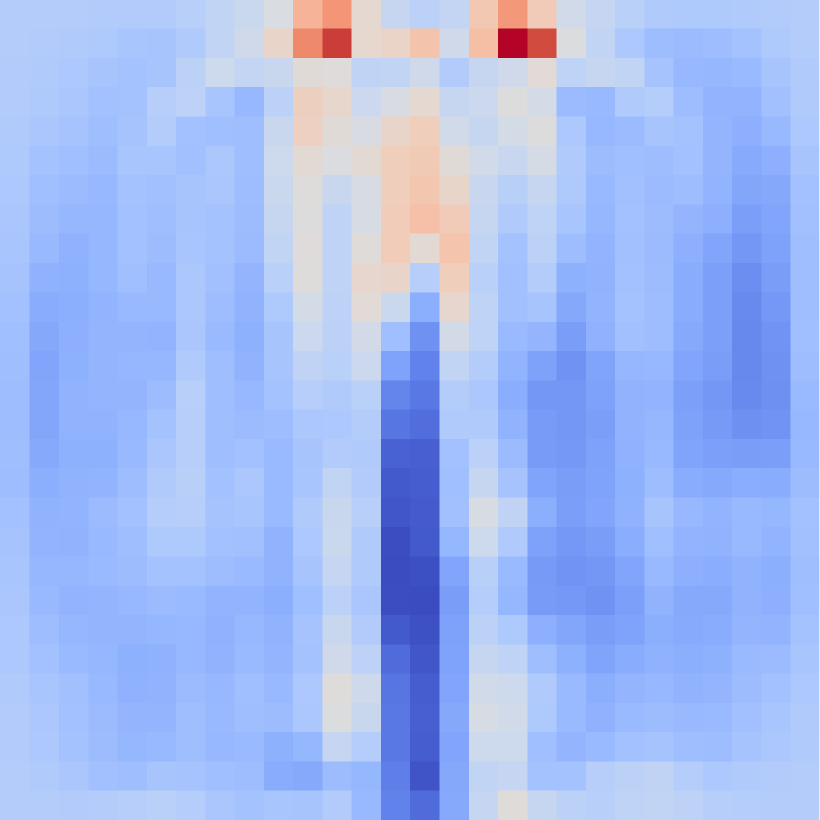} \\[3pt]

        \includegraphics[width=0.115\textwidth]{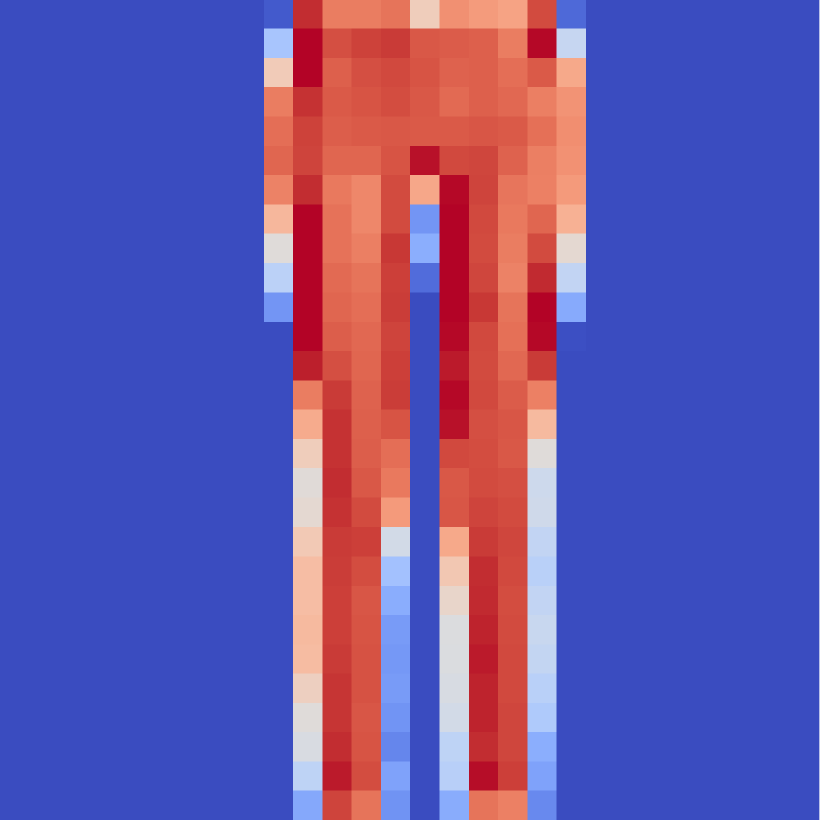} &
        \includegraphics[width=0.115\textwidth]{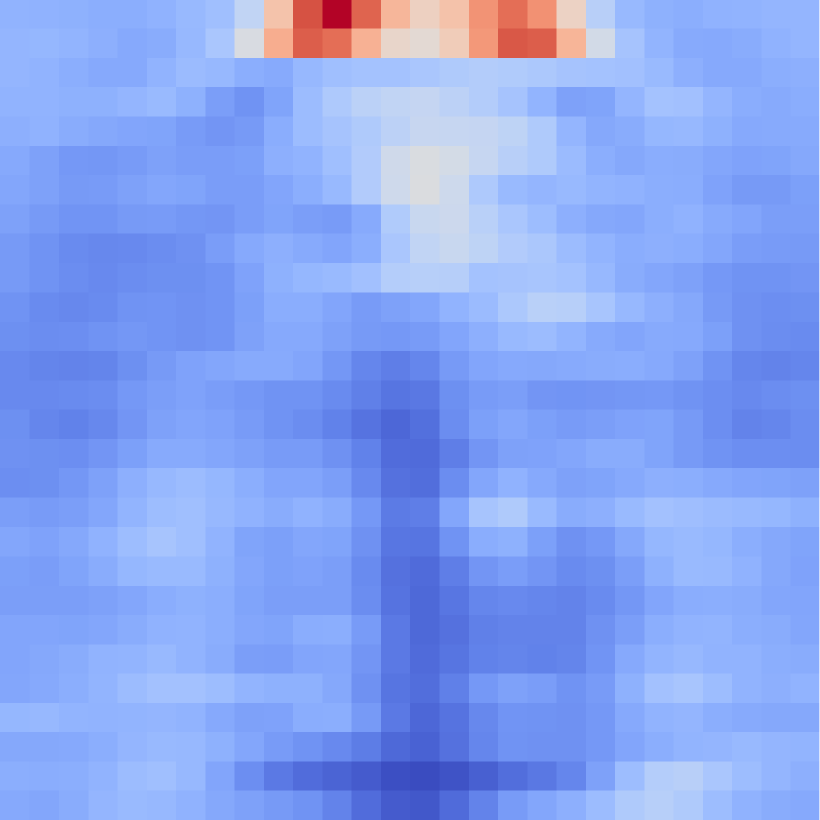} &
        \includegraphics[width=0.115\textwidth]{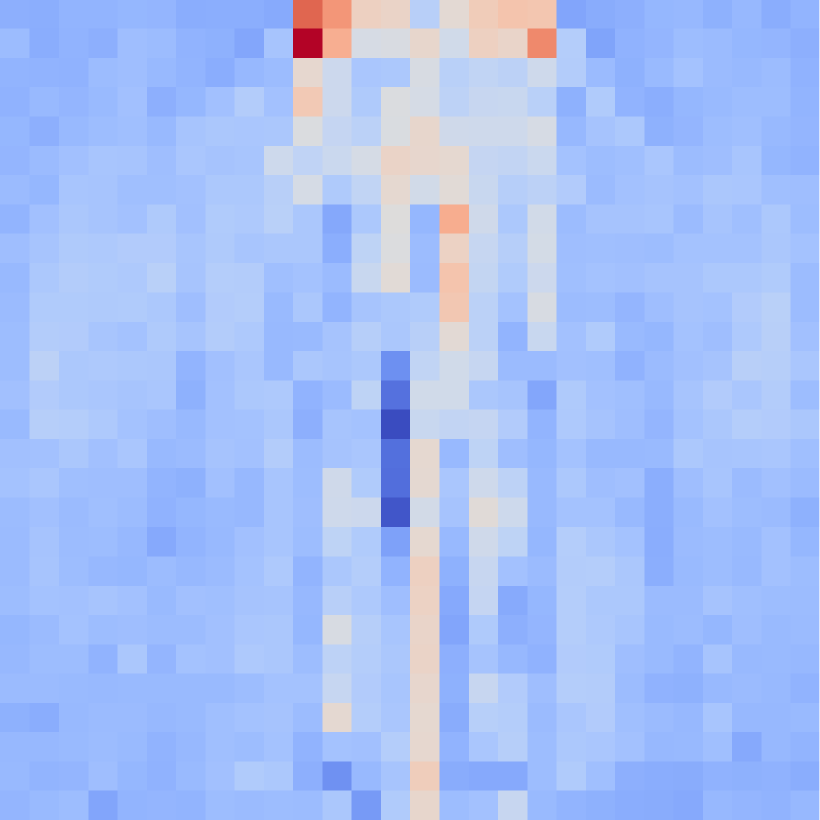} &
        \includegraphics[width=0.115\textwidth]{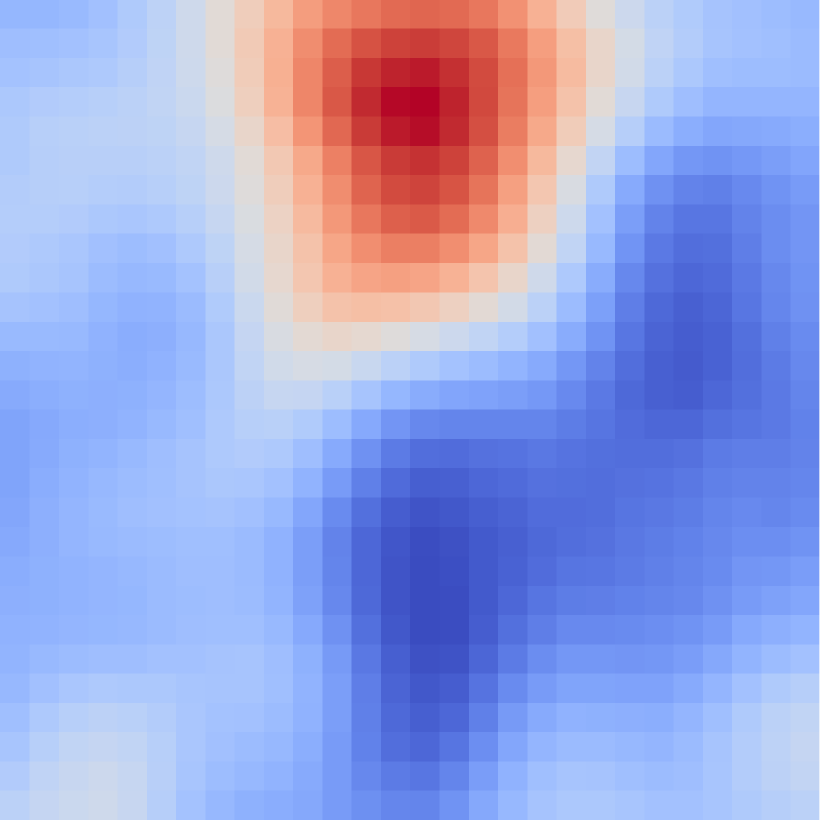} &
        \includegraphics[width=0.115\textwidth]{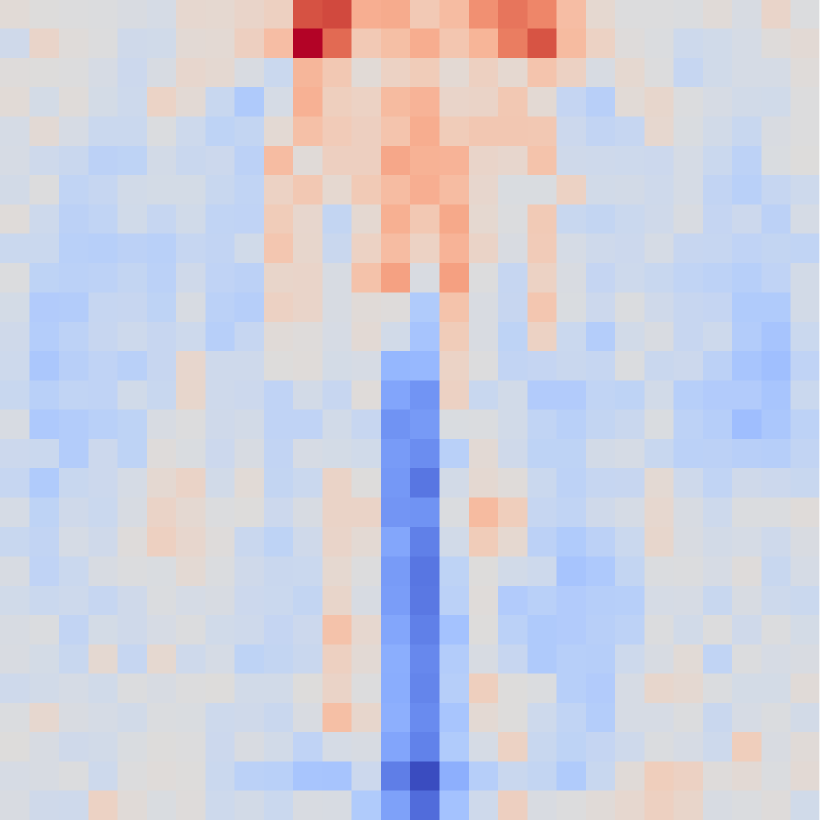} &
        \includegraphics[width=0.115\textwidth]{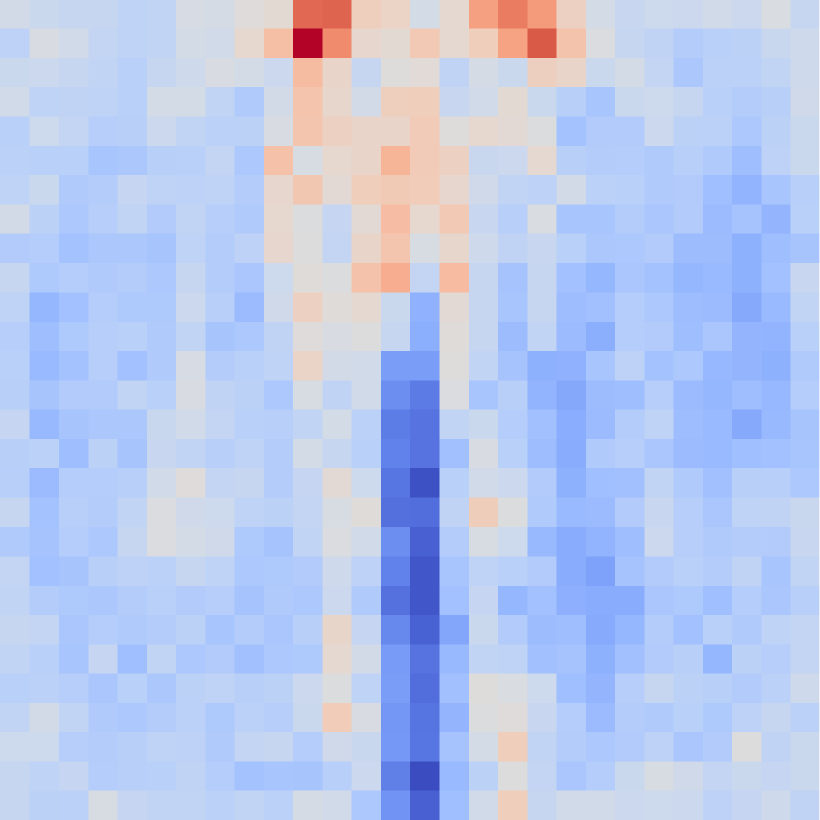} &
        \includegraphics[width=0.115\textwidth]{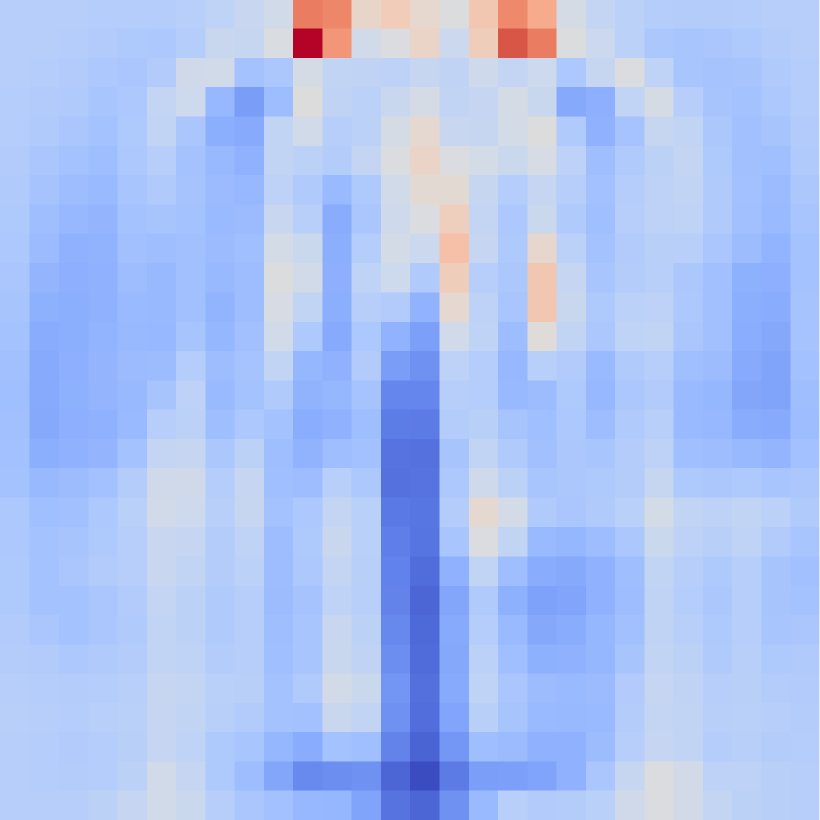} &
        \includegraphics[width=0.115\textwidth]{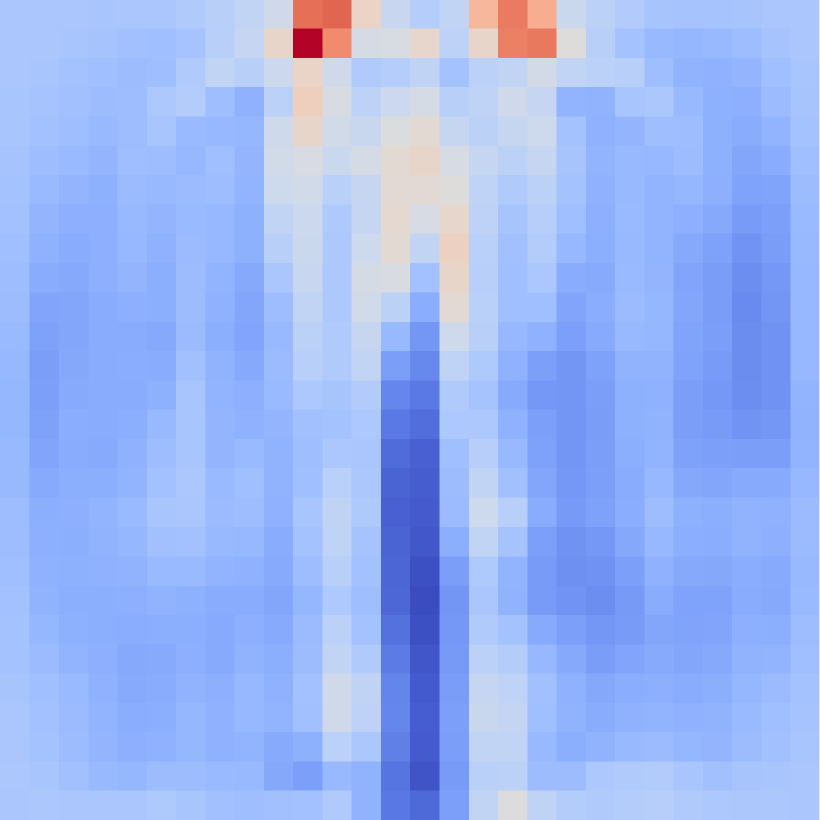} \\[3pt]

        \includegraphics[width=0.115\textwidth]{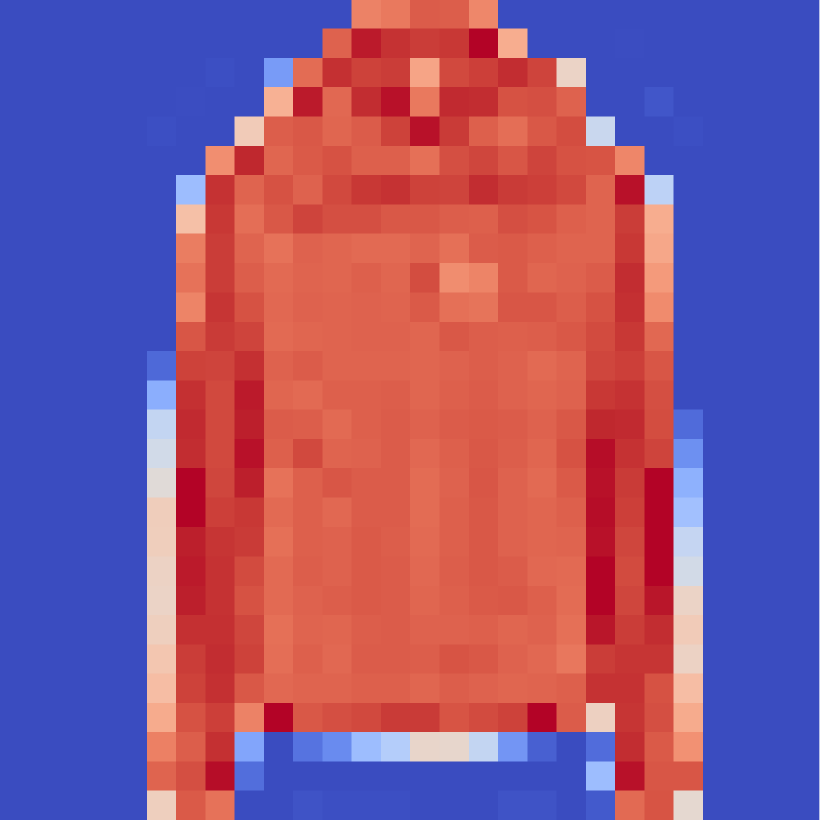} &
        \includegraphics[width=0.115\textwidth]{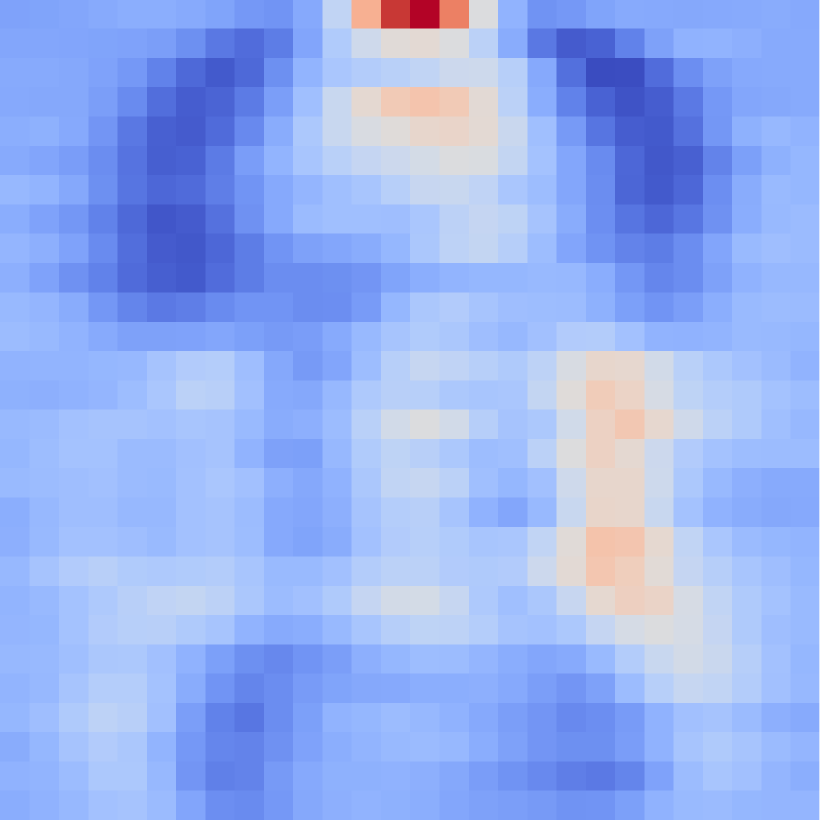} &
        \includegraphics[width=0.115\textwidth]{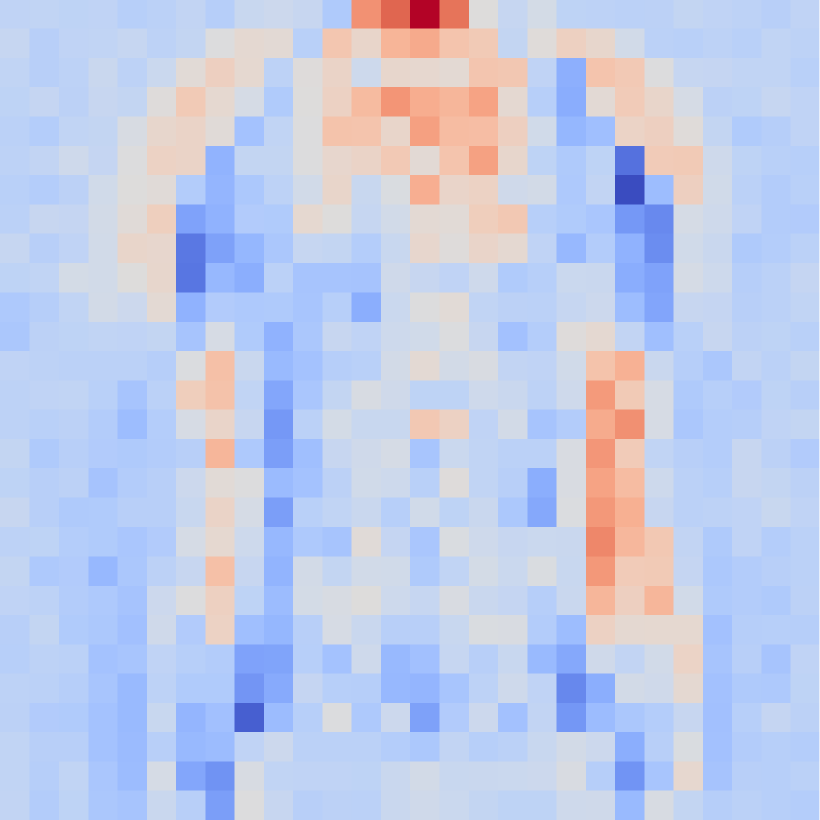} &
        \includegraphics[width=0.115\textwidth]{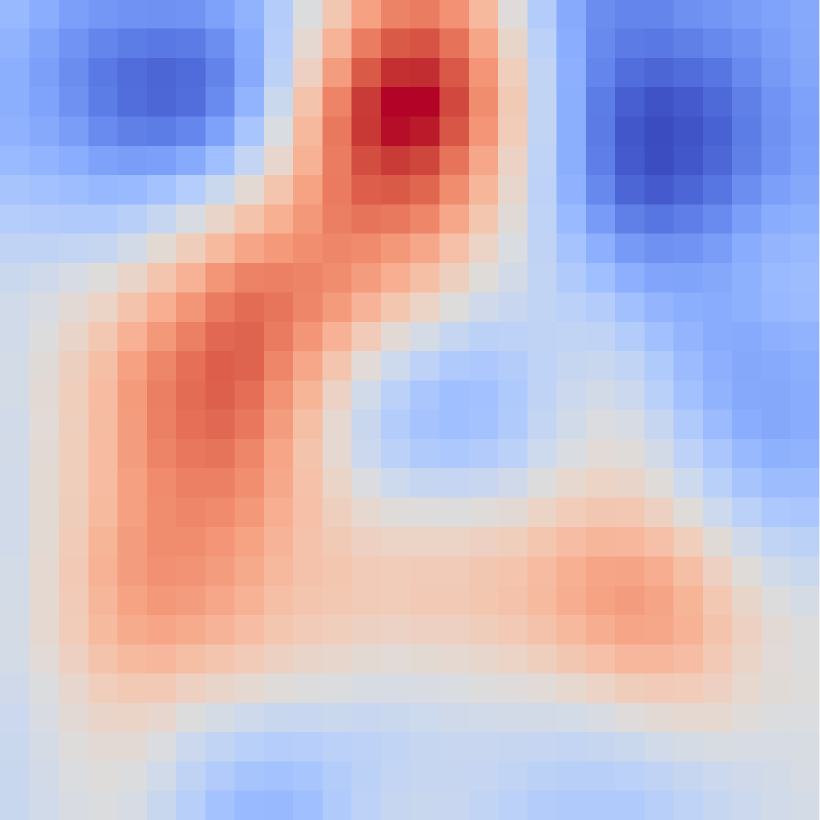} &
        \includegraphics[width=0.115\textwidth]{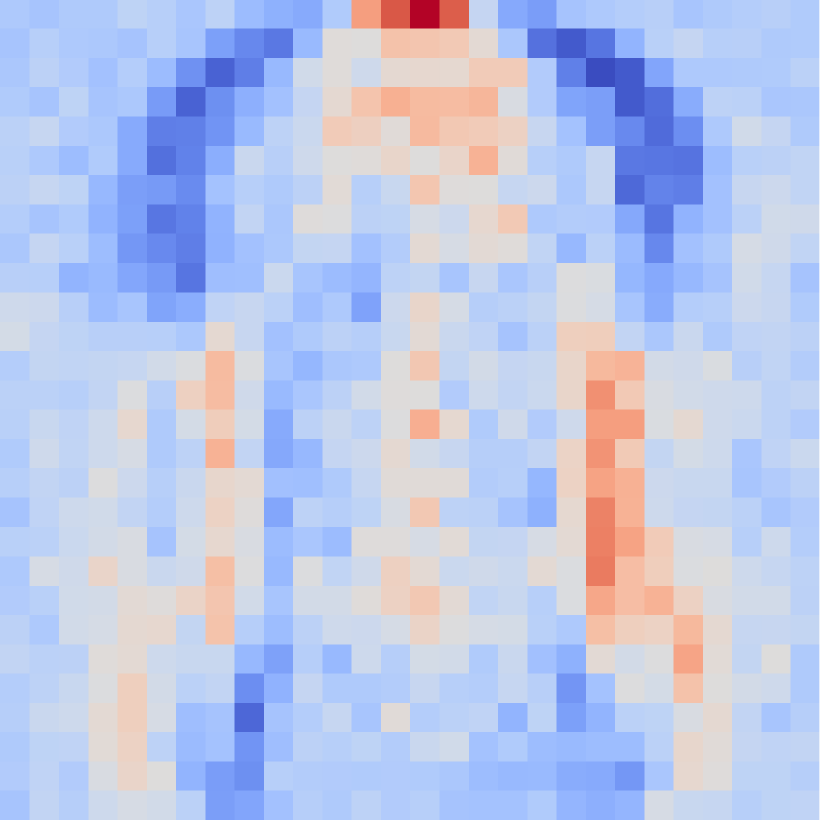} &
        \includegraphics[width=0.115\textwidth]{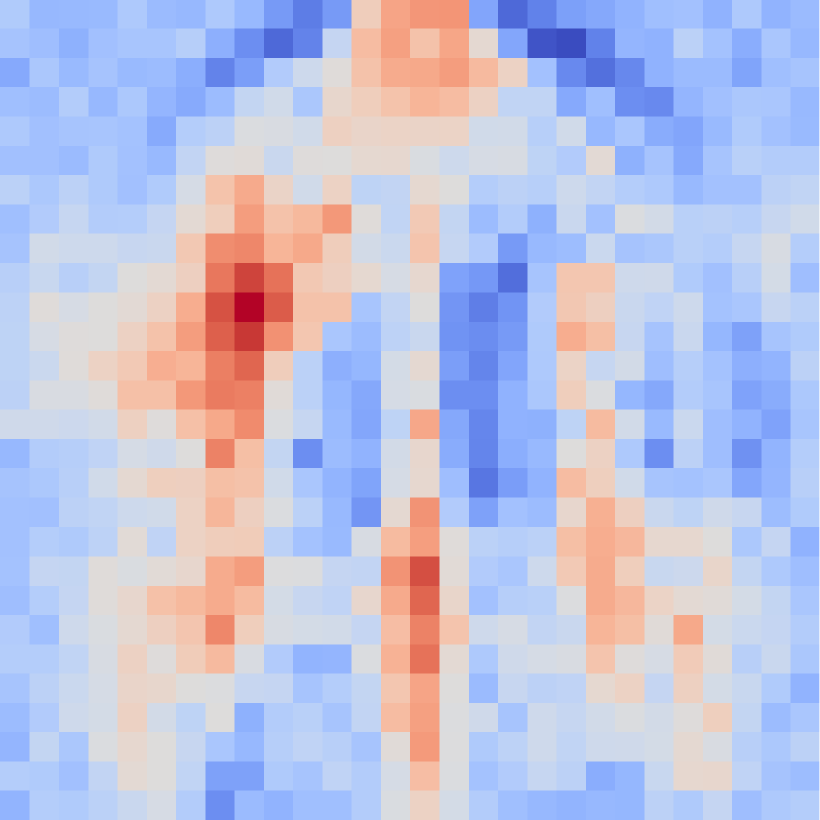} &
        \includegraphics[width=0.115\textwidth]{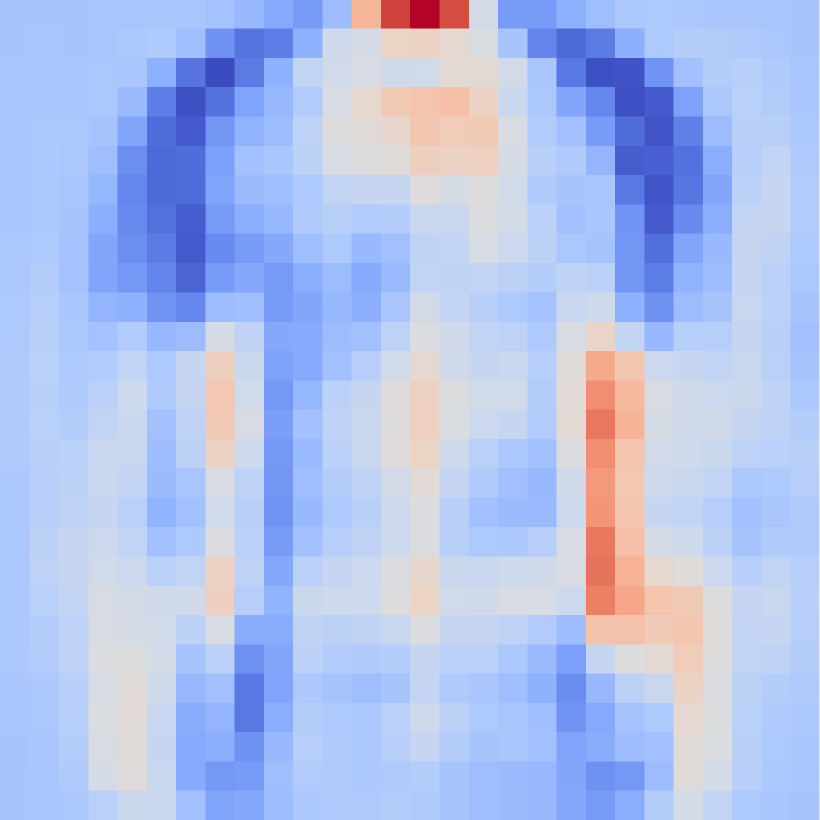} &
        \includegraphics[width=0.115\textwidth]{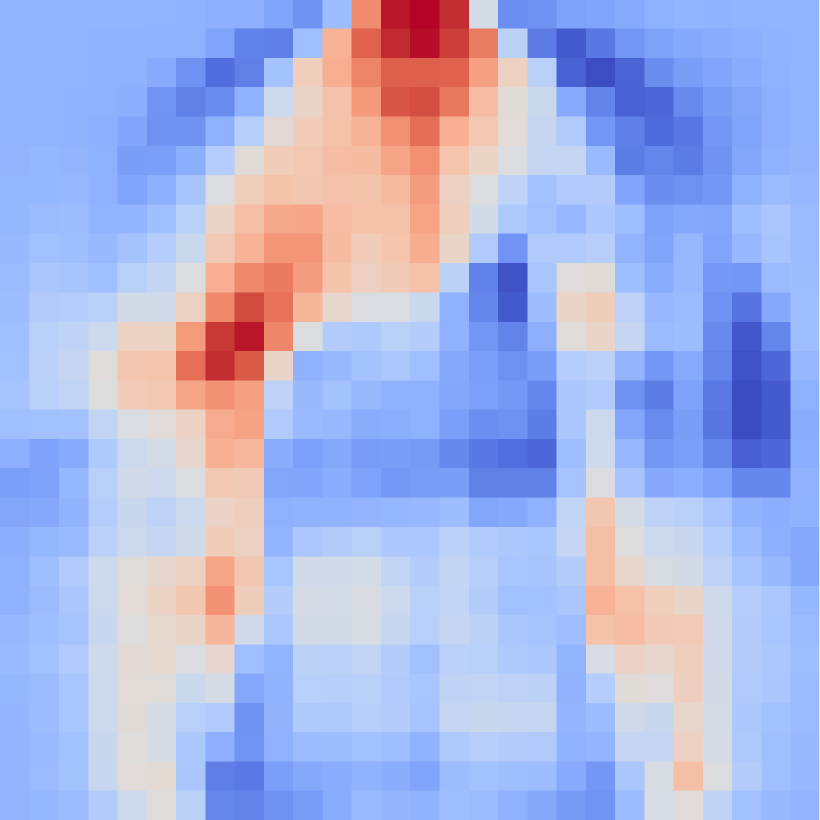} \\[3pt]

        \includegraphics[width=0.115\textwidth]{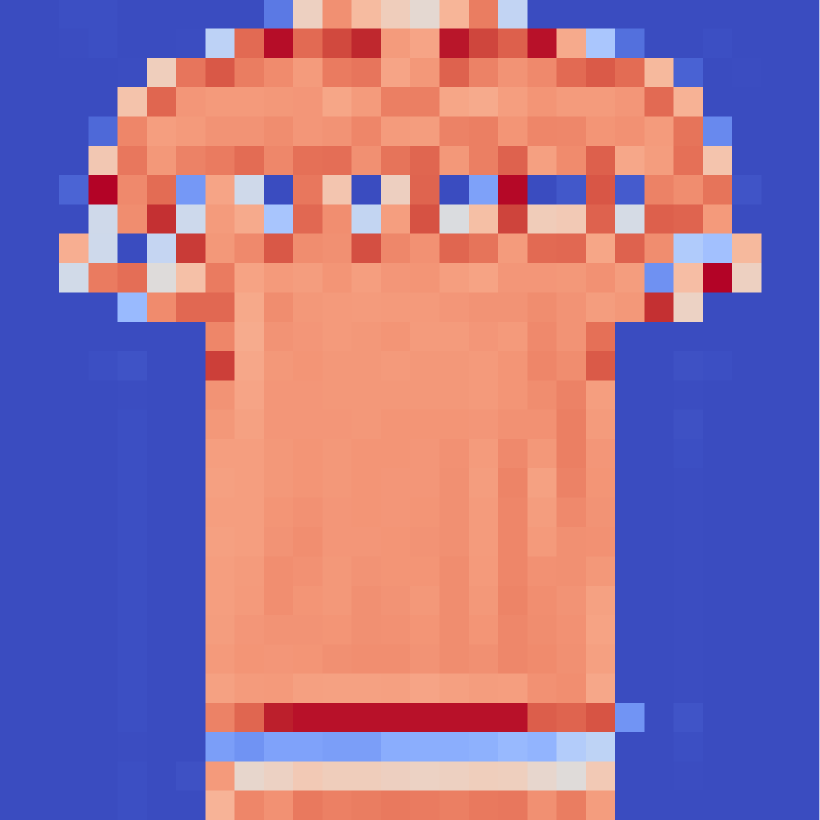} &
        \includegraphics[width=0.115\textwidth]{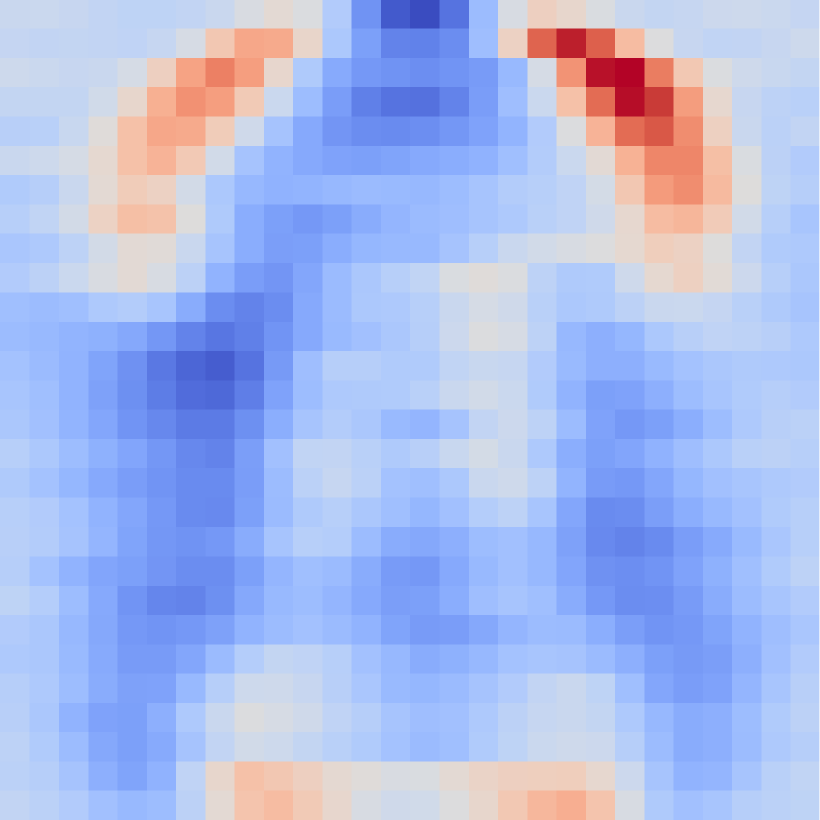} &
        \includegraphics[width=0.115\textwidth]{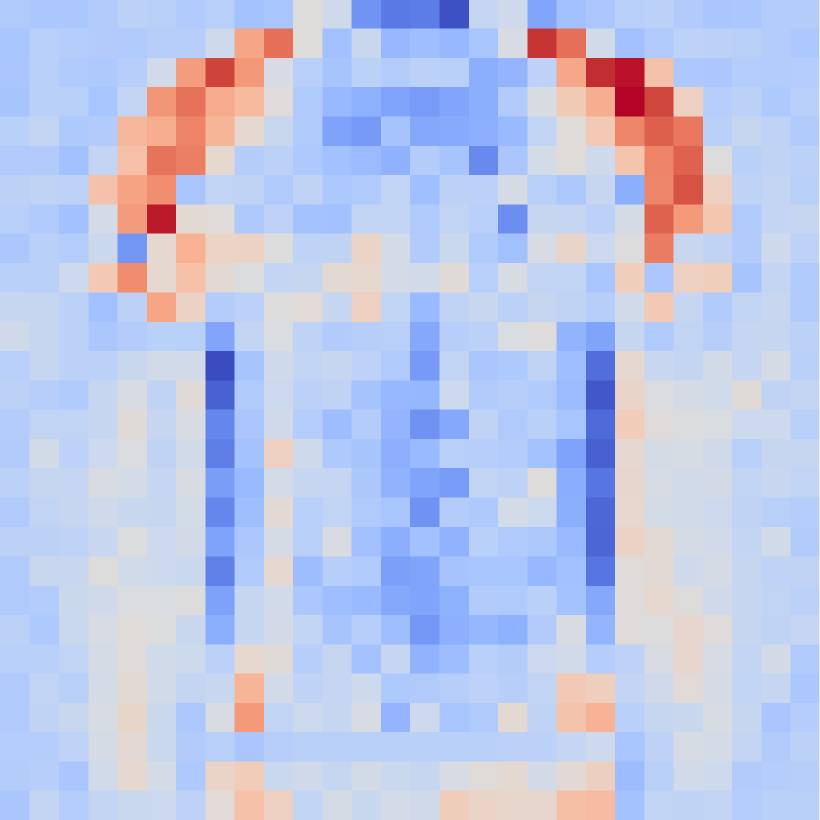} &
        \includegraphics[width=0.115\textwidth]{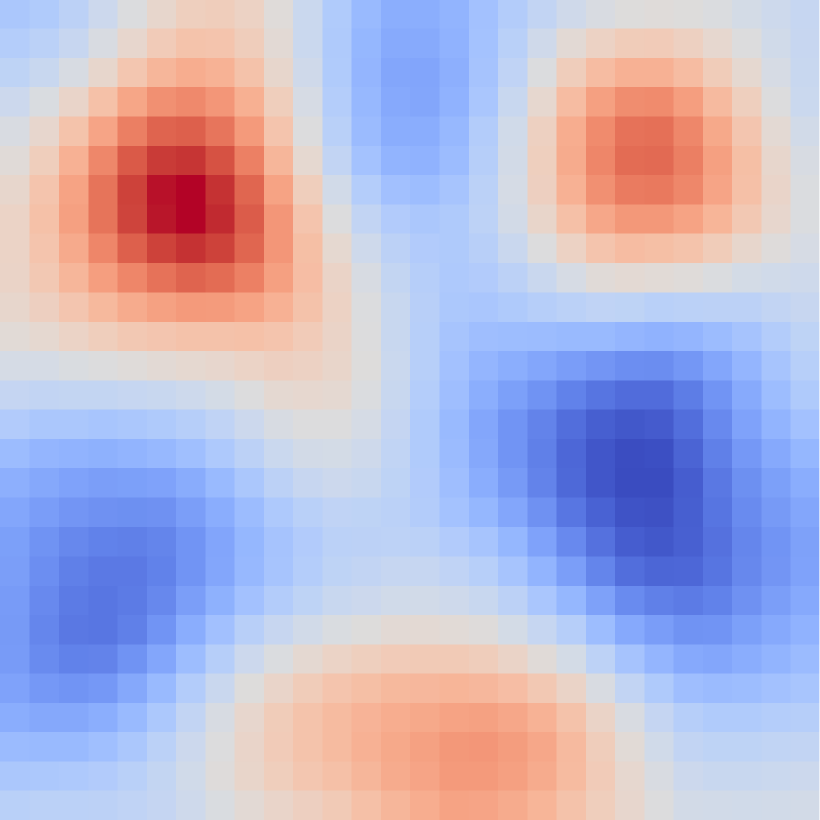} &
        \includegraphics[width=0.115\textwidth]{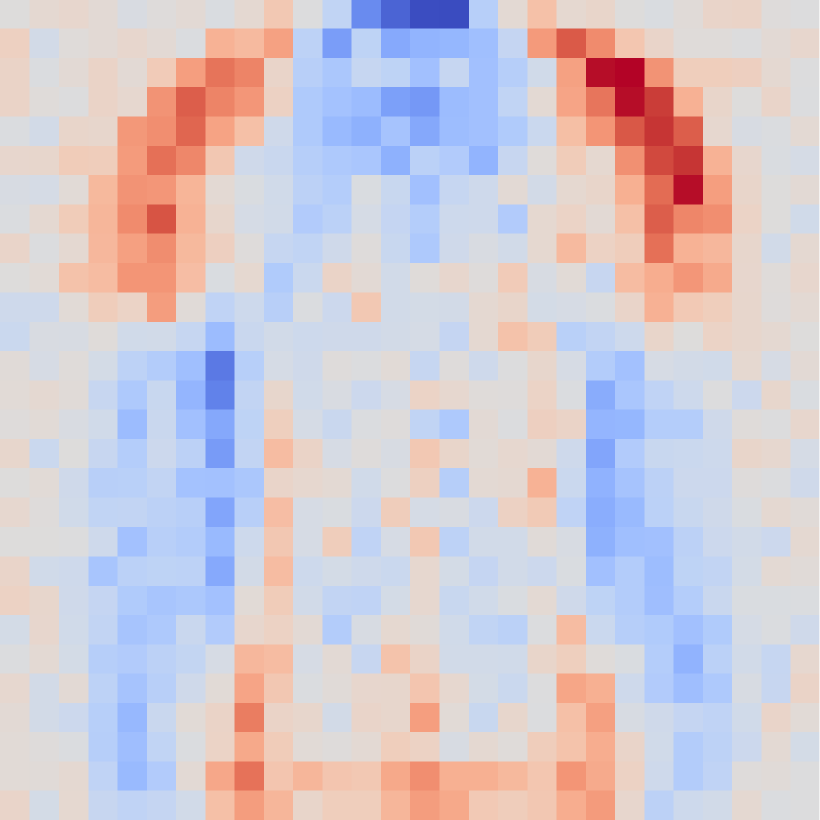} &
        \includegraphics[width=0.115\textwidth]{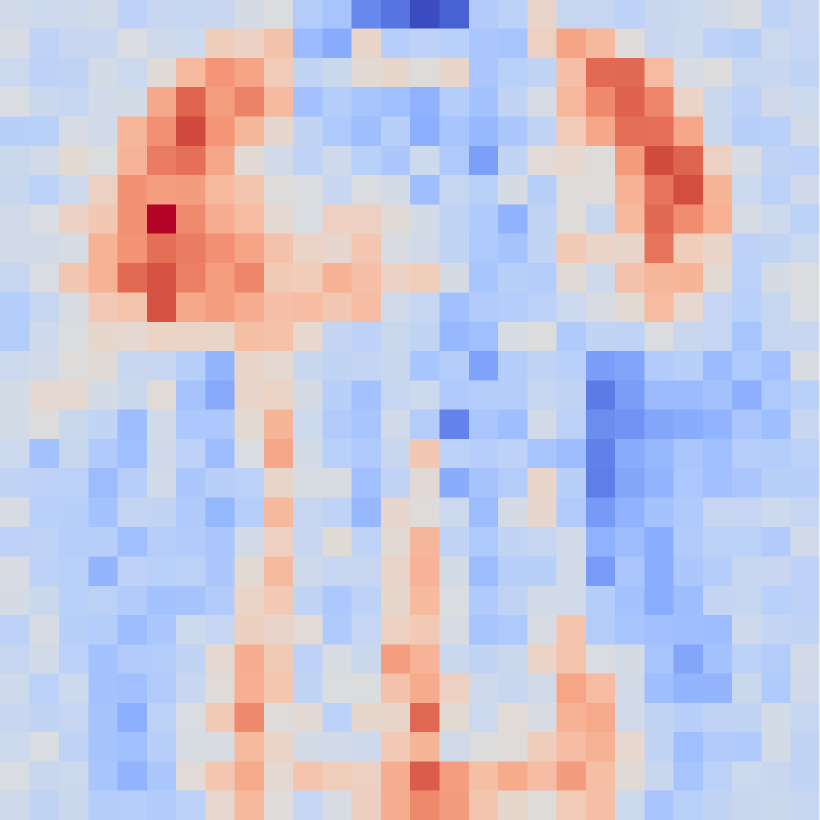} &
        \includegraphics[width=0.115\textwidth]{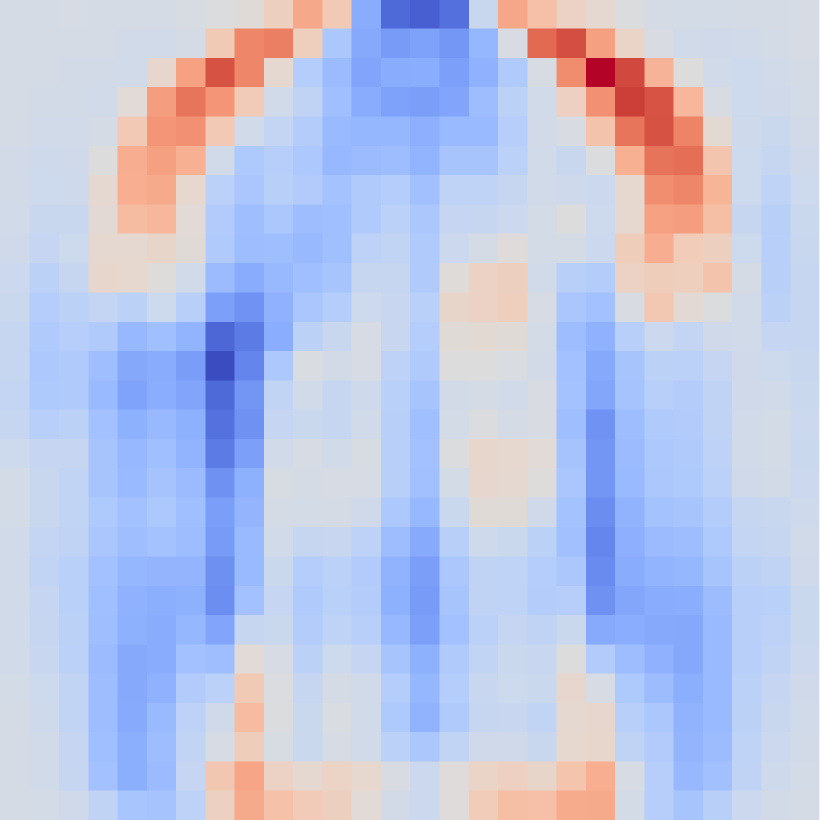} &
        \includegraphics[width=0.115\textwidth]{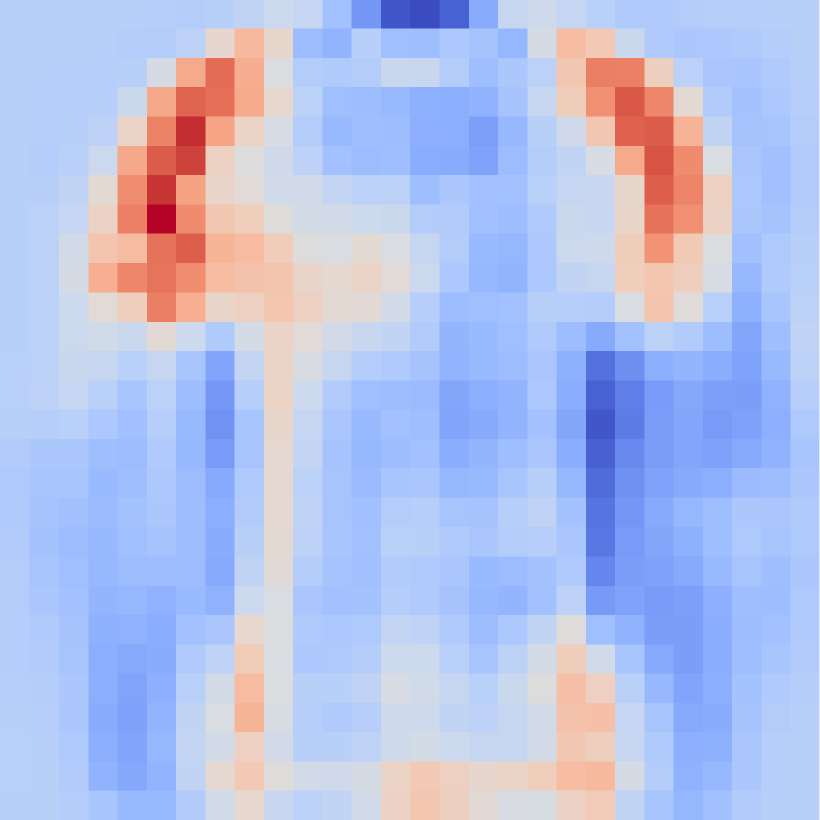} \\[3pt]

        \includegraphics[width=0.115\textwidth]{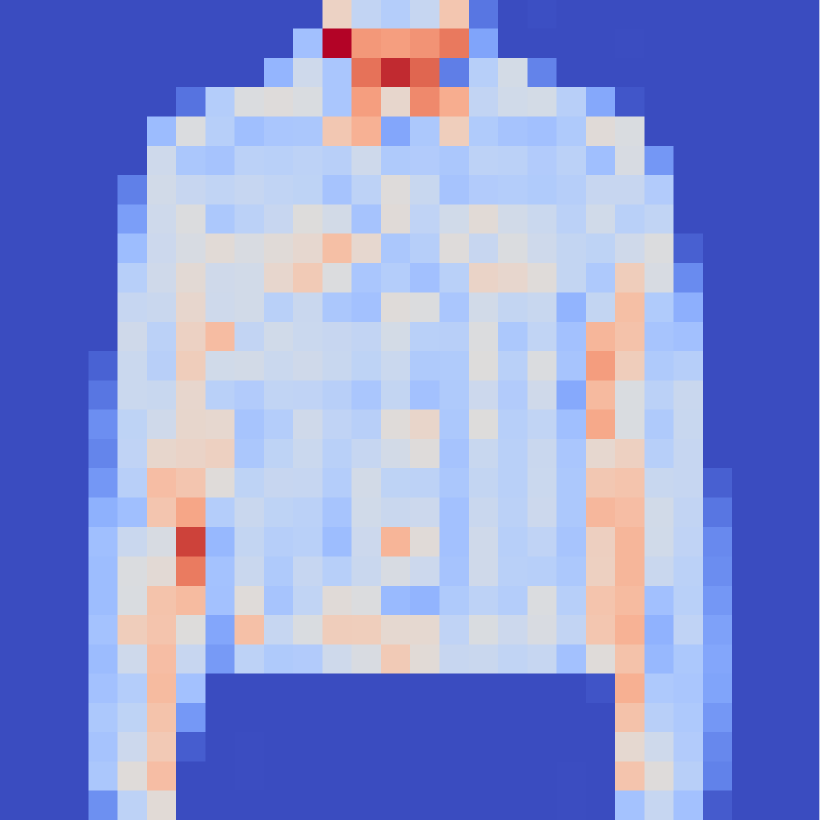} &
        \includegraphics[width=0.115\textwidth]{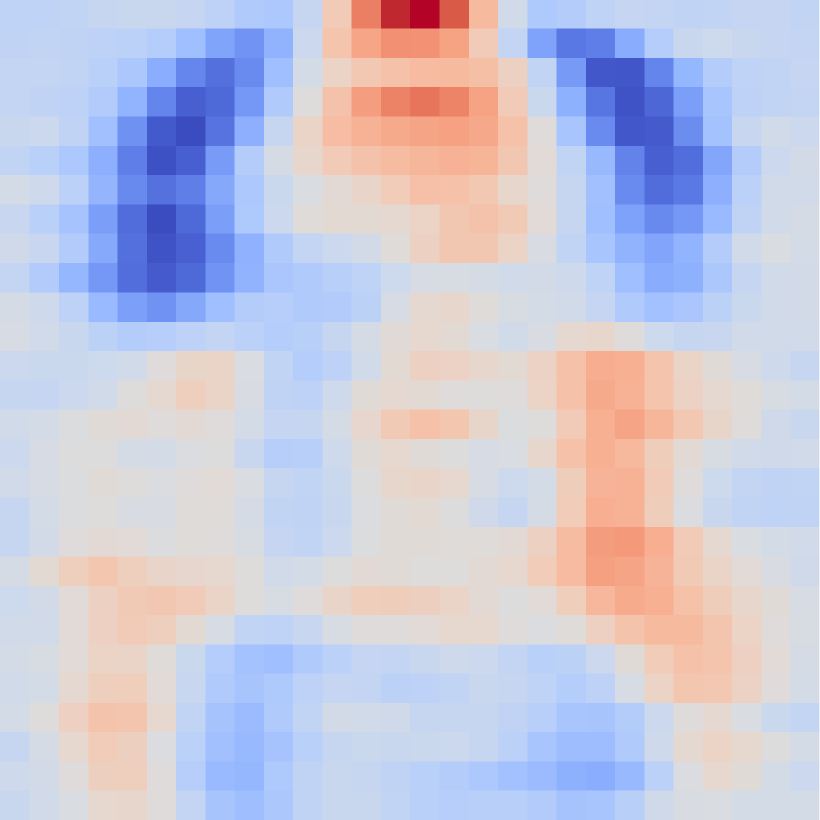} &
        \includegraphics[width=0.115\textwidth]{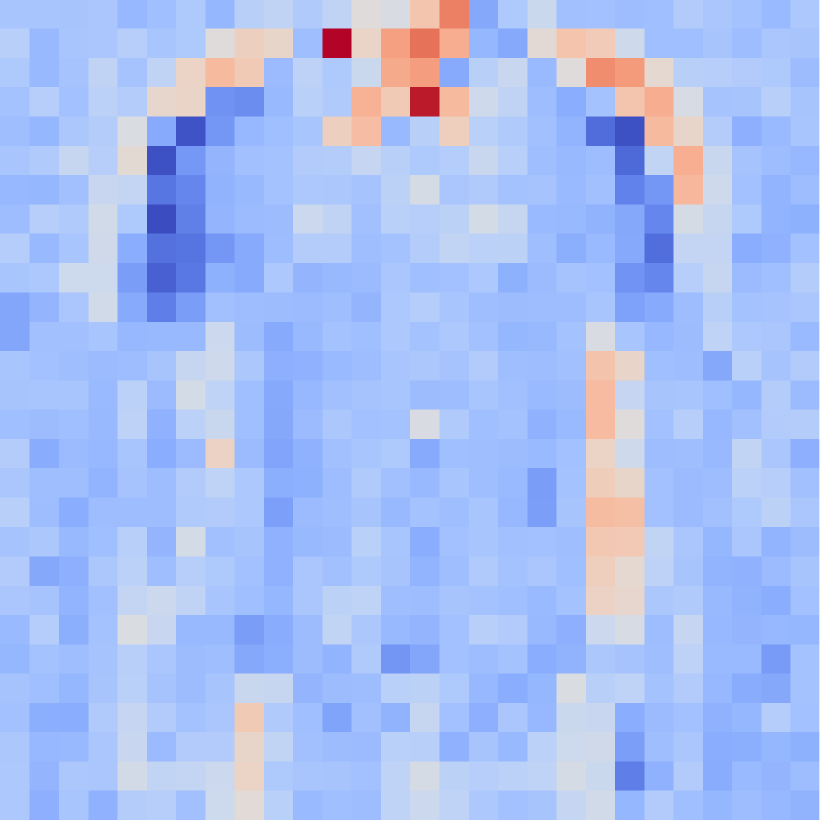} &
        \includegraphics[width=0.115\textwidth]{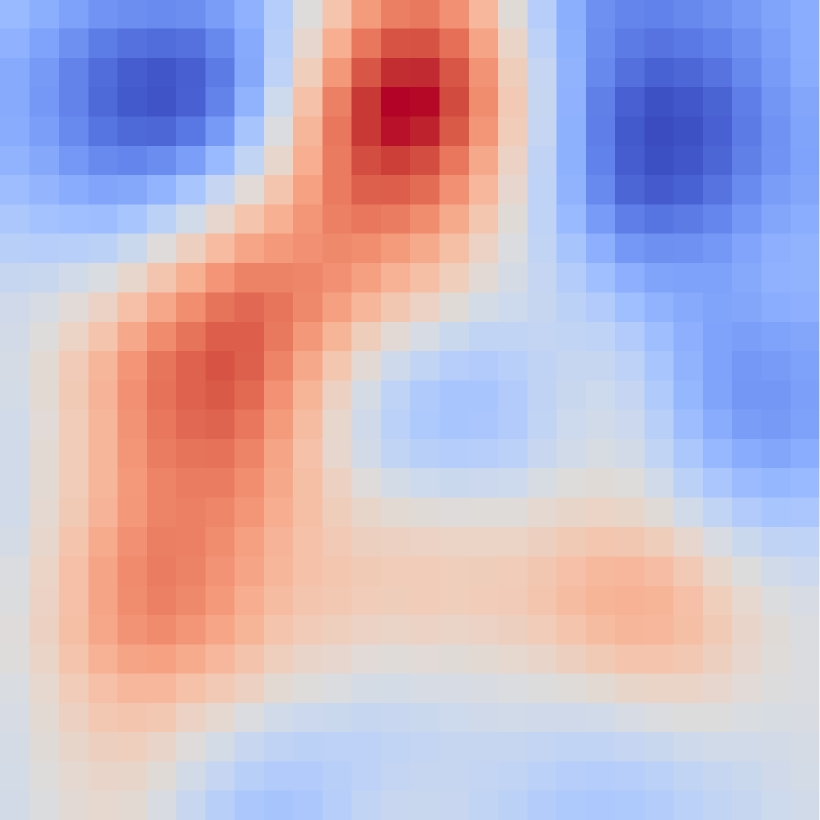} &
        \includegraphics[width=0.115\textwidth]{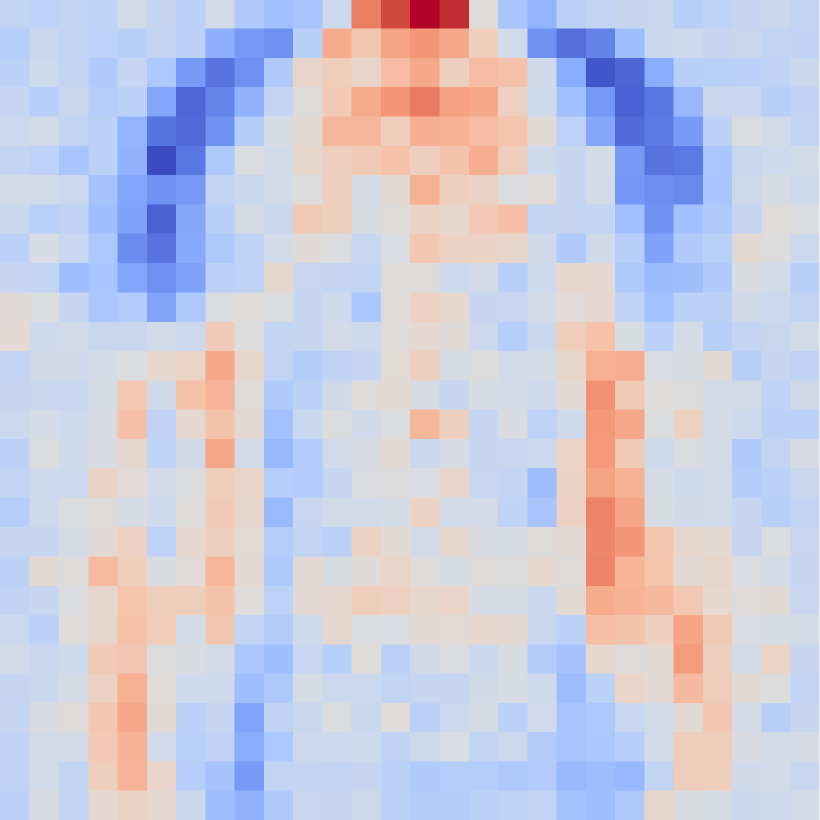} &
        \includegraphics[width=0.115\textwidth]{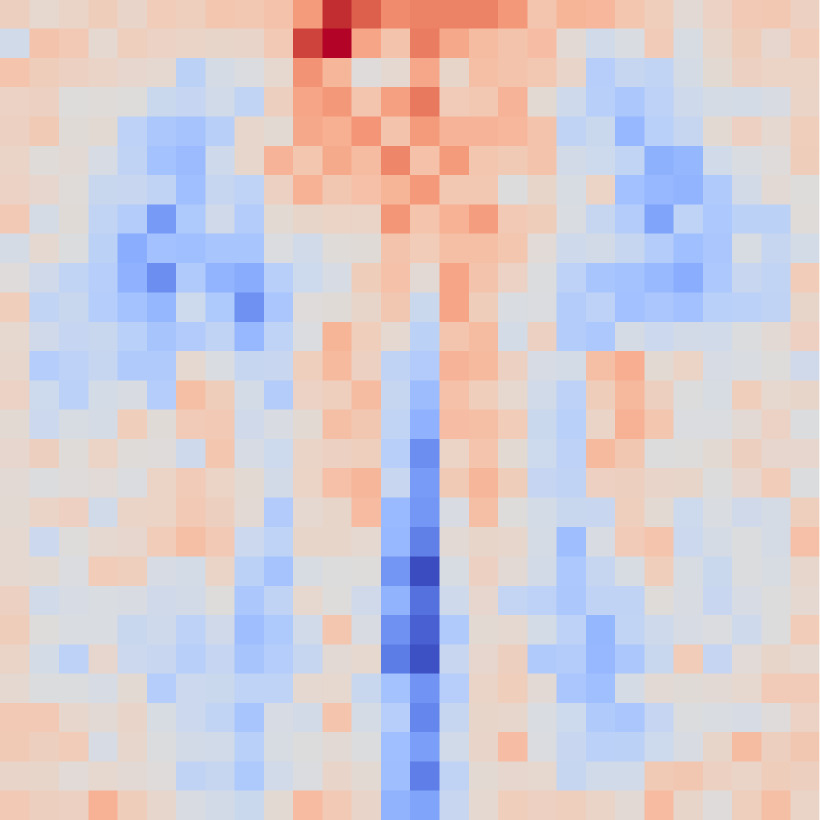} &
        \includegraphics[width=0.115\textwidth]{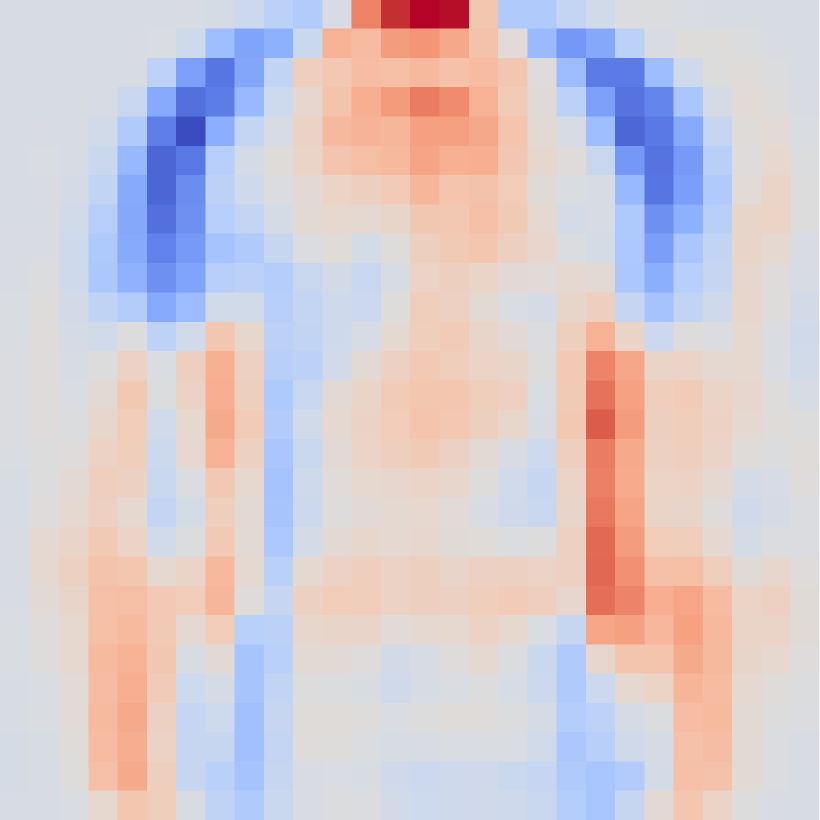} &
        \includegraphics[width=0.115\textwidth]{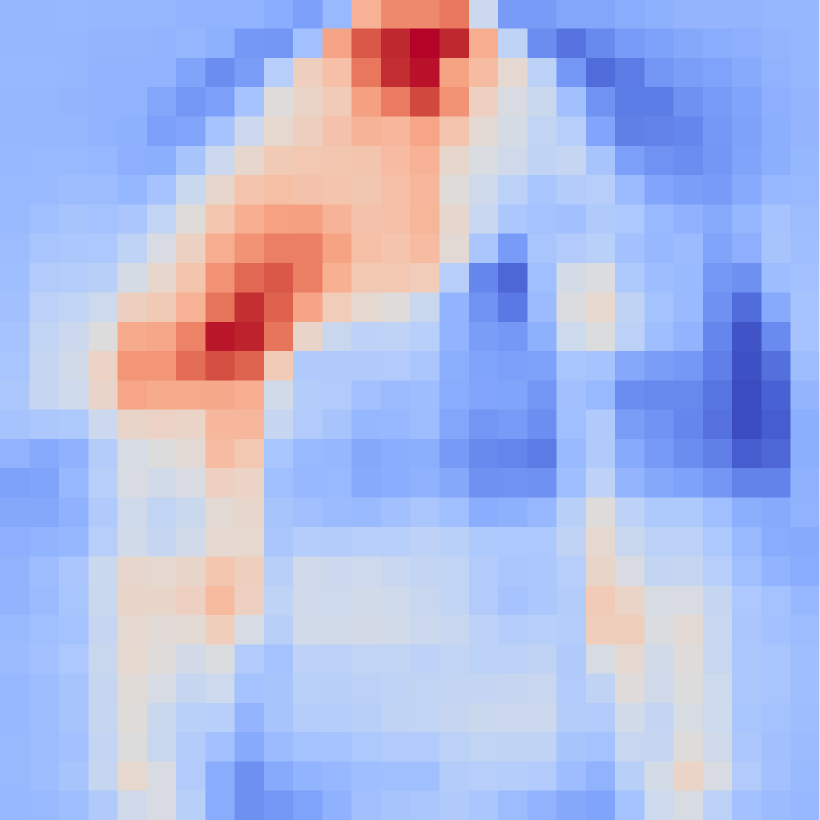} \\[3pt]
       
    \end{tabular}

    \caption{The importance scores of different attribution methods for the Fashion MNIST datasets. Both XWP and XWP\textsuperscript{c} 
    align with other pattern recognition techniques, yet they produce heatmaps that are significantly cleaner and more interpretable.}
    \label{fig:fmnist}
    \Description{A grid of 9x8 importance scores of different attribution methods for the Fashion MNIST datasets. Starting
    from the leftmost column, the image sample is presented, followed by different attribution methods: Occlusion, Shapley Values, Integrated Gradients, 
    LRP, XWP and XWP\textsuperscript{c}. Both XWP and XWP\textsuperscript{c} align with other pattern recognition techniques, yet they produce heatmaps 
    that are significantly cleaner and more interpretable.}

\end{figure*}

\end{document}